%% file: main.tex
\documentclass[10pt,twocolumn,letterpaper]{article}

\usepackage[pagenumbers]{cvpr} 

\usepackage{multirow}
\usepackage{bbm}
\newcommand{\ind}{\mathbbm{1}}
\usepackage{booktabs}   
\usepackage{tabularx}   
\usepackage{graphicx}   
\usepackage{pifont}     
\usepackage{xcolor}     
\usepackage{colortbl}   

\usepackage[nointegrals]{wasysym}
\newcommand{\pmark}{\LEFTcircle}

\usepackage[table]{xcolor}
\definecolor{goodat}{HTML}{FAF0F3}
\definecolor{worseat}{HTML}{E3F2FD}
\newcommand{\good}[1]{\cellcolor{goodat}#1}
\newcommand{\bad}[1]{\cellcolor{worseat}#1}

\usepackage[most]{tcolorbox}

\newcommand{\question}[1]{\par\medskip\noindent\colorbox{gray!12}{\parbox{\dimexpr\linewidth-2\fboxsep\relax}{\textbf{#1}}}\par\nobreak\smallskip\noindent\ignorespaces}

\newtcolorbox{promptbox}[2][]{enhanced, breakable,
  colback=white, colframe=black!60, boxrule=0.5pt, arc=0pt, sharp corners,
  colbacktitle=white, coltitle=black, fonttitle=\bfseries\footnotesize, titlerule=0.3pt, titlerule style={black!60},
  left=6pt, right=6pt, top=3pt, bottom=3pt, toptitle=2pt, bottomtitle=2pt,
  fontupper=\footnotesize\ttfamily, title={#2}, #1}

\definecolor{qualpass}{HTML}{42B47B}
\definecolor{qualphys}{HTML}{EF7462}
\definecolor{qualsa}{HTML}{4A98ED}
\definecolor{qualavoid}{HTML}{9E9E9E}

\newcommand{\cmark}{\textcolor{green!50!black}{\ding{51}}}
\newcommand{\xmark}{\textcolor{red!75!black}{\ding{55}}}

\definecolor{cvprblue}{rgb}{0.21,0.49,0.74}
\usepackage[pagebackref,breaklinks,colorlinks,allcolors=cvprblue]{hyperref}

\def\confName{CVPR}
\def\confYear{2027}

\title{PhyRubric: Rubric-Grounded Diagnostic Evaluation of Physical Commonsense in Text-to-Video Generation}

\author{
Huiyi Chen \quad Weitai Kang \quad Shicheng Fan \quad Xinshuo Wang \\
Zhixiang Shen \quad Yan Yan \quad Philip S. Yu \quad Baoxin Li \\[4pt]
Department of Computer Science, University of Illinois Chicago\\
{\tt\small \{hchen270, wkang126, sfan25, xwang921, zshen33, yyan55, psyu, baoxinli\}@uic.edu} \\[2pt]
{\small\url{https://github.com/chenyil6/phyRubric}}
}

\begin{document}
\maketitle
\input{sec/0_abstract}    
\input{sec/1_intro}

\input{sec/2_related_work}

\input{sec/3_phyRubric}

\input{sec/4_experiment}

\input{sec/5_conclusion}

\input{sec/6_llm_statement}

\input{sec/ack}

{
    \small
    \bibliographystyle{ieeenat_fullname}
    \bibliography{main}
}

\input{sec/X_suppl}

\end{document}

%% file: sec/0_abstract.tex
\begin{abstract}
As text-to-video (T2V) models are increasingly positioned as world simulators, it is essential to evaluate whether their videos conform to physical commonsense. However, existing benchmarks rely on annotators who rate each video with a single score based on intuition, yielding ratings that are subjective and do not tell which physical principle is violated. We introduce PhyRubric, a rubric-grounded diagnostic framework for evaluating physical commonsense in T2V generation. For each caption, PhyRubric constructs a caption-specific rubric before any video is seen, decomposing the caption into semantic and physical items, and derives scores from item-level verdicts grounded in timestamped evidence. We instantiate this framework as PhyRubric-Bench, which spans 34 categories across five physical domains, and scale it up to PhyRubric-Train with 3,788 videos, on which we fine-tune the open-source PhyRubric-Judge. Empirical results across ten T2V models show that physical commonsense remains far from solved, and that PhyRubric-Judge correlates more closely with human judgments than existing open-source judges while matching the frontier proprietary Gemini 4 Argon. Further analysis provides actionable insights for T2V models.
\end{abstract}

%% file: sec/1_intro.tex
\section{Introduction}
\label{sec:intro}

\begin{figure}[t]
\centering
\includegraphics[width=\linewidth]{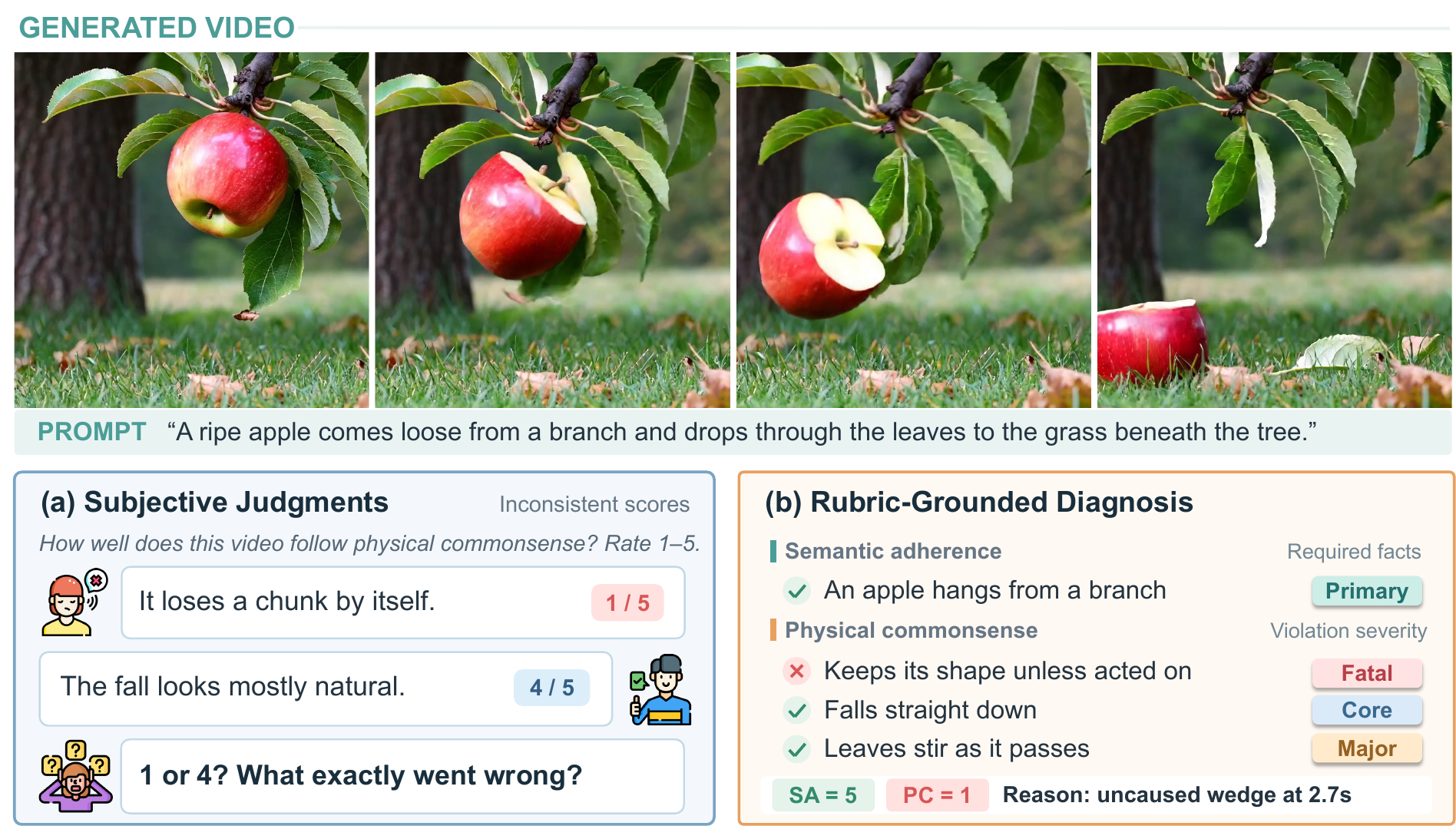}
\caption{Intuition-based rating vs. rubric-grounded evaluation. (a) Two raters give different scores to the same video, and the scores alone do not say what went wrong. (b) Checking the video against a caption-specific rubric item by item yields scores with the violated expectation and its timestamp.}
\label{fig:motivation}
\end{figure}

Text-to-video (T2V) generation has advanced rapidly, and recent T2V models~\cite{wan2025wan,wu2025hunyuanvideo,agarwal2025cosmos,chen2023videocrafter1,zheng2024open,sun2024sora,ma2025step} can now render complex scenes from text prompts. As these models are increasingly positioned as world simulators~\cite{zhu2024sora,kang2024far,agarwal2025cosmos}, a basic requirement is that their videos conform to physical commonsense, the shared expectations about the physical world that everyone acquires from everyday experience, without any textbook physics~\cite{kubricht2017intuitive,spelke2007core}. A reliable and comprehensive evaluation of this ability is therefore essential for tracking progress and guiding model development.

Existing benchmarks for physical commonsense have made notable progress~\cite{bansal2025videophy,bansal2026videophy,meng2024towards,gu2026phyworldbench}, but they share two limitations (Fig.~\ref{fig:motivation}). \textbf{First, the ratings are subjective.} Annotators rate each video and its caption with a single score on a discrete scale, but no rule defines how physical violations map to scores. Each rater thus applies an intuitive mapping, weighing the severity of violations by personal standards, so the same video may receive different scores. This inconsistency introduces noise into both benchmark results and the automatic evaluators trained on these ratings. \textbf{Second, the ratings are uninterpretable.} A discrete scale is easier to rate than a continuous one, since fractional scores would be harder to assign and more scattered. However, a single score reflects only how implausible a video is overall, not which physical principle is violated. Evaluations can thus only rank models and categories, yielding coarse findings that offer little guidance for improvement.


To address these limitations, we introduce \textbf{PhyRubric}, a rubric-grounded diagnostic framework for evaluating physical commonsense in T2V generation (Fig.~\ref{fig:overview}). For each caption, PhyRubric constructs a caption-specific rubric before any video is seen, and reuses it for every video generated from that caption. The rubric decomposes the caption into verifiable items of two types: \textit{semantic items} list the visible facts the caption requires, while \textit{physical items} specify which visible events count as violations and are each assigned a severity level inspired by the order in which infants acquire physical knowledge~\cite{spelke2007core,baillargeon2004infants,lin2022infants}. Given a video, each item receives a verdict grounded in timestamped evidence, and scores are derived deterministically from these verdicts rather than rated directly. Every score thus comes with a diagnosis of which expectation failed and why.

We instantiate this framework as \textbf{PhyRubric-Bench}. Unlike prior benchmarks dominated by mechanics (Suppl.~Tab.~\ref{tab:benchmark}), it spans 34 categories across five physical domains: mechanics, optics, heat, electricity, and acoustics\footnote{Since generated videos are silent, the acoustics domain covers only the visible consequences of sound, such as water rippling beside a speaker.}. Every caption describes a complete causal event and states its outcome explicitly, so that semantic adherence requires rendering the cause, the action, and its physical consequence, \eg, ``the key switch is turned, and the lamp lights'' rather than ``the key switch is turned.''

Beyond PhyRubric-Bench, we scale up to \textbf{PhyRubric-Train}, 947 captions that share the taxonomy of the benchmark but no scenes, each paired with its own rubric. Here the rubric plays a different role: rather than defining the evaluation as in PhyRubric-Bench, it guides an agentic annotator, telling it which expectations to check and where to look, yielding 3,788 labeled videos from four T2V models with 45.9K rubric-item verdicts. On these labels, we train \textbf{PhyRubric-Judge}, an open-source VLM that needs no rubric at inference. Given only a caption and a video, it predicts semantic adherence and physical commonsense scores together with a diagnosis of which physical expectation is violated and why.

We evaluate ten T2V models on PhyRubric-Bench and find that basic physical commonsense remains far from solved: even the strongest model, Seedance 2.5, produces videos that are both faithful to the caption and physically plausible in fewer than half of cases. Moreover, PhyRubric-Judge correlates more closely with human judgments than the judges released with existing benchmarks, generalizes well to unseen generators and domains, and provides a diagnosis that locates each violation.

Our contributions are summarized as follows:
\begin{itemize}
\item We propose \textbf{PhyRubric}, a rubric-grounded framework that constructs a caption-specific rubric before any video is seen and derives every score from evidence-grounded item verdicts, so that each score comes with a diagnosis.
\item We construct \textbf{PhyRubric-Bench}, spanning 34 categories across five physical domains with human-verified rubrics and labels, and \textbf{PhyRubric-Train}, labeled by a rubric-guided agentic annotator, on which we train the open-source \textbf{PhyRubric-Judge} that generalizes to unseen generators and domains.
\item Rather than proposing a definitive standard, we show that our rubric-grounded evaluation serves as an analysis tool that yields deeper insights: semantic adherence is the main bottleneck, physical errors are all-or-nothing, and models often avoid physical errors by not depicting the event at all, a shortcut that prior protocols reward.
\end{itemize}

%% file: sec/2_related_work.tex
\section{Related Work}
\label{sec:related}

\subsection{Physical Commonsense Benchmarks for Text-to-Video Generation}
The rapid advancement of text-to-video (T2V) generation has motivated a range of benchmarks for its assessment~\cite{huang2024vbench,zheng2025vbench,huang2025vbench++,liu2024evalcrafter,wang2026very,sun2025t2v,liu2023fetv,liao2024evaluation}. Moving beyond general visual quality, a growing line of work focuses on physical commonsense in T2V generation~\cite{bansal2025videophy,bansal2026videophy,meng2024towards,gu2026phyworldbench,guo2025t2vphysbench,sanli2026physvidbench,km2026phyeduvideo}, while other works study physical understanding in settings where models are conditioned on observed frames and predict how the scene evolves~\cite{motamed2026generative,kang2024far,lin2026phyground}. However, most of these physical commonsense benchmarks are dominated by mechanics~\cite{bansal2025videophy,bansal2026videophy,guo2025t2vphysbench}, and some include prompts that require specialized knowledge beyond everyday experience~\cite{meng2024towards,gu2026phyworldbench,guo2025t2vphysbench,km2026phyeduvideo}. In contrast, PhyRubric-Bench spans five physical domains with prompts grounded in physical commonsense, and pairs each prompt with a caption-specific rubric that enables diagnostic evaluation.

\subsection{VLM-as-a-Judge for Physical Commonsense}

Existing evaluation of video generation mainly follows two paradigms: standard metrics such as FVD~\cite{unterthiner2018towards}, SSIM~\cite{wang2004image}, and CLIP score~\cite{hessel2021clipscore}, and VLM-based evaluation~\cite{chen2024mllm,pu2025judge,han2025video} such as VideoScore~\cite{he2024videoscore}. As standard metrics cannot capture physical plausibility, physical commonsense evaluation mainly adopts the latter, employing VLMs as dedicated judges. On the open-source side, judges such as VideoCon-Physics~\cite{bansal2025videophy} and VideoPhy-2-AutoEval~\cite{bansal2026videophy} are fine-tuned on the human annotations of their corresponding datasets. On the proprietary side, recent works prompt commercial models such as GPT-4o, Gemini-2.5-Pro, and GPT-5.5 as judges~\cite{meng2024towards,gu2026phyworldbench,sanli2026physvidbench,pothiraj2026physics}. Nevertheless, most of these judges output a holistic score rather than a detailed diagnosis, failing to reveal what goes wrong and why. 

\subsection{Rubric-Based Evaluation}

A rubric is an explicit set of criteria that decomposes a holistic quality judgment into verifiable items, making the assessment transparent, consistent, and interpretable~\cite{jonsson2007use,chen2026holistic}. Recently, rubrics have been widely adopted for fine-grained evaluation of large language models (LLMs) and vision-language models~\cite{kim2024prometheus,lee2024prometheus}, and increasingly tailor the criteria to each individual query rather than applying a generic rubric to all samples~\cite{kim2025biggen,cook2024ticking,arora2025healthbench}. In text-to-image generation, a related line of work decomposes prompts into fine-grained questions to assess text-image alignment~\cite{hu2023tifa,cho2024davidsonian}. However, rubric-based evaluation remains underexplored for physical commonsense in video generation. In this work, rubrics play two roles: they serve as the evaluation criteria of PhyRubric-Bench, and they guide an agentic annotator to label training data for PhyRubric-Judge.

%% file: sec/3_phyRubric.tex
\begin{figure*}[t]
\centering
\includegraphics[width=0.9\textwidth]{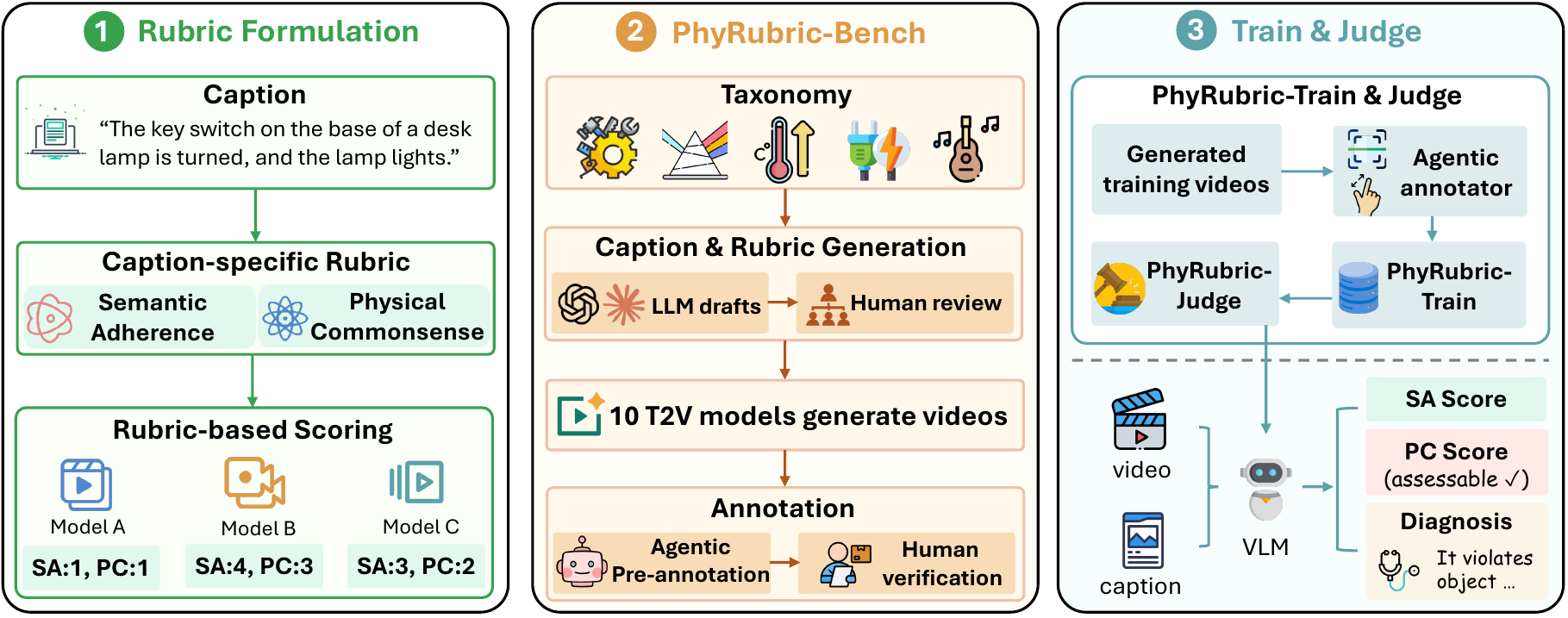}
\caption{\textbf{Overview of PhyRubric.} (1) \textbf{Rubric formulation}: each caption is paired with a rubric covering semantic adherence (SA) and physical commonsense (PC), constructed before any video is seen and reused to score videos from all models. (2) \textbf{PhyRubric-Bench}: captions spanning five physical domains and their rubrics are drafted by LLMs and reviewed by humans. Videos from 10 T2V models are pre-annotated by an agent and verified by humans. (3) \textbf{Train \& Judge}: an agentic annotator labels training videos to build PhyRubric-Train, on which we train PhyRubric-Judge. Given a video and its caption, the judge outputs an SA score, whether PC is assessable, a PC score, and a diagnosis of the violated expectation with timestamped evidence.}

\label{fig:overview}

\end{figure*}

\section{PhyRubric}
\label{sec:method}

As illustrated in Fig.~\ref{fig:overview}, PhyRubric fixes a rubric for each caption before any video is seen and shares it across all videos from that caption. This keeps the criteria identical across videos, so that scores from different generators are directly comparable and cannot be swayed by what a particular video happens to show. Moreover, a rubric reviewed by humans once can be reused for any number of videos, including those from future generators.

\subsection{Caption-Specific Rubric Formulation}
\label{sec:rubric}

\begin{figure}[t]
\centering
\includegraphics[width=\linewidth]{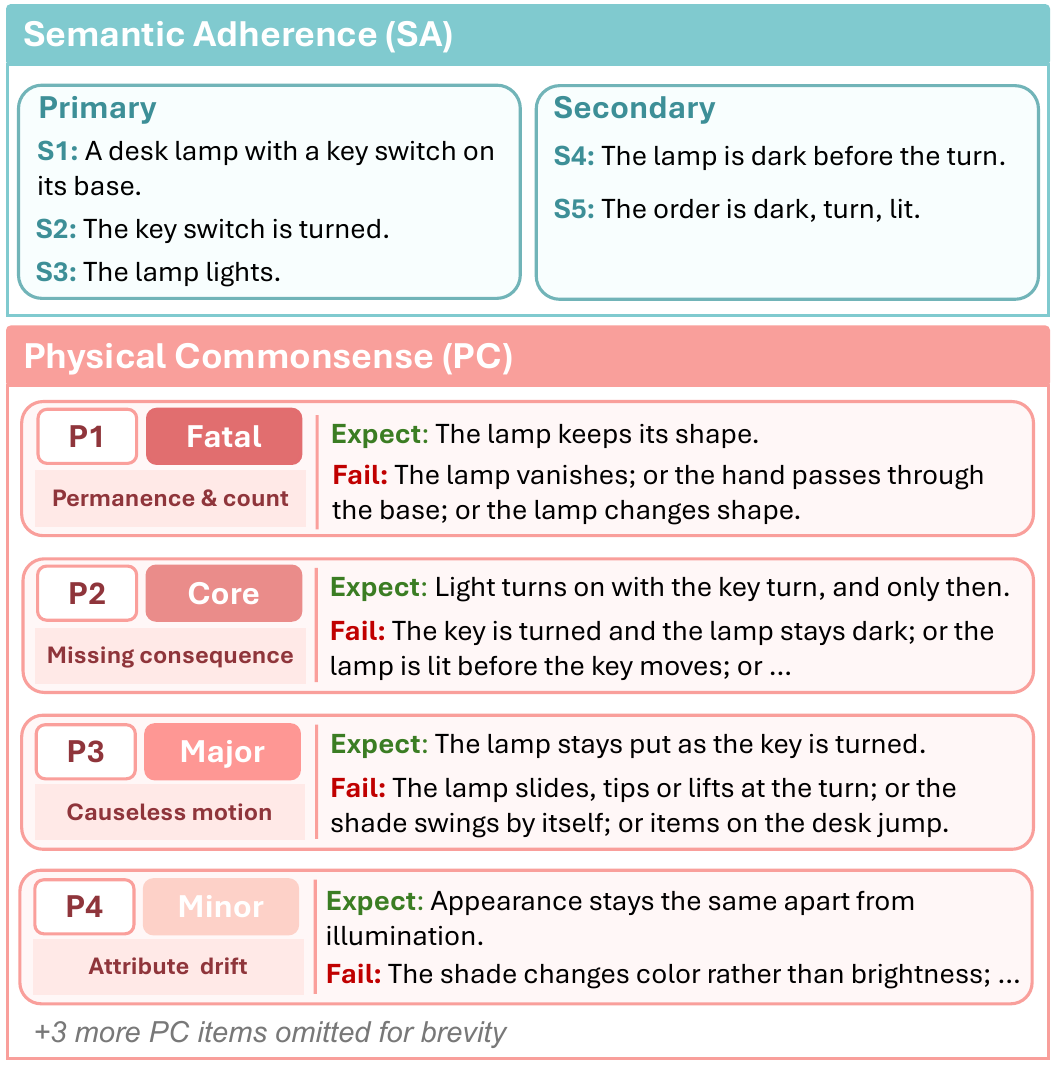}
\caption{A simplified rubric for the caption ``The key switch on the base of a desk lamp is turned, and the lamp lights.'' The full rubric is provided in Suppl.~Sec.~\ref{app:rubric_full}.}
\label{fig:rubric_example}
\vspace{-8pt}
\end{figure}

We evaluate generated videos along the two dimensions introduced by VideoPhy-2~\cite{bansal2026videophy}, semantic adherence (SA) and physical commonsense (PC), each on a 1--5 scale. Unlike prior benchmarks, where annotators assign a holistic score to each dimension, we pair every caption with a rubric that specifies what a video must satisfy to be faithful and physically plausible for that caption. The rubric consists of two groups of items, one for SA and one for PC, and each score is derived from item-level verdicts rather than rated directly. A simplified rubric example is shown in Fig.~\ref{fig:rubric_example}.

\noindent
\textbf{Semantic Adherence (SA).} SA measures whether the video shows what the caption describes. Its rubric items are the visible facts the caption requires, in two tiers:
\begin{itemize}
\item \textbf{Primary}: the main objects and each event in the caption.
\item \textbf{Secondary}: the remaining facts, including stated qualifiers (\eg, once, slowly, wooden), initial states implied by the caption, and the order of events.
\end{itemize}

\noindent
\textbf{Physical Commonsense (PC).} This metric evaluates whether the events shown in the video follow physical commonsense. We define four severity levels for physical violations, inspired by the order in which infants acquire physical knowledge in developmental psychology~\cite{spelke2007core,baillargeon2004infants,lin2022infants}:
\begin{itemize}
\item \textbf{Fatal}: the scene is impossible as a whole, violating core object principles such as permanence, solidity, or continuity (\eg, objects passing through each other, or appearing or vanishing without cause).
\item \textbf{Core}: the central interaction described by the caption does not happen or happens incorrectly (\eg, in reverse or through a broken mechanism).
\item \textbf{Major}: a clear violation outside the central interaction, such as a missing secondary consequence (\eg, the lamp lights but the desk beneath it stays dark).
\item \textbf{Minor}: a visual flaw that leaves the event intact, such as color or texture drift of untouched objects.
\end{itemize}
Each PC item is assigned one of these severity levels and one of ten violation types (\eg, permanence, solidity, causeless motion, process direction). Following the same developmental order, the ten types are grouped into L1 core object principles, L2 everyday dynamics, and L3 attribute experience (Suppl.~Tab.~\ref{tab:violation}). Each PC item is then judged by a pair of verifiable statements:
\begin{itemize}
\item \textbf{Expect}: what a physically correct video should show, specifying the moment, region, or event to inspect.
\item \textbf{Fail}: the concrete visible signs that count as a violation, and the item fails only if one of these signs is observed.
\end{itemize}

\noindent
\textbf{Verdicts.} An SA item is judged \emph{pass} or \emph{fail}. A PC item has a third verdict, \emph{not shown}, which applies when the moment, region, or event specified by its Expect statement never appears in the video. Otherwise, the item fails if any sign in its Fail statement is observed, and passes if not. Note that \emph{not shown} means the situation is absent, not that the expected behavior failed to occur. For example, for the caption ``The key switch on the base of a desk lamp is turned, and the lamp lights'' (Fig.~\ref{fig:rubric_example}), if the key is never turned in the video, P2 is \emph{not shown} rather than failed, since the missing action is already penalized by the SA item S2. Since a fail can only be triggered by a listed sign, every negative verdict is grounded in a specific observation rather than an overall impression, and carries the severity and violation type of its item. Consistent with the nature of commonsense, PC items judge only the direction and order of change rather than its rate or extent, and never penalize a cause for being out of view (\eg, a heat source or power supply).

\noindent
\textbf{Scoring Rules.} Both scores lie on a 1--5 scale and are derived from the verdicts by fixed rules. For SA, a video judged entirely unrelated to the caption scores 1. Otherwise, it scores 5 if all items pass, 4 if only secondary items fail, 3 if exactly one primary item fails, and 2 if two or more primary items fail. For PC, let $F$ be the set of failed items. A video with $F=\emptyset$ scores 5; otherwise,
\begin{equation}
    \mathrm{PC} = \max\big(1,\; \min_{i \in F} c(s_i) - \max(0,\, h - 1)\big),
\end{equation}
where $s_i$ is the severity of item $i$, $c(\cdot)$ maps fatal, core, major, and minor to caps of 1, 2, 3, and 4, and $h$ is the number of failed items of major severity or above. In words, the most severe failure sets the cap, and each additional failure of major severity or above lowers the score by one. Each rubric is checked to contain at least two primary and one secondary SA items and at least one PC item of every severity, so that every score from 1 to 5 is attainable.

\subsection{PhyRubric-Bench}
\label{sec:bench}

\begin{figure*}[t]
\centering
\includegraphics[width=0.86\textwidth]{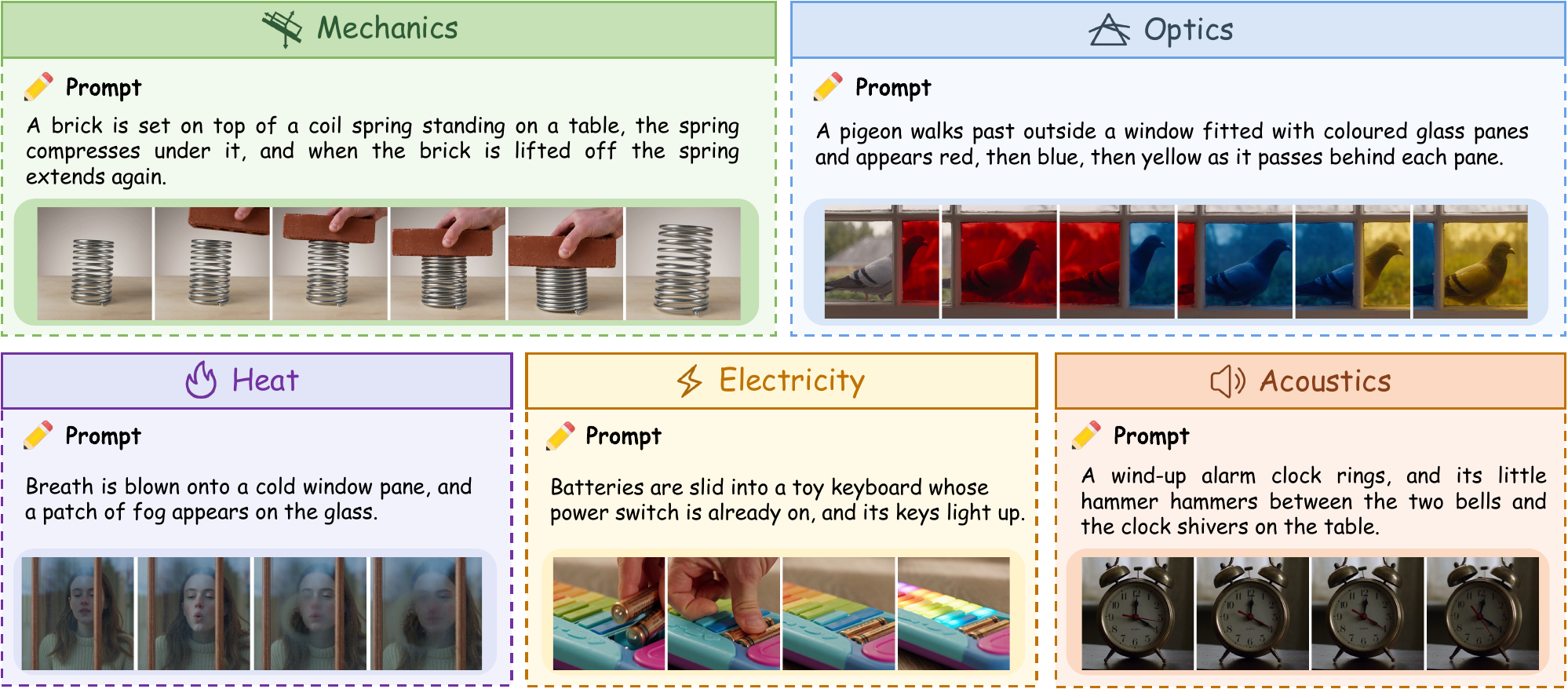}
\caption{Representative examples from PhyRubric-Bench. Each caption describes a single everyday event and states its visible outcome.}
\label{fig:domains}
\vspace{-6pt}
\end{figure*}

Suppl.~Tab.~\ref{tab:benchmark} compares PhyRubric-Bench with prior benchmarks, which are dominated by mechanics. PhyRubric-Bench covers all five physical domains with 34 categories (Fig.~\ref{fig:domains}), and every caption tests only physical commonsense and is paired with a human-verified rubric.

\noindent
\textbf{Caption and rubric generation.} Captions are drafted by Claude Fable 5.1~\cite{anthropic2026fable} and GPT-5.6 Sol~\cite{openai2026gpt56} and refined by human reviewers until each describes a single causal event that is visible within about five seconds and has an explicitly stated outcome. Stating the outcome serves two purposes. (1) \textit{Fixed expectations}: what should happen is determined before any video is seen, so the rubric can specify the expected events and violations rather than guess among plausible outcomes. (2) \textit{Diagnostic failures}: if a model is told what should happen yet fails to render it, the failure reflects missing physical knowledge rather than creative freedom.

In addition, captions include a person only when the human body itself is part of the physics, since current generators often produce distorted human bodies~\cite{zheng2025vbench,zhang2025human,ma2025evaluating}. Each phenomenon is instantiated in three to five distinct scenes for diversity, yielding 1,192 captions, which are split into 245 for PhyRubric-Bench and 947 for PhyRubric-Train (Sec.~\ref{sec:judge}) with no scene shared between the two. Each caption is then paired with a rubric, drafted by an LLM prompted with example rubrics from the same domain, checked automatically for format and score reachability, and revised by human reviewers until approved.

\noindent
\textbf{Video generation and annotation.} For evaluation, the 245 benchmark captions are rendered by ten text-to-video models, six of which are unseen during judge training, yielding 2,433 videos\footnote{The remaining 17 of the $10\times245$ videos were not produced because the generators' built-in safety filters declined the captions.}. Each video is first pre-annotated item by item by our agentic annotator (Sec.~\ref{sec:judge}), and human annotators then verify every verdict against the video to obtain the ground truth. Since the rubric already specifies what counts as a violation, annotators only need to check whether the listed signs appear, rather than judging physical plausibility themselves. More construction details and statistics are given in Suppl.~Sec.~\ref{sup:stats}.

\subsection{PhyRubric-Train and PhyRubric-Judge}
\label{sec:judge}

\noindent
\textbf{Training data annotation.} Human verification does not scale to thousands of videos, so we label the training set with the same agentic annotator used to pre-annotate PhyRubric-Bench, built on Claude Sonnet 5.5~\cite{anthropic2026sonnet}. Instead of a fixed set of frames, the annotator first scans a timestamped overview of the clip and then zooms in at full frame rate where and when it needs to, until each rubric item can be decided from frames that clearly show the relevant event. This matters because physical violations often involve small objects or brief moments that uniform sampling misses. For every item, it returns a verdict with timestamped evidence, and the scores follow from the verdicts by the rules in Sec.~\ref{sec:rubric}. Applied to the 947 training captions, each rendered by four generators, it yields PhyRubric-Train with 3,788 videos and 45.9K item-level verdicts. Suppl.~Sec.~\ref{sup:annotator} details its protocol and reliability.

\noindent
\textbf{Judge.} The agentic annotator produces reliable labels but is impractical as a general-purpose evaluator: it relies on a proprietary model, makes repeated tool calls per video, and needs a rubric for every new caption. We therefore train PhyRubric-Judge, an open-source VLM, on PhyRubric-Train. Starting from Qwen3-VL-8B~\cite{bai2025qwen3}, we explore configurations that vary the rubric input, the output format, and the training objective (SFT or GRPO~\cite{shao2024deepseekmath}). Our default judge takes only the caption and uniformly sampled frames, so that it applies to any new caption and receives the same input as existing judges. \textit{It outputs the SA score, whether PC is assessable, the PC score, and a diagnosis.} The training targets come from PhyRubric-Train, with the diagnosis assembled from the item-level evidence: a summary of what the video shows, the missing caption facts, and each physical violation with its evidence. The judge thus learns not only what score to give but why (Suppl.~Sec.~\ref{sup:suppl_ablation}).

%% file: sec/4_experiment.tex
\section{Experiments}

\input{Tab/leaderboard}

\subsection{Experimental Setup}
\label{sec:exp_setup}

\noindent\textbf{Video Generative Models.} PhyRubric-Bench is rendered by ten video generators. Four of them are open-source T2V models that also render PhyRubric-Train: CogVideoX1.5-5B~\cite{yang2025cogvideox}, HunyuanVideo-1.5-8B~\cite{wu2025hunyuanvideo}, LTX-Video-2B~\cite{hacohen2024ltx}, and Wan2.2-T2V-A14B~\cite{wan2025wan}. The other six are unseen during training and measure how well a judge generalizes to new generators: Wan2.2-TI2V-5B~\cite{wan2025wan}, LTX-2.3-22B~\cite{hacohen2026ltx}, Cosmos-Predict2.5-2B and Cosmos-Predict2.5-14B~\cite{agarwal2025cosmos}, and the proprietary Seedance 2.5~\cite{seedance25} and Veo 3.1~\cite{google2025veo31}. All videos are five to six seconds long.

\noindent\textbf{Evaluation Metrics.} Each video is rated on a 1--5 scale for SA and PC. We report the mean SA over all videos and the mean PC over assessable videos only. We also report the joint performance (JP), the percentage of all videos with SA$\geq$4 and PC$\geq$4, \ie, videos that are both faithful to the caption and physically plausible. Following VideoPhy-2~\cite{bansal2026videophy}, we use JP as the primary metric. A high PC alone only indicates that what is shown is plausible (\eg, a nearly static video can easily appear plausible), so SA is needed to confirm that the model produces the described event.

\noindent\textbf{Implementation Details.} For video generation, each open-source model is run with its official checkpoint and default sampling recipe, generating one video per caption with a fixed seed of 42. HunyuanVideo-1.5 and Wan2.2-T2V-A14B are rendered at 480p, CogVideoX1.5-5B at 768p, and the remaining open-source models at 704p. Seedance 2.5 and Veo 3.1 are accessed through their official APIs at 480p and 720p, respectively. Per-model settings are listed in Suppl.~Sec.~\ref{sup:generation}. For PhyRubric-Judge, we fine-tune Qwen3-VL-8B-Instruct~\cite{bai2025qwen3} on PhyRubric-Train with LoRA (rank 16)~\cite{hu2021lora} on all linear layers of the language model, keeping the vision encoder and projector frozen. The judge takes the caption and 16 uniformly sampled frames, each resized to at most $448\times448$. We train for 2 epochs (900 steps) with an effective batch size of 8 and a cosine learning rate schedule peaking at $10^{-4}$. At inference, the judge decodes greedily, and we additionally compute a soft score~\cite{liu2023g,wu2023q} as the expected rating under the probabilities of the five score tokens (Suppl.~Sec.~\ref{sup:suppl_metrics}). All experiments are run on NVIDIA GH200 Superchips.

\begin{figure*}[t]
\centering
\begin{subfigure}[b]{0.298\textwidth}
  \centering
  \includegraphics[height=1.71in]{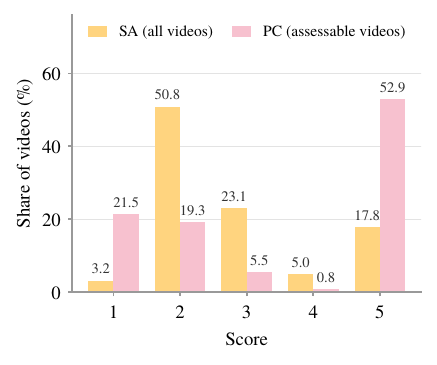}
  \captionsetup{margin={13pt,0pt}}
  \caption{Distribution of SA and PC scores}
  \label{fig:score_dist}
\end{subfigure}\hfill
\begin{subfigure}[b]{0.320\textwidth}
  \centering
  \includegraphics[height=1.71in]{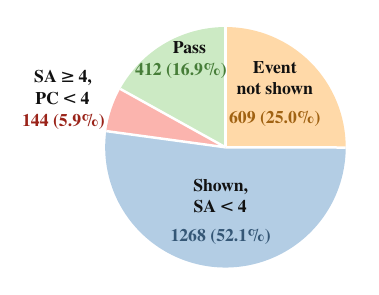}
  \captionsetup{margin={28pt,0pt}}
  \caption{Outcome of all videos}
  \label{fig:outcome_pie}
\end{subfigure}\hfill
\begin{subfigure}[b]{0.349\textwidth}
  \centering
  \includegraphics[height=1.71in]{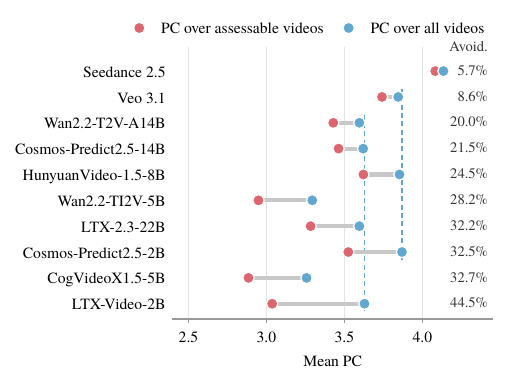}
  \captionsetup{margin={45pt,0pt}}
  \caption{PC over assessable vs.\ all videos}
  \label{fig:pc_inflation}
\end{subfigure}
\caption{\textbf{Statistics of PhyRubric-Bench.} (a) SA over all videos and PC over assessable videos. (b) Outcome of each video: the event is not shown, shown incompletely (SA$<$4), shown but physically implausible (SA$\geq$4, PC$<$4), or passing (SA$\geq$4, PC$\geq$4). (c) Mean PC per generator over assessable and all videos, sorted by avoidance rate (right column). Dashed lines mark two generators for reference.}

\label{fig:bench_analysis}
\vspace{-3pt} 
\end{figure*}

\subsection{Main Results}
\label{sec:main_results}

Tab.~\ref{tab:leaderboard} reports the results of ten video generators on PhyRubric-Bench across five physics domains. We summarize the main findings below.

\noindent\textbf{Physical commonsense remains far from solved.} The two proprietary models lead by a wide margin: Seedance 2.5 achieves an average JP of 44.20\% and Veo 3.1 36.20\%, two to three times that of the strongest open-source model, HunyuanVideo-1.5-8B (14.93\%). Even so, more than half of their videos are still not both faithful to the caption and physically plausible.
Moreover, every open-source model has an average SA below 3, indicating that at least one primary element of the caption is typically missing from the generated video. Even among assessable videos, where the intended event does occur, their average PC never exceeds 3.5, suggesting that the event typically violates at least one expectation of major severity or above.

\noindent\textbf{Video generative models handle optics best but struggle with heat and acoustics.}
As shown in Tab.~\ref{tab:leaderboard}, Optics is the strongest domain for eight of the ten evaluated models, and ranks second (or tied for second) for the remaining two models, Seedance 2.5 and CogVideoX1.5-5B. In contrast, Heat and Acoustics are the weakest domains for most models. Even the proprietary models reach only 30.00\% (Seedance 2.5) and 15.00\% (Veo 3.1) JP on Acoustics, and two open-source models score 0.00\% in this domain. We conjecture that optical phenomena (\eg, reflection and shadows) are visually salient and well represented in training data, whereas thermal processes involve gradual state changes and acoustic phenomena must be conveyed through indirect visual cues (\eg, sound-induced ripples on a liquid surface), both of which are harder to render.

\noindent\textbf{Semantic adherence is the main bottleneck, and physical errors are all-or-nothing.}
Fig.~\ref{fig:score_dist} shows the distribution of SA and PC scores. SA concentrates at 2 (50.8\% of all videos), the score given when two or more primary elements of the caption are missing, while only 5.0\% of videos receive a 4, the score for a video that misses only secondary details: most generated videos fail to depict the captioned event itself, before physics even comes into play. In contrast, once the event is fully shown, its physics is usually correct: as shown in Fig.~\ref{fig:outcome_pie}, only 5.9\% of videos show the event fully but violate physics (SA$\geq$4, PC$<$4), compared with 16.9\% that pass, meaning that 74.1\% of fully shown events are rendered with plausible physics. Physical errors themselves are all-or-nothing: PC over assessable videos is bimodal, with 52.9\% scoring 5, \ie, no violation at all, and 40.8\% scoring 1 or 2, \ie, a fatal or core violation, whereas scores of 3 and 4, \ie, only major or minor violations, together account for 6.3\%. This suggests that when current models do show the event, they either render its physics correctly or break it at the most basic level, and rarely get the core event right but slip on peripheral details.

\noindent\textbf{Models take a shortcut to high PC by not showing the event.}
Prior protocols rate PC on every video, so a video that never attempts the event leaves nothing for physics to contradict and receives a high PC. In PhyRubric, the PC items on the central event of such a video are judged \emph{not shown}, so the video is marked unassessable and excluded from PC. We define the avoidance rate of a model as the share of its videos that never show the event. Such avoidance is common: 25.0\% of all videos never show the event (Fig.~\ref{fig:outcome_pie}), and the avoidance rate ranges from 20.0\% to 44.5\% across open-source models, compared with 5.7\% and 8.6\% for Seedance 2.5 and Veo 3.1 (Fig.~\ref{fig:pc_inflation}). We regard avoidance as a failure of physical commonsense rather than merely of instruction following: every caption in PhyRubric-Bench states the outcome explicitly, so a model that still renders a static scene or a different event suggests that it does not know how the stated cause produces the stated effect. If avoided videos are scored as in prior protocols, the PC score of every open-source model rises by 0.16--0.59, and the gain grows with its avoidance rate (Fig.~\ref{fig:pc_inflation}). LTX-Video-2B, with the highest avoidance rate (44.5\%), would then rank above Cosmos-Predict2.5-14B and Wan2.2-T2V-A14B. PC alone is most misleading for Cosmos-Predict2.5-2B: under prior protocols its PC would be second only to Seedance 2.5 and above Veo 3.1, yet its JP (6.69\%) ranks eighth of ten, as it rarely depicts the event.

\subsection{Evaluation of PhyRubric-Judge}
\label{sec:judge_eval}

We assess PhyRubric-Judge by its agreement with the human labels of PhyRubric-Bench. We apply each judge to every video in PhyRubric-Bench and report two metrics on its soft score: Pearson correlation ($r$)~\cite{pearson1895note} with the human scores, and ROC-AUC~\cite{hanley1982meaning} for distinguishing videos that humans rate as pass (score$\geq$4) from those rated as fail, separately for SA and PC. Suppl.~Sec.~\ref{sup:suppl_metrics} reports the results of all judges on the full set of metrics adopted by prior work.

\noindent\textbf{Baselines.} We compare PhyRubric-Judge with four groups of judges. (1) \emph{Zero-shot VLMs}: Qwen3-VL-8B, Qwen3-VL-32B, and Qwen3-VL-235B-A22B~\cite{bai2025qwen3}. The 8B model is also the untrained starting point of our judge. (2) \emph{General video evaluators}: VideoScore~\cite{he2024videoscore} and VideoScore2~\cite{he2025videoscore2}, fine-tuned to rate generated videos along general quality dimensions. (3) \emph{Physics-specific evaluators}: VideoCon-Physics~\cite{bansal2025videophy} and VideoPhy-2-AutoEval~\cite{bansal2026videophy}, each fine-tuned on human annotations of its own benchmark. (4) \emph{Proprietary VLM}: Gemini 4 Argon~\cite{google2026gemini4argon}, a frontier closed-source model, prompted with the same instructions as our judge. Details are in the Suppl.~Sec.~\ref{sup:suppl_baselines}.

\input{Tab/judge}

\noindent\textbf{PhyRubric-Judge aligns best with human judgments.} As shown in Tab.~\ref{tab:judge}, PhyRubric-Judge achieves the best agreement with human judgments on both SA and PC across all metrics. On SA, it reaches a correlation of 0.61, well above the best open-source baseline, Qwen3-VL-32B (0.49). PC is the harder dimension: open-source zero-shot VLMs perform close to chance (AUC 0.53--0.58), and even the strongest open-source evaluator, VideoPhy-2-AutoEval, reaches only $r=0.23$, despite being fine-tuned in two stages on 62K human annotations from VideoPhy and VideoPhy-2. In contrast, PhyRubric-Judge reaches 0.32 while being trained on only 3,600 automatically labeled videos. Notably, this 8B model is on par with the proprietary Gemini 4 Argon ($r=0.60$ on SA and $0.31$ on PC), showing that rubric-grounded supervision enables a small open model to match a frontier proprietary model.

\input{Tab/generalization}

\noindent\textbf{PhyRubric-Judge generalizes to unseen generators and domains.} PhyRubric-Train contains videos from only four of the ten generators. To test whether the judge learns physical commonsense rather than the artifacts of these four generators, Tab.~\ref{tab:generalization} (a) compares its agreement on benchmark videos from these generators (Seen) and from the other six (Unseen). On unseen generators, PhyRubric-Judge loses only 0.03 in SA and 0.06 in PC correlation, and still outperforms every baseline. To test generalization across domains, Tab.~\ref{tab:generalization} (b) trains five additional judges, each with one domain removed from PhyRubric-Train, and evaluates each on its held-out domain. SA correlation drops by at most 0.04, and PC correlation by at most 0.09 (on mechanics).

\subsection{Error Analysis}
\label{sec:error_analysis}

\begin{figure}[t]
\centering
\includegraphics[width=0.93\linewidth]{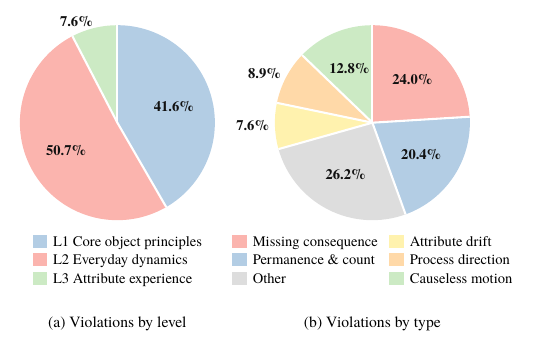}
\caption{Distribution of PC violations in the assessable videos of PhyRubric-Bench, according to the human labels. Each violation type is counted once per video. (a) By level of the violation taxonomy. (b) By violation type, showing the five most frequent types and the rest as \emph{Other}.}
\label{fig:error_pies}
\vspace{-8pt}
\end{figure}

\begin{figure}[t]
\centering
\includegraphics[width=\linewidth]{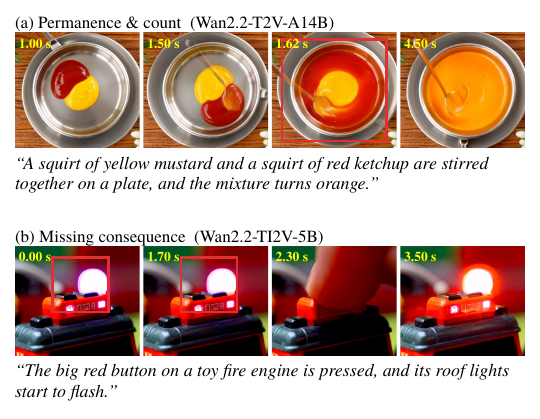}
\caption{Two failure cases from PhyRubric-Bench. (a) A small squirt of ketchup becomes a full bowl and the spoon jumps between 1.50\,s and 1.62\,s. (b) The roof light is lit from the first frame, before the finger reaches the button at 2.30\,s.}
\label{fig:error_case}
\vspace{-8pt}
\end{figure}

To better understand the limitations of current T2V models, we categorize their failed items by the three levels and ten violation types in PhyRubric, with more analysis in Suppl.~Sec.~\ref{sup:error_analysis}.

\noindent
\textbf{The physical commonsense of current T2V models is still limited at the most basic level.} As shown in Fig.~\ref{fig:error_pies} (a), 41.6\% of the violations belong to L1 (core object principles) and 50.7\% to L2 (everyday dynamics), while only 7.6\% belong to L3 (attribute experience). This indicates that current models rarely fail on fine attributes, but instead break principles that hold for any object, such as persistence over time, and the way common events unfold. Fig.~\ref{fig:error_pies} (b) shows that the two most frequent types, missing consequence (24.0\%) and permanence \& count (20.4\%), together account for nearly half of all violations. A permanence \& count violation occurs when an object appears from nothing, vanishes, teleports, duplicates, or merges with another during the video. For example, in Fig.~\ref{fig:error_case} (a), the sauce becomes fully stirred and the spoon teleports to a new position between 1.50\,s and 1.62\,s, a severe violation that any viewer would notice at a glance. A missing consequence violation occurs when the trigger of an event is shown but its effect never follows, or occurs at the wrong time. In Fig.~\ref{fig:error_case}(b), the lights are on from the start, before the finger even reaches the button. Both videos have high visual quality, yet their obvious physical errors make them distinguishable from real videos.

%% file: Tab/leaderboard.tex
\begin{table*}[t]
\centering
\caption{\textbf{Physical commonsense of T2V models on PhyRubric-Bench.} SA and PC are mean scores on a 1--5 scale; PC is computed over assessable videos only. JP is the joint performance (\%), the fraction of all videos with SA$\geq$4 and PC$\geq$4. Avg averages over domains. The best proprietary model is in \textbf{bold} and the best open-source model is \underline{underlined}. For each model, \colorbox{goodat}{pink} and \colorbox{worseat}{blue} shading mark the JP of its best and worst domains, respectively.}
\label{tab:leaderboard}
\resizebox{\textwidth}{!}{%
\begin{tabular}{l|ccc|ccc|ccc|ccc|ccc|ccc}
\toprule
 & \multicolumn{3}{c|}{\textbf{Mechanics}} & \multicolumn{3}{c|}{\textbf{Optics}} & \multicolumn{3}{c|}{\textbf{Heat}} & \multicolumn{3}{c|}{\textbf{Electricity}} & \multicolumn{3}{c|}{\textbf{Acoustics}} & \multicolumn{3}{c}{\textbf{Avg}} \\
\cmidrule(lr){2-4} \cmidrule(lr){5-7} \cmidrule(lr){8-10} \cmidrule(lr){11-13} \cmidrule(lr){14-16} \cmidrule(lr){17-19}
\textbf{Model} & SA & PC & JP & SA & PC & JP & SA & PC & JP & SA & PC & JP & SA & PC & JP & SA & PC & JP \\
\midrule
\multicolumn{19}{c}{Proprietary Models} \\
\midrule
Seedance 2.5~\cite{seedance25} & \textbf{3.95} & \textbf{4.24} & \textbf{48.00} & \textbf{3.96} & \textbf{3.96} & \textbf{48.98} & 3.42 & \textbf{4.14} & \textbf{38.00} & \textbf{3.94} & \textbf{4.13} & \good{\textbf{56.00}} & \textbf{3.60} & \textbf{3.47} & \bad{\textbf{30.00}} & \textbf{3.77} & \textbf{3.99} & \textbf{44.20} \\
Veo 3.1~\cite{google2025veo31} & 3.88 & 4.06 & 44.00 & 3.78 & 3.89 & \good{48.00} & \textbf{3.48} & 3.92 & 36.00 & 3.80 & 3.35 & 38.00 & 3.45 & 2.65 & \bad{15.00} & 3.68 & 3.57 & 36.20 \\
\midrule
\multicolumn{19}{c}{Open-Source Models} \\
\midrule
LTX-Video-2B~\cite{hacohen2024ltx} & 2.28 & 2.59 & 5.33 & 2.56 & 2.94 & \good{6.00} & 2.20 & 3.73 & \bad{2.00} & 2.30 & \underline{3.42} & \bad{2.00} & 2.00 & 2.50 & 5.00 & 2.27 & 3.04 & 4.07 \\
Cosmos-Predict2.5-2B~\cite{agarwal2025cosmos} & 2.29 & 3.16 & 5.56 & 2.67 & \underline{4.36} & \good{12.50} & 2.32 & 3.86 & \bad{4.00} & 2.38 & 3.22 & 6.38 & 2.60 & 2.23 & 5.00 & 2.45 & 3.37 & 6.69 \\
Wan2.2-TI2V-5B~\cite{wan2025wan} & 2.73 & 2.72 & 9.33 & 2.88 & 3.41 & \good{16.00} & 2.44 & 3.29 & 10.00 & 2.64 & 2.83 & 4.00 & 2.45 & 2.00 & \bad{0.00} & 2.63 & 2.85 & 7.87 \\
CogVideoX1.5-5B~\cite{yang2025cogvideox} & 2.21 & 2.52 & \bad{1.33} & 2.64 & 2.63 & 6.00 & 2.34 & 3.78 & \good{8.00} & 2.34 & 2.76 & 6.00 & 2.50 & 2.71 & 5.00 & 2.41 & 2.88 & 5.27 \\
HunyuanVideo-1.5-8B~\cite{wu2025hunyuanvideo} & 2.83 & \underline{3.93} & 14.67 & 3.20 & 3.87 & \good{\underline{28.00}} & \underline{2.64} & \underline{3.93} & \underline{18.00} & 2.42 & 2.70 & \bad{4.00} & 2.40 & \underline{2.88} & \underline{10.00} & 2.70 & \underline{3.46} & \underline{14.93} \\
Wan2.2-T2V-A14B~\cite{wan2025wan} & 2.95 & 3.43 & \underline{21.33} & \underline{3.44} & 3.75 & \good{26.00} & 2.26 & 3.53 & \bad{2.00} & \underline{2.96} & 3.29 & \underline{20.00} & 2.60 & 2.71 & 5.00 & \underline{2.84} & 3.34 & 14.87 \\
Cosmos-Predict2.5-14B~\cite{agarwal2025cosmos} & 2.69 & 3.68 & 13.89 & 2.96 & 3.58 & \good{14.58} & 2.46 & 3.69 & 12.00 & 2.62 & 3.15 & 10.64 & \underline{2.65} & 2.29 & \bad{\underline{10.00}} & 2.68 & 3.28 & 12.22 \\
LTX-2.3-22B~\cite{hacohen2026ltx} & \underline{2.96} & 3.31 & 17.33 & 2.96 & 3.62 & \good{18.00} & 2.48 & 3.15 & 8.00 & 2.50 & 3.20 & 14.00 & 2.45 & 2.82 & \bad{0.00} & 2.67 & 3.22 & 11.47 \\
\bottomrule
\end{tabular}%
}
\vspace{-8pt}
\end{table*}

%% file: Tab/judge.tex
\begin{table}[t]
\centering
\caption{Agreement between judges and human labels on PhyRubric-Bench. $r$: Pearson correlation with human scores; AUC: ROC-AUC for distinguishing videos rated as pass (score$\geq$4) from fail by humans. Both are computed on each judge's soft score. Best results in \textbf{bold}, second best \underline{underlined}.}
\label{tab:judge}
\resizebox{0.85\linewidth}{!}{%
\begin{tabular}{l|cc|cc}
\toprule
 & \multicolumn{2}{c|}{\textbf{SA}} & \multicolumn{2}{c}{\textbf{PC}} \\
\cmidrule(lr){2-3} \cmidrule(lr){4-5}
\textbf{Judge} & $r\uparrow$ & AUC$\uparrow$ & $r\uparrow$ & AUC$\uparrow$ \\
\midrule
\multicolumn{5}{l}{\textit{Zero-Shot VLMs}} \\
Qwen3-VL-8B~\cite{bai2025qwen3}   & 0.46 & 0.78 & 0.02 & 0.53 \\
Qwen3-VL-32B~\cite{bai2025qwen3}  & 0.49 & 0.81 & 0.14 & 0.58 \\
Qwen3-VL-235B~\cite{bai2025qwen3} & 0.46 & 0.81 & 0.14 & 0.57 \\
\midrule
\multicolumn{5}{l}{\textit{General Video Evaluators}} \\
VideoScore~\cite{he2024videoscore}   & $-$0.01 & 0.48 & 0.11 & 0.59 \\
VideoScore2~\cite{he2025videoscore2} & 0.32 & 0.68 & 0.11 & 0.56 \\
\midrule
\multicolumn{5}{l}{\textit{Physics-Specific Evaluators}} \\
VideoCon-Physics~\cite{bansal2025videophy}     & 0.16 & 0.59 & 0.18 & 0.61 \\
VideoPhy-2-AutoEval~\cite{bansal2026videophy}  & 0.21 & 0.63 & 0.23 & 0.63 \\
\midrule
\multicolumn{5}{l}{\textit{Proprietary VLM}} \\
Gemini 4 Argon~\cite{google2026gemini4argon} & \underline{0.60} & \textbf{0.84} & \underline{0.31} & \textbf{0.68} \\
\midrule
\textbf{PhyRubric-Judge (Ours)} & \textbf{0.61} & \textbf{0.84} & \textbf{0.32} & \textbf{0.68} \\
\bottomrule
\end{tabular}%
}
\vspace{-6pt}
\end{table}

%% file: Tab/generalization.tex
\begin{table}[t]
\centering
\caption{Generalization of PhyRubric-Judge. (a) Videos from the four generators used to build PhyRubric-Train (Seen, 980 videos) and from the six held-out generators (Unseen, 1,453 videos). (b) Per-domain agreement of the judge trained on all five domains (Full) and with that domain held out (w/o). Metrics as in Tab.~\ref{tab:judge}.}
\label{tab:generalization}
\resizebox{\linewidth}{!}{%
\begin{tabular}{l|cccc|cccc}
\toprule
\multicolumn{9}{l}{\textit{(a) Seen vs.\ unseen generators}} \\
 & \multicolumn{4}{c|}{\textbf{Seen}} & \multicolumn{4}{c}{\textbf{Unseen}} \\
 & \multicolumn{2}{c}{SA} & \multicolumn{2}{c|}{PC} & \multicolumn{2}{c}{SA} & \multicolumn{2}{c}{PC} \\
\cmidrule(lr){2-3} \cmidrule(lr){4-5} \cmidrule(lr){6-7} \cmidrule(lr){8-9}
\textbf{Judge} & $r$ & AUC & $r$ & AUC & $r$ & AUC & $r$ & AUC \\
\midrule
Qwen3-VL-32B        & 0.48 & 0.82 & 0.18 & 0.60 & 0.49 & 0.80 & 0.11 & 0.57 \\
VideoScore2         & 0.32 & 0.69 & 0.12 & 0.55 & 0.30 & 0.67 & 0.10 & 0.56 \\
VideoPhy-2-AutoEval & 0.25 & 0.66 & 0.24 & 0.64 & 0.18 & 0.62 & 0.23 & 0.63 \\
\textbf{PhyRubric-Judge (Ours)} & \textbf{0.62} & \textbf{0.86} & \textbf{0.35} & \textbf{0.69} & \textbf{0.59} & \textbf{0.83} & \textbf{0.29} & \textbf{0.67} \\
\midrule
\multicolumn{9}{l}{\textit{(b) Held-out physical domain}} \\
 & \multicolumn{4}{c|}{\textbf{Full}} & \multicolumn{4}{c}{\textbf{w/o the domain}} \\
 & \multicolumn{2}{c}{SA} & \multicolumn{2}{c|}{PC} & \multicolumn{2}{c}{SA} & \multicolumn{2}{c}{PC} \\
\cmidrule(lr){2-3} \cmidrule(lr){4-5} \cmidrule(lr){6-7} \cmidrule(lr){8-9}
\textbf{Domain} & $r$ & AUC & $r$ & AUC & $r$ & AUC & $r$ & AUC \\
\midrule
Mechanics   & 0.67 & 0.88 & 0.37 & 0.72 & 0.65 & 0.87 & 0.28 & 0.67 \\
Optics      & 0.57 & 0.81 & 0.30 & 0.66 & 0.53 & 0.79 & 0.24 & 0.63 \\
Heat        & 0.58 & 0.86 & 0.24 & 0.66 & 0.54 & 0.85 & 0.21 & 0.64 \\
Electricity & 0.59 & 0.83 & 0.34 & 0.70 & 0.56 & 0.82 & 0.30 & 0.69 \\
Acoustics   & 0.54 & 0.81 & 0.19 & 0.58 & 0.52 & 0.79 & 0.23 & 0.60 \\
\bottomrule
\end{tabular}%
}
\vspace{-10pt}
\end{table}

%% file: sec/5_conclusion.tex
\section{Conclusion}
\label{sec:conclusion}

We presented PhyRubric, a rubric-grounded framework for diagnostic evaluation of physical commonsense in T2V generation, which fixes a caption-specific rubric before any video is seen and derives every score from evidence-grounded item verdicts. We instantiated it as PhyRubric-Bench across five physical domains and scaled it up to PhyRubric-Train, on which we trained PhyRubric-Judge, an open-source judge that outperforms existing open-source judges, matches Gemini 4 Argon, and generalizes to unseen generators and domains. Our evaluation of T2V models reveals that physical commonsense remains far from solved, and our in-depth analysis provides actionable insights for developing future T2V models.

%% file: sec/6_llm_statement.tex
\section*{LLM Usage Statement}

Beyond the LLMs that are described in the paper, Claude models were used to assist with code implementation and to polish the writing. All LLM-generated code was reviewed by the authors.

%% file: sec/ack.tex
\section*{Acknowledgments}
This work used the Delta and DeltaAI systems at the National Center for Supercomputing Applications (NCSA) through allocation CIS261732 from the Advanced Cyberinfrastructure Coordination Ecosystem: Services \& Support (ACCESS) program~\cite{boerner2023access}, which is supported by U.S. National Science Foundation (NSF) grants \#2138259, \#2138286, \#2138307, \#2137603, and \#2138296. Delta and DeltaAI~\cite{bode2025deltaai} are supported by NSF awards OAC 2005572 and OAC 2320345, respectively, and the State of Illinois. This work also used Google Cloud through the CloudBank project, which is supported by NSF grant \#1925001.

%% file: sec/X_suppl.tex
\clearpage
\setcounter{page}{1}
\maketitlesupplementary
\appendix

\section{Dataset and Benchmark}

\input{Tab/benchmark}

\subsection{Comparison with Prior Benchmarks}
\label{sup:comparison}

Tab.~\ref{tab:benchmark} compares PhyRubric with prior benchmarks that evaluate physical commonsense in text-to-video models. Each cell is determined from the benchmark's paper and released code and data, following the criteria below.

\noindent\textbf{Domains.} Prior benchmarks center on mechanics, such as collisions, falling, and fluid flow. Heat is a main category only in PhyGenBench. PhyWorldBench covers heat in two of its 50 subcategories and electricity in one, about 4\% and 2\% of its prompts, whereas each domain makes up about a fifth of PhyRubric's captions. No prior benchmark covers acoustics. These under-covered domains are exactly those in which the cause (\eg, an electric current, heat, or sound) is invisible, so a model must link a visible trigger to a visible effect without seeing what connects them.

\noindent\textbf{Commonsense only.} We mark this column when every caption can be judged from everyday experience. PhyWorldBench includes an Anti-Physics category whose prompts deliberately violate physics, PhyGenBench organizes its prompts by 27 physical laws, and T2VPhysBench builds its prompts from first-principles laws such as energy conservation. Judging these videos requires treating physical violations as correct or applying formal physics knowledge, neither of which reflects how non-experts judge videos.

\noindent\textbf{Rubric.} Several benchmarks pair each prompt with questions or standards: PhyGenBench generates physics questions with GPT-4o, PhyWorldBench lists physics standards per prompt, and PhysVidBench derives yes/no questions from each prompt. None of them, however, assigns a severity to its items or specifies the visible signs that count as a violation, so a failed question weighs the same whether an object vanishes or a color drifts slightly, and the score is an average of answers rather than a rule over severities. PhyRubric fixes both, and every rubric is verified by humans before any video is scored.

\noindent\textbf{Train set.} Only VideoPhy and VideoPhy-2 release human-labeled videos for training a judge, from which VideoCon-Physics and VideoPhy-2-AutoEval are trained. Their labels are holistic 1--5 ratings, whereas PhyRubric-Train provides item-level verdicts with timestamped evidence.

\noindent\textbf{Explainable.} A judge is explainable when it reports which requirement fails and what in the video supports that verdict. The judges of VideoPhy, VideoPhy-2, and PhyGenBench output scores only. PhyWorldBench and PhysVidBench are partially explainable: each yes/no answer refers to one requirement, so a negative answer indicates what is missing, but neither cites evidence from the video, and both judges answer from a text description of the video rather than from the video itself.

\noindent\textbf{Open-source.} We mark this column when a physics-specific judge is trained and its weights are released. VideoPhy and VideoPhy-2 release their judges, VideoCon-Physics and VideoPhy-2-AutoEval. PhyWorldBench, PhyGenBench, and PhysVidBench instead rely on general-purpose models, mainly proprietary ones (OpenAI o1, GPT-4o, and Gemini 2.5 Flash, respectively). T2VPhysBench uses human evaluation only and has no judge, which we mark as not applicable.

\subsection{Statistics}
\label{sup:stats}

\input{Tab/taxonomy}
\input{Tab/violation}

\begin{figure}[t]
\centering
\includegraphics[width=0.8\linewidth]{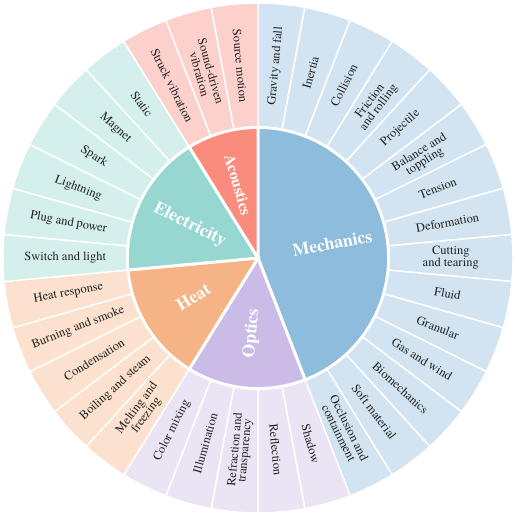}
\caption{Taxonomy of PhyRubric-Bench. The inner circle shows the five physical domains, sized by the number of captions, and the outer ring shows the 34 categories.}
\label{fig:taxonomy}
\end{figure}

Fig.~\ref{fig:taxonomy} shows the taxonomy of PhyRubric-Bench, and Tab.~\ref{tab:taxonomy} lists the categories of each domain and the number of training and evaluation captions. Captions average 21.1 words in PhyRubric-Train and 22.8 in PhyRubric-Bench. Each rubric contains 5.1 SA items and 7.0 PC items on average, and each PC item carries one of ten violation types (Tab.~\ref{tab:violation}). The 3,788 training videos carry 45.9K item-level verdicts, and the 2,433 benchmark videos carry 29.3K.

\subsection{Caption and Rubric Construction}
\label{sup:construction}
\label{app:rubric_full}

\noindent\textbf{Caption Construction.} Captions are drafted one category at a time. For each of the 34 categories, Claude Fable 5.1 and GPT-5.6 Sol receive the scope note of the category and the drafting instructions in Sec.~\ref{sup:prompts_data}, and each model proposes a batch of candidates. Beyond the requirements in Sec.~\ref{sec:bench}, the instructions add three rules. (1) A caption states the trigger and the visible outcome in plain words but never how the outcome unfolds. (2) It contains no law names, numbers, decorative adjectives, or camera directions. (3) Its outcome must hold under every reasonable rendering rather than depend on a threshold, such as how hard an object is pushed. Each candidate also comes with two hidden fields that neither the generator nor the VLM judge ever sees: one to three violation types that the event puts at risk, and an \emph{expectation} briefly describing how the outcome should unfold. Both guide rubric drafting.

Candidates are first checked automatically for a well-formed record, a valid category, and the absence of numerals, and then reviewed manually in two stages. First, the categories are divided between two reviewers. Each reviewer marks every candidate in the assigned categories as keep, edit, or remove, provides a revised caption for each edit, and compares the captions within each category for repeated scenes, lack of variety, and events that belong to another domain. The other reviewer then cross-checks every decision and revision. Second, a third reviewer goes through all remaining captions to ensure consistent standards across categories. Disagreements are resolved by discussion. Review removed 242 candidates: 99 in heat, 48 in electricity, 48 in mechanics, 24 in acoustics, and 23 in optics.

\noindent\textbf{Rubric Construction.} Every caption in both splits receives a rubric by the same procedure. Claude Fable 5.1 drafts the rubric from the caption, the scope note of its category, its violation types and expectation, and approved rubrics from the same domain as examples, following the instructions in Sec.~\ref{sup:prompts_data}. The instructions specify what each item may say: an SA item states a fact the caption requires and nothing else, and a PC item pairs a physical requirement with two to four visible signs of failure, constrains direction and order but never magnitude, and never requires an invisible cause such as heat, current, or sound to be seen.

Every draft is first run through a checking program, which flags a draft that is malformed, alters the caption, has too few or too many items, or makes some score from 1 to 5 unreachable. It also flags numeric thresholds, vague words, SA items that judge physics or are conditional, secondary SA items that overlap a PC item, and declared violation types that no PC item covers.

The 1,192 drafts are then divided evenly among three reviewers regardless of category. Each reviewer reads every assigned draft together with its flags and decides whether it passes, checking whether each item tests what the caption puts at risk and and whether every sign of failure is visible in the video. A draft that does not pass is revised with the LLM, checked by the program again, and reviewed until it passes. Finally, for cross-checking, each reviewer re-reads a random 10\% of the rubrics approved by another reviewer, in a cycle, and agrees with the original decision in 93\% of cases. The remaining cases, mostly wording refinements rather than incorrect items, are discussed and revised.
Fig.~\ref{fig:rubric_full} shows the full rubric of the desk-lamp caption in Fig.~\ref{fig:rubric_example}.

\input{figs/rubric_full}

\subsection{Annotation}
\label{sup:supp_annotator}
\label{sup:annotator}

\noindent\textbf{Agentic annotator.} Both splits are first labeled by Claude Sonnet 5.5 running as a tool-using agent. Videos are labeled by independent agent instances, each handling five videos from mixed generators. Every video is labeled in a fresh context, so the agent sees only its caption, rubric, and video, never the labels of other videos. The agent sees the video only through a frame-inspection tool with two operations. \textbf{Overview} returns 16 uniformly spaced frames of the whole clip, each stamped with its timestamp. \textbf{Zoom} returns up to 16 consecutive frames at the full frame rate within a time window chosen by the agent, optionally cropped to a region. For every video, the agent first takes an overview and writes a description of the whole clip before reading the rubric in detail. It then zooms at least once into the time window and region where the central event should occur, and zooms again whenever a decision hinges on a moment it has not yet seen at full frame rate. An item that involves a small object or a fast action is never decided from the overview alone. The full instructions are given in Sec.~\ref{sup:prompts_data}.

The agent applies the verdict definitions of Sec.~\ref{sec:rubric} with one additional requirement: an item passes only when the agent can point to the frames that show it, since the caption states what was requested, not what the video contains. For every item, the agent writes its verdict with one or two sentences of timestamped evidence, and for a failed PC item it also records which sign in the Fail statement was observed. The agent never outputs scores; SA and PC are computed from its verdicts by the rules in Sec.~\ref{sec:rubric}. PhyRubric-Train uses these labels as they are.

\noindent\textbf{Human verification.} For PhyRubric-Bench, the three reviewers who built the rubrics verify every verdict against the video. Each reviewer watches the video, reads every verdict together with its evidence, and keeps it or changes it; the agent's original labels are stored separately so that every change can be counted. Generator names are hidden and videos are shuffled across generators, so that a reviewer cannot be biased toward or against a particular model. The three reviewers first verify 20 videos together to align their standards, and these are excluded from the statistics below. Next, 150 videos, 15 per generator, are each verified independently by two reviewers, with each pair of reviewers covering one third of them. The remaining videos are divided evenly among the three reviewers.

\noindent\textbf{Overturn rate.} Reviewers changed 2.8\% of the SA verdicts and 5.6\% of the PC verdicts. Most PC changes (61\%) turn a pass into a fail, \ie, a violation the agent missed, typically a brief or small one, which matches the error patterns of Sec.~\ref{sec:error_analysis}. After verification, 88.1\% of the videos keep both their SA and PC scores, and most of the remaining ones change by one point. On the doubly verified videos, the two reviewers agree on 95.3\% of the verdicts (Cohen's $\kappa={}$0.87), so the corrections reflect what the video shows rather than the preference of one reviewer. Because PhyRubric-Train is labeled by the same agent under the same instructions, the overturn rate also approximately estimates the error rate of its training labels, about 3\% for SA and 6\% for PC verdicts.

\section{Experimental Setup}
\label{sup:suppl_setup}

\subsection{Generation Settings}
\label{sup:generation}

\begin{table}[t]
\centering
\caption{Generation settings of the ten video generators. Open-source models run with their official checkpoints and default sampling recipes through the \texttt{diffusers} pipelines, one video per caption with seed 42; the two proprietary models are accessed through their official APIs, which expose neither a seed nor sampling parameters. CFG: classifier-free guidance scale; for Wan2.2-T2V-A14B the two values belong to the high- and low-noise experts. The distilled LTX models use a fixed timestep schedule without guidance.}
\label{tab:generation}
\small
\setlength{\tabcolsep}{4pt}
\resizebox{\linewidth}{!}{%
\begin{tabular}{l c c c c c}
\toprule
\textbf{Model} & \textbf{Resolution} & \textbf{Frames} & \textbf{FPS} & \textbf{Steps} & \textbf{CFG} \\
\midrule
CogVideoX1.5-5B        & $1360\times768$ & 81  & 16 & 50 & 6.0 \\
HunyuanVideo-1.5-8B    & $848\times480$  & 121 & 24 & 50 & 6.0 \\
LTX-Video-2B           & $1216\times704$ & 121 & 24 & 8+4 & -- \\
Wan2.2-T2V-A14B        & $832\times480$  & 81  & 16 & 40 & 4.0 / 3.0 \\
Wan2.2-TI2V-5B         & $1280\times704$ & 121 & 24 & 50 & 5.0 \\
LTX-2.3-22B            & $1280\times704$ & 121 & 24 & 8  & -- \\
Cosmos-Predict2.5-2B   & $1280\times704$ & 93  & 16 & 36 & 7.0 \\
Cosmos-Predict2.5-14B  & $1280\times704$ & 93  & 16 & 36 & 7.0 \\
Seedance 2.5           & $854\times480$  & 121 & 24 & -- & -- \\
Veo 3.1                & $1280\times720$ & 144 & 24 & -- & -- \\
\bottomrule
\end{tabular}}
\end{table}

Tab.~\ref{tab:generation} lists the settings of every generator. Each caption is passed verbatim as the prompt, with no prompt rewriting, style suffix, or negative prompt beyond a model's own default. Clip length follows each model's native recipe, between 5.0 and 6.0 seconds. Seedance 2.5 is run at 480p, 16:9, and 5 seconds, with audio and watermark disabled. Veo 3.1 (\texttt{veo-3.1-generate-001}) is run at 720p and 6 seconds, without audio.

\subsection{Evaluation Metrics}
\label{sup:suppl_metrics}

\input{Tab/suppl_metric}

\noindent\textbf{Hard and soft judge scores.}
A generative judge emits an integer score, which we call the \textbf{hard} score.
Following G-Eval~\cite{liu2023g} and Q-Align~\cite{wu2023q}, we also extract the probabilities the model assigns
to the tokens ``1''--``5'' at the position where it generates the score, and define the \textbf{soft} score as the
expectation under the renormalised distribution:
\begin{equation}
  s^{\text{soft}} = \sum_{k=1}^{5} k \cdot \frac{p(k)}{\sum_{j=1}^{5} p(j)},
  \label{eq:soft}
\end{equation}
where $p(k)$ is the probability of token ``$k$''.
Unlike the hard score, the soft score can rank clips that receive the same integer, which reflects the judge's
confidence and \textit{avoids the many ties that a five-valued integer score inevitably produces}, so that correlation
and ranking metrics can measure finer-grained agreement with humans.

VideoCon-Physics~\cite{bansal2025videophy} and VideoScore~\cite{he2024videoscore} output continuous scores by construction
(a Yes-probability and a regression output, respectively), so they have a soft score but no hard score.
VideoPhy-2-AutoEval~\cite{bansal2026videophy}, like PhyRubric-Judge, emits integer scores. We compute its soft score with the same construction, without changing its prompt or decoding, so that all judges are compared on soft scores.

\noindent\textbf{Metrics used by prior judges.}
Existing judges are validated with different metrics~\cite{bansal2025videophy,bansal2026videophy,he2024videoscore,he2025videoscore2}.
We report all of them for every judge (Tab.~\ref{tab:suppl_metric_overview}) and group them into three families below,
where, for each clip $i$ and each dimension (SA or PC), $x_i$ is the judge's (hard or soft) score,
$y_i\in\{1,\dots,5\}$ the human label, and $n$ the number of clips.

\noindent\textbf{A. Correlation.}
These metrics measure whether the judge orders clips in the same way as humans, regardless of the score scale. We report them on both the hard and the soft score, because the prior judges validate with different score types:
VideoPhy-2-AutoEval reports Pearson on its integer score, VideoScore2 on a probability-derived float.
\begin{itemize}
\item \textbf{Pearson} $r$ is the linear correlation between $x$ and $y$:
\begin{equation}
  r = \frac{\sum_i (x_i-\bar{x})(y_i-\bar{y})}{\sqrt{\sum_i (x_i-\bar{x})^2}\sqrt{\sum_i (y_i-\bar{y})^2}}.
\end{equation}
It ranges in $[-1, 1]$, and a higher value indicates stronger linear agreement with human labels.

\item \textbf{Spearman} $\rho$ is the Pearson correlation computed on ranks instead of raw scores, where clips
with equal scores share the average of their ranks. It ranges in $[-1, 1]$, and a higher value indicates that the
judge ranks clips more consistently with human labels, even if the two scores are not linearly related.

\end{itemize}

\noindent\textbf{B. Ranking.}
\begin{itemize}
  \item \textbf{ROC-AUC} splits clips by the human label into positives $\mathcal{P} = \{i : y_i \geq 4\}$ and
  negatives $\mathcal{N}$, and measures the probability that a positive clip receives a higher judge score than a
  negative one, counting ties as one half:
  \begin{equation}
    \mathrm{AUC} = \frac{1}{|\mathcal{P}||\mathcal{N}|}\sum_{i\in\mathcal{P}}\sum_{j\in\mathcal{N}}
    \big[\ind(x_i > x_j) + \tfrac{1}{2}\ind(x_i = x_j)\big].
  \end{equation}
It ranges in $[0, 1]$, where $0.5$ corresponds to random ordering and a higher value is better.
It thresholds only the human label, so it applies to judge scores on any scale.
Following VideoCon-Physics~\cite{bansal2025videophy}, which computes ROC-AUC on its continuous Yes-probability,
we compute it on continuous scores: for judges that output integers, we use their soft scores (Eq.~\ref{eq:soft}).
\end{itemize}

\noindent\textbf{C. Agreement.}
These metrics check whether the judge's decisions or scores match the human labels, in two ways.

\noindent(1) \textit{Binary agreement.}
Both scores are thresholded into pass/fail decisions, $z_i = \ind(y_i \geq 4)$ and $\hat{z}_i = \ind(x_i \geq \theta)$,
where $\theta$ is the judge's own pass line (Tab.~\ref{tab:suppl_passlines}).
We compute the metrics for SA and PC separately, and for their joint decision, under which a clip passes only if both pass.
\begin{itemize}
  \item \textbf{Binary accuracy} is the fraction of clips with $\hat{z}_i = z_i$.
  \item \textbf{F1} is the harmonic mean of precision (the fraction of judge-passed clips that humans also pass)
  and recall (the fraction of human-passed clips that the judge also passes), with pass as the positive class.
\end{itemize}

\noindent(2) \textit{Score-level agreement.}
Exact and within-1 accuracy are reported for judges that output integer 1--5 scores, and are computed on the
hard score. They measure whether the score can be used directly on the human scale, rather than how well it ranks clips.
\begin{itemize}
  \item \textbf{Exact accuracy} is the fraction of clips with $x_i = y_i$.
  \item \textbf{Within-1 accuracy} is the fraction of clips with $|x_i - y_i| \leq 1$, tolerating the one-point
  disagreements that are common among human annotators.
\end{itemize}
Both range in $[0,1]$, and a higher value is better.

\input{Tab/suppl_passlines}

\subsection{Baseline Setup}
\label{sup:suppl_baselines}

Tab.~\ref{tab:suppl_passlines} lists the dimensions, native output and pass line of every judge.
All baselines are run on the PhyRubric-Bench videos with their official checkpoints, prompts, frame sampling and
decoding. Each baseline receives the caption and the video, but not the rubric.

\medskip

\noindent\textbf{VideoCon-Physics}~\cite{bansal2025videophy} and \textbf{VideoPhy-2-AutoEval}~\cite{bansal2026videophy} share the mPLUG-Owl-video 7B architecture. We follow their official implementations, including the prompts (Sec.~\ref{sup:suppl_prompts_baselines}) and 32 uniformly sampled frames at $224\times224$. Each is queried once for SA and once for PC. VideoCon-Physics uses the normalized probability of answering ``Yes'' as its soft score:
\begin{equation}
  s = \frac{p(\text{Yes})}{p(\text{Yes}) + p(\text{No})}.
\end{equation}
VideoPhy-2-AutoEval takes its greedily decoded digit as the hard score, and its soft score follows Eq.~\ref{eq:soft}.

\medskip

\noindent\textbf{VideoScore}~\cite{he2024videoscore} follows its official implementation, with up to 48 uniformly
sampled frames and a single prompt containing the caption (Sec.~\ref{sup:suppl_prompts_baselines}).
Instead of generating text, it outputs five scores in $[1,4]$ from a regression head, one per evaluation dimension.
We use \emph{text-to-video alignment} as SA and \emph{factual consistency} as PC.

\medskip

\noindent\textbf{VideoScore2}~\cite{he2025videoscore2} follows its official implementation, with video sampled at 2 fps,
its three-dimension prompt containing the caption (Sec.~\ref{sup:suppl_prompts_baselines}), and sampling at
temperature 0.7 with a fixed seed per clip. It first generates a reasoning trace and then an integer score for each
dimension, which we parse as the hard score. Its soft score applies Eq.~\ref{eq:soft} at each score token.
We use \emph{text-to-video alignment} as SA and \emph{physical and common-sense consistency} as PC.

\medskip
\noindent\textbf{Zero-shot VLMs.} Qwen3-VL-8B-Instruct, Qwen3-VL-32B-Instruct and Qwen3-VL-235B-A22B-Instruct receive the same system prompt and input as PhyRubric-Judge (Sec.~\ref{sup:suppl_prompts_ours}), namely the SA and PC rating scales described in words, the caption, and 16 uniformly sampled frames at up to $448\times448$. They decode greedily and are parsed identically. The 8B model is the base of PhyRubric-Judge.

\medskip
\noindent\textbf{Proprietary VLM.} Gemini 4 Argon receives the same system prompt and output format as PhyRubric-Judge (Sec.~\ref{sup:suppl_prompts_ours}). The video is sent directly and sampled by the model at 3 fps, which yields about 15 frames for a five-second clip, close to the 16 frames given to the local judges. Thinking cannot be disabled, so we set it to the lowest level the model allows (LOW), and we decode at temperature 0. The integer the model outputs is its hard score. The model exposes no token probabilities, so Eq.~\ref{eq:soft} cannot be applied. Instead, in a second run, we add two fields to the requested JSON that ask the model for its probability of each score from 1 to 5, separately for SA and PC, and take the expectation of this stated distribution as the soft score.

\section{Additional Experimental Results}
\label{sup:suppl_results}

\input{Tab/familyA}
\input{Tab/familyB}
\input{Tab/familyC}

\subsection{Comparison with Existing VLM Judges}
\label{sup:suppl_judges}

Tab.~\ref{tab:suppl_judges_corr}--\ref{tab:suppl_judges_agree} report families A--C for every judge of
Tab.~\ref{tab:judge}. Judges whose dimensions differ from ours are mapped to SA and PC as listed in
Tab.~\ref{tab:suppl_passlines}. All baselines use their official checkpoints, prompts, frame sampling and decoding;
implementation details are given in Sec.~\ref{sup:suppl_setup}.
PhyRubric-Judge achieves the best result on every metric among open-source judges, and is on par with the proprietary Gemini 4 Argon.

\medskip
\noindent\textbf{Comparison with Gemini 4 Argon.} In Tab.~\ref{tab:suppl_judges_corr}, Gemini 4 Argon leads on the hard-score rows. This is because its thinking cannot be disabled (Sec.~\ref{sup:suppl_baselines}), so its integer score follows a reasoning trace, whereas the hard score of PhyRubric-Judge, like that of every other baseline, is a single greedy token without reasoning. A direct comparison of hard scores therefore favors Gemini 4 Argon. Moreover, a paired bootstrap over clips (2,000 resamples) shows that none of these gaps is significant, as all 95\% confidence intervals include zero.

In Tab.~\ref{tab:suppl_judges_agree}, Gemini 4 Argon is best on binary accuracy, on SA and jointly. These numbers are inflated because it rarely predicts pass: it passes only 10\% of clips on SA and 3\% jointly, whereas humans pass 20\% and 14\%. Since most clips fail, even a trivial judge that fails every clip reaches an accuracy of 0.80 on SA and 0.86 jointly. F1 does not reward this behaviour, and PhyRubric-Judge achieves the higher F1 in every column. Moreover, its joint pass rate (17\%) is close to the human rate (14\%), so it yields usable per-generator scores for a leaderboard, whereas Gemini 4 Argon would leave most generators with a JP near zero (Tab.~\ref{tab:suppl_leaderboard_judge}).

\subsection{Reproducing the Leaderboard with PhyRubric-Judge}
\label{sup:suppl_per_gen}

\input{Tab/leaderboard_judge}

The clip-level agreement of Sec.~\ref{sup:suppl_judges} is a demanding test. In practice, a judge is used to score a generator, where per-clip errors average out. We therefore recompute the leaderboard of Tab.~\ref{tab:leaderboard} from judge scores (Tab.~\ref{tab:suppl_leaderboard_judge}).

PhyRubric-Judge reproduces the human leaderboard closely ($r=0.93$, $\rho=0.87$) and keeps its JP values on the human scale, with an MAE of 4.65 points. It separates the two proprietary generators from the open-source ones by a wide margin, although it swaps their order, and it overestimates the weakest open-source generators. Gemini 4 Argon reaches a slightly higher Pearson correlation ($r=0.94$), but this is driven by the two proprietary generators alone: it compresses the other eight to a JP below 3\%, so it ranks them less reliably ($\rho=0.81$) and deviates far more from the human values (MAE 12.65). PhyRubric-Judge thus provides a leaderboard that is usable in absolute terms, not only in ranking.

\subsection{Ablations of PhyRubric-Judge}
\label{sup:suppl_ablation}

\input{Tab/ablation}
\input{figs/ablation_formats}

Tab.~\ref{tab:ablation} examines four design choices of PhyRubric-Judge. Each variant trains Qwen3-VL-8B with the recipe of Sec.~\ref{sec:exp_setup} and changes only one factor. Fig.~\ref{fig:ablation_formats} shows the input and output of each variant.

\question{Should the judge see the rubric?}
The annotator needs the rubric, whereas the judge takes only the caption (Sec.~\ref{sec:judge}). We test this choice in three ways. (1) \textbf{Giving the judge the rubric as input and training it to output the scores only improves correlation by 0.06 on SA and 0.05 on PC}. The rubric thus carries information that the judge cannot fully recover from the caption, but every new caption would then need a rubric before the judge can be applied. (2) \textbf{Giving the rubric and training the judge to output item-wise verdicts instead of scores does not help}, as the scores derived from its verdicts are less accurate than directly predicted ones. (3) \textbf{The rubric is of no use to the untrained model.} Given the rubric, zero-shot Qwen3-VL-8B becomes worse on both dimensions, and its PC correlation falls below zero. We therefore keep the caption-only input, so that the judge applies to any new caption, and use the rubric at annotation time only.

\question{Where should the diagnosis go?}
The judge emits the scores first and the diagnosis after them. We compare two alternatives in which the diagnosis is generated first. (1) The diagnosis is the first field of the same JSON object. (2) The diagnosis is a chain of thought inside \texttt{<think>} tags, followed by a JSON object that holds the scores only. The diagnosis text is identical in all three cases, and only its position changes. Both alternatives lose 0.06--0.07 on SA and 0.12 on PC. The reason is that the soft score is the expectation over the score tokens (Eq.~\ref{eq:soft}). When the score comes first, this distribution still reflects the uncertainty of the judge. After a committed diagnosis, it becomes nearly one-hot, and this uncertainty is lost. Scores first thus keeps an informative soft score and a diagnosis.

\question{Does reinforcement learning help?}
Starting from the SFT judge, we continue training the same LoRA with GRPO for one epoch, with 4 samples per prompt at temperature 1, a batch of 8 prompts, a learning rate of $10^{-5}$, and no KL term. The reward is a weighted sum of four terms,
\begin{equation}
  R = r_{\text{SA}} + r_{\text{PC}} + 0.5\, r_{\text{fmt}} + 0.5\, r_{\text{assess}},
\end{equation}
where $r_{\text{fmt}}$ rewards a valid output format and $r_{\text{assess}}$ rewards a correct \texttt{pc\_assessable} flag. The score terms measure how close the predicted score $\hat{s}_d$ is to the label $s_d$:
\begin{equation}
  r_{d} = 1 - \frac{|\hat{s}_d - s_d|}{4}, \quad d \in \{\text{SA}, \text{PC}\}.
\end{equation}
We run this recipe twice, (1) without thinking, starting from Default, and (2) with thinking, starting from Diagnosis as thinking, and compare each run with its own starting point.

As shown in Tab.~\ref{tab:ablation}(c), neither run improves over its SFT starting point, and the thinking-free run is clearly better than the one with thinking, especially on PC ($r$ of 0.314 vs.\ 0.086). A possible reason is that GRPO has mainly proven effective on reasoning tasks with verifiable answers, such as mathematics, and may bring less to a perceptual task such as judging physical commonsense in video. We leave this to future work and keep SFT as the default.

\begin{figure}[t]
\centering
\includegraphics[width=\linewidth]{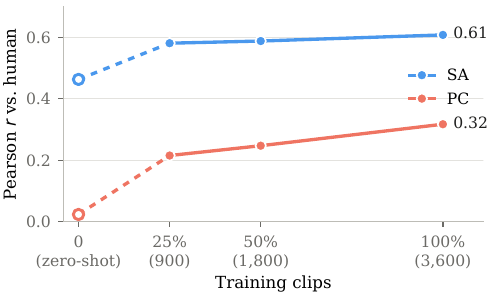}
\caption{Pearson $r$ with human labels against the fraction of PhyRubric-Train used for training. The hollow point at 0\% (dashed) is zero-shot Qwen3-VL-8B.}
\label{fig:data_scaling}
\end{figure}

\question{How much training data is needed?}
PhyRubric-Train is split into 3,600 videos for training and 188 videos from 47 held-out captions for monitoring the training loss. All judges in this paper are trained on the 3,600-video split. We train on random 25\% and 50\% subsets of it (900 and 1,800 videos) with the same number of epochs (Fig.~\ref{fig:data_scaling}). SA improves only slightly with data (0.58, 0.59, 0.61), whereas PC improves steadily (0.22, 0.25, 0.32). Physical plausibility is thus the dimension that benefits most from more supervision. Since the training labels are produced automatically by the annotator, the judge can be strengthened further at low cost, by generating and labeling more videos without human annotation.

\subsection{Can Prompting Fix Physical Violations?}
\label{sup:suppl_regen}

\input{Tab/regen}
\input{figs/regen_case1}

\noindent\textbf{Setup.} We ask whether feeding the diagnosis back into the prompt helps a generator fix its failures. From the 245 Wan2.2-T2V-A14B videos of PhyRubric-Bench, we take the 193 that PhyRubric-Judge scores SA$\leq$3 or PC$\leq$3 and sample 100 of them, stratified by domain. Each caption is re-rendered once under two conditions with the same sampler settings: the original caption with a new seed, which controls for resampling, and a prompt of 45--70 words written by Claude Sonnet~5.5 from the caption and the judge's diagnosis of the original video, which names the agent, states the cause before its effect, and fixes the order of events. All 200 new videos are labeled blind by the annotator with the original rubrics (Sec.~\ref{sup:annotator}), and Tab.~\ref{tab:regen} compares them with the original videos.

\noindent\textbf{The rewrite fixes adherence, not physics.} The diagnosis-driven rewrite raises the mean SA from 2.55 to 3.19 and the SA pass rate from 11\% to 31\%, against 2.76 and 18\% for resampling alone, and its gain over the new seed is significant. PC, in contrast, does not move: neither condition differs from the original in mean PC or in PC pass rate, and the joint pass rate rises only because more videos now pass SA. The rewrite can make the generator depict the caption, but it does not make it obey physics.

\noindent\textbf{What the diagnosis changes.} Fig.~\ref{fig:regen_case1} shows an example. The diagnosis states that the balloon floats on a string and that no throw, no arc and therefore no order of events is seen. The rewrite is specific about exactly these points: it starts from the arm swinging and releasing the balloon, names the high arc above the fence and the splash, and keeps the scene of the original video. The generator follows this prompt, and SA rises from 2 to 5. The diagnosis turns the caption into a prompt that states what the generator left out.

\begin{figure*}[t]
\centering
\begin{subfigure}[b]{0.53\textwidth}
  \centering
  \includegraphics[height=2.0in]{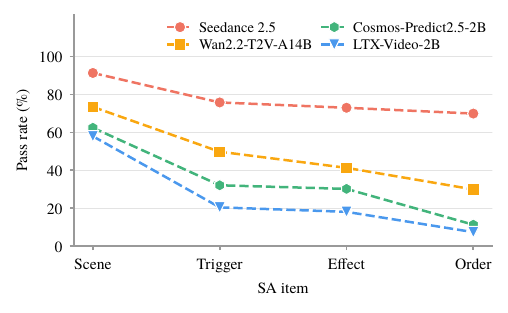}
  \caption{SA item pass rate along the causal episode}
  \label{fig:sa_chain}
\end{subfigure}\hfill
\begin{subfigure}[b]{0.45\textwidth}
  \centering
  \includegraphics[height=2.0in]{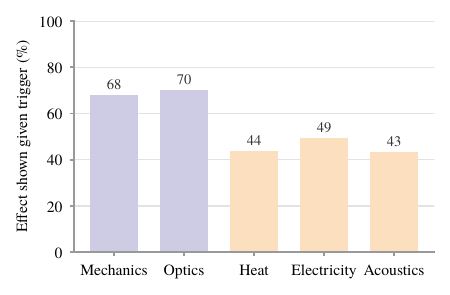}
  \caption{Effect shown given the trigger, per domain}
  \label{fig:sa_effect}
\end{subfigure}
\caption{Semantic adherence along the causal episode on PhyRubric-Bench. (a) Pass rate of SA items at each stage, scene, trigger, effect, and their order, for four generators. (b) Fraction of videos that show every effect among those that show the trigger, per domain.}
\label{fig:sa_analysis}
\end{figure*}

\input{Tab/error_types}
\input{Tab/suppl_error_types_gen}

\section{More Analysis}

\subsection{Error Analysis}
\label{sup:error_analysis}

The ten violation types are defined in Tab.~\ref{tab:violation} (Sec.~\ref{sup:stats}). Tab.~\ref{tab:error_types} reports, for each domain, the share of assessable videos with at least one failed PC item of each type, and Tab.~\ref{tab:error_types_gen} breaks the same rates down by generator.

\noindent\textbf{Adherence decays along the causal episode.} Every caption in PhyRubric describes a complete episode, namely a scene, a trigger, its effect, and their order, and the SA items follow this structure. Fig.~\ref{fig:sa_analysis}(a) plots the pass rate of each stage. Models render the scene most reliably, but the pass rate falls for the trigger and its effect, and falls further for their order. The decay is steepest for the weakest models. Fig.~\ref{fig:sa_analysis}(b) focuses on the step from trigger to effect. In mechanics and optics, where the effect follows directly from contact or light, 68\% and 70\% of triggered videos show it. In heat, electricity, and acoustics, where the cause reaches the effect through an invisible medium, only 43--49\% do.

\noindent\textbf{Violation types follow the physics of each domain.} Tab.~\ref{tab:error_types} shows that the violation types are not spread evenly over the domains but follow what each domain asks for. Missing consequence is the most frequent type overall (17.3\% of assessable videos) and peaks in acoustics (44.4\%) and electricity (24.6\%), where an invisible sound or current must produce a visible effect, against 9.1\% in mechanics. Acoustics also leads in causeless motion (21.8\%), since a vibration that should follow a sound is often rendered as an object that shivers on its own. Process direction concentrates in heat and electricity (11.2\% and 13.0\%), where a melting process reverses or a lamp switches back off. Optics carries the appearance types, namely shape and identity (11.7\%), where a reflection or a refracted object does not match its source, and attribute drift (9.3\%), where the color of an untouched object changes. Mechanics, where several objects interact, leads in permanence and count (18.5\%), material response (8.6\%), and trajectory and momentum (8.2\%). Each domain thus exposes a different class of failure, and a benchmark concentrated on mechanics would leave most of these failures unmeasured.

\noindent\textbf{Object permanence separates the generators, while missing consequence is shared.} Tab.~\ref{tab:error_types_gen} breaks the same rates down by generator, and the two most frequent types behave differently. Permanence and count failures vary by a factor of six. Objects appear, vanish, or duplicate in 33.3\% of the assessable videos of CogVideoX1.5-5B and in 22--23\% of those of LTX-Video-2B and Wan2.2-TI2V-5B, but in only 5.2\% for Seedance~2.5 and 8.6\% for HunyuanVideo-1.5 and Cosmos-Predict2.5-14B, with Veo~3.1 in between at 12.9\%. Missing consequence, in contrast, stays between 11\% and 22\% for every generator, and causeless motion between 6\% and 13\%. Missing consequence is the most frequent type for all generators except the two weakest, where permanence and count leads. Keeping objects consistent over time thus separates the generators, whereas linking a cause to its visible effect is a weakness they all share and remains the dominant failure even of the generators that largely keep objects consistent.

\subsection{Qualitative Examples}
\label{sup:qualitative}

\begin{figure*}[t]
\centering
\includegraphics[width=\textwidth]{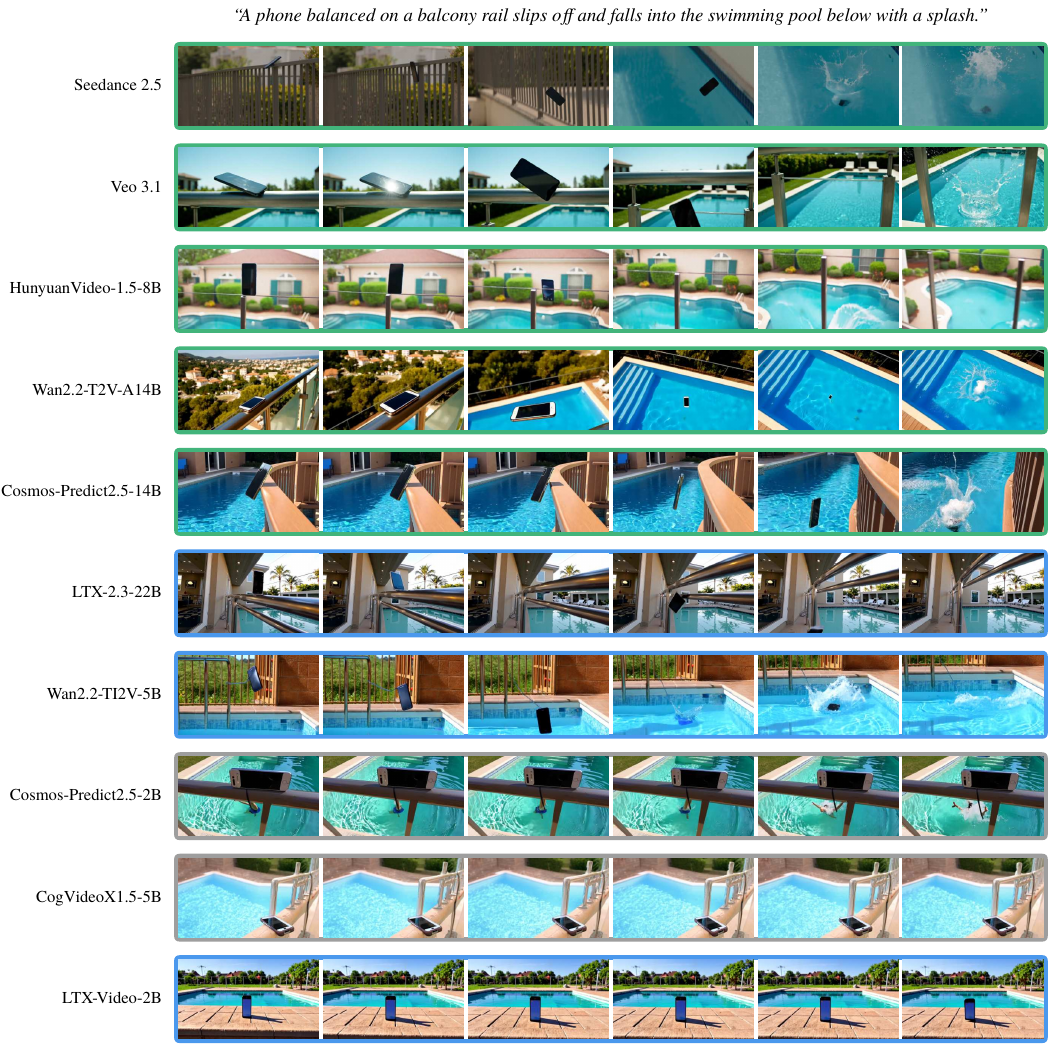}
\caption{Videos of ten generators for a mechanics caption. Each row shows six evenly spaced frames of one video, and the border color gives the human verdict. \textcolor{qualpass}{Green}: the event is shown and physically plausible (SA$\geq$4, PC$\geq$4). \textcolor{qualphys}{Red}: the event is shown but violates physics (SA$\geq$4, PC$<$4). \textcolor{qualsa}{Blue}: the event is shown incompletely (SA$<$4). \textcolor{qualavoid}{Gray}: the event is not shown.}
\label{fig:qual_mech}
\end{figure*}

\begin{figure*}[t]
\centering
\includegraphics[width=\textwidth]{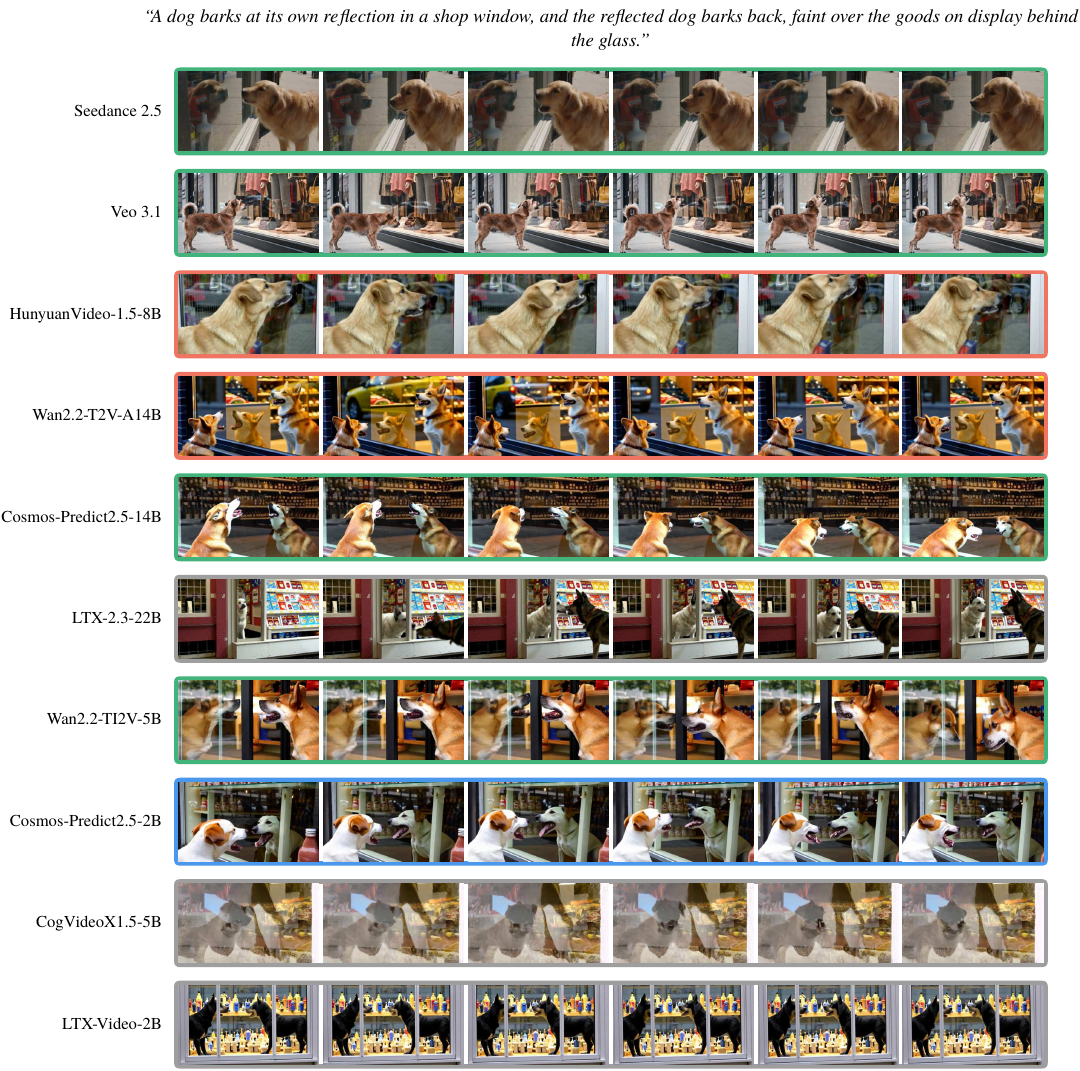}
\caption{Videos of ten generators for an optics caption; border colors as in Fig.~\ref{fig:qual_mech}.}
\label{fig:qual_opt}
\end{figure*}

\begin{figure*}[t]
\centering
\includegraphics[width=\textwidth]{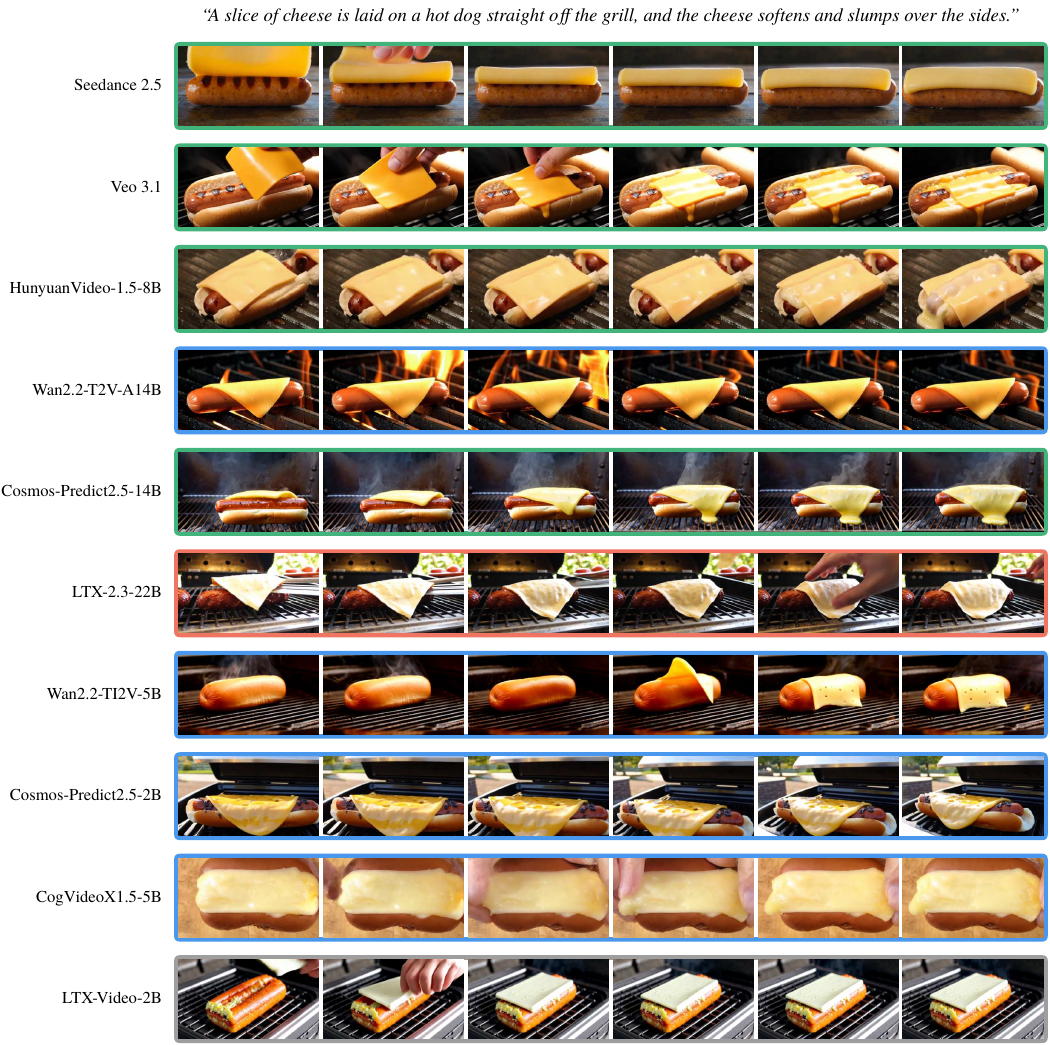}
\caption{Videos of ten generators for a heat caption; border colors as in Fig.~\ref{fig:qual_mech}.}
\label{fig:qual_heat}
\end{figure*}

\begin{figure*}[t]
\centering
\includegraphics[width=\textwidth]{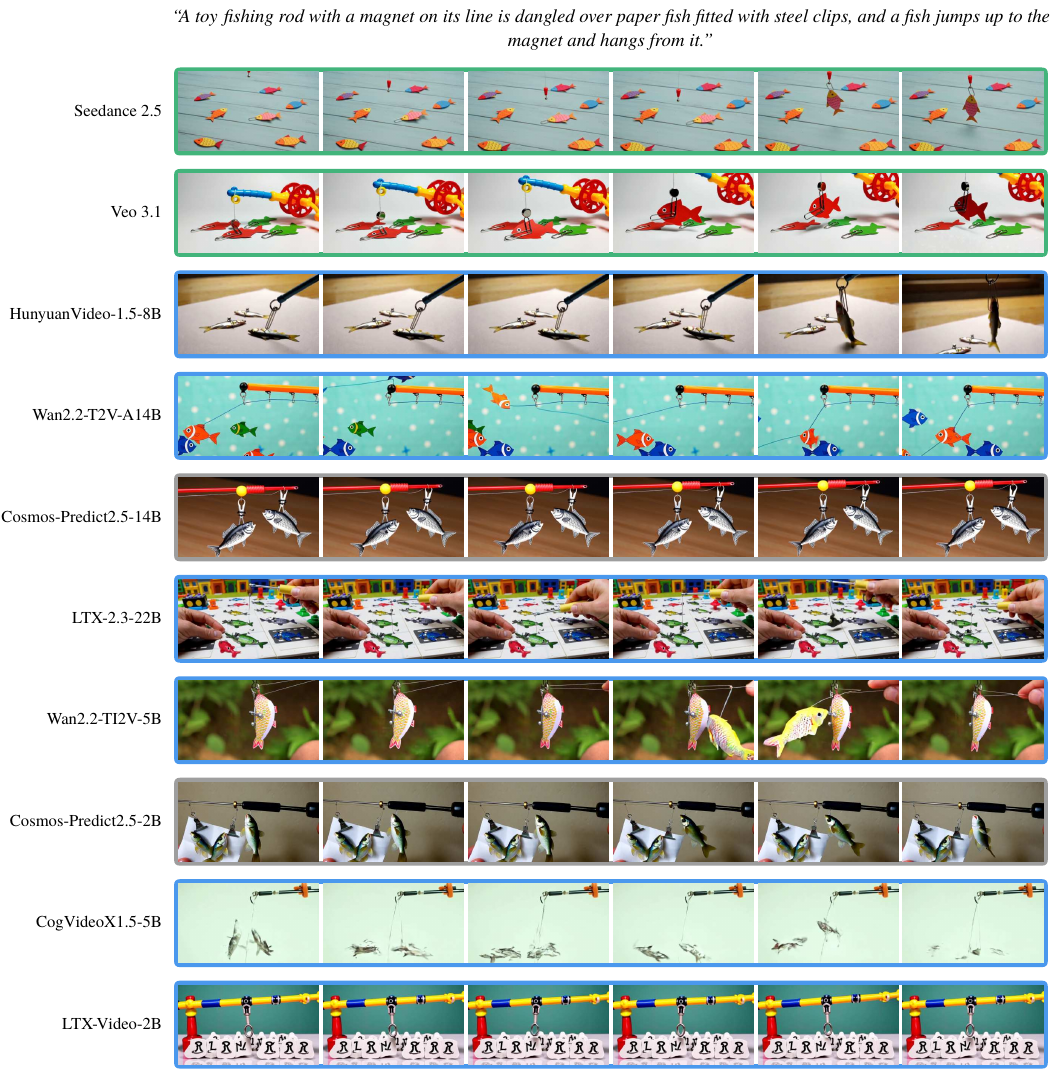}
\caption{Videos of ten generators for an electricity caption; border colors as in Fig.~\ref{fig:qual_mech}.}
\label{fig:qual_elec}
\end{figure*}

\begin{figure*}[t]
\centering
\includegraphics[width=\textwidth]{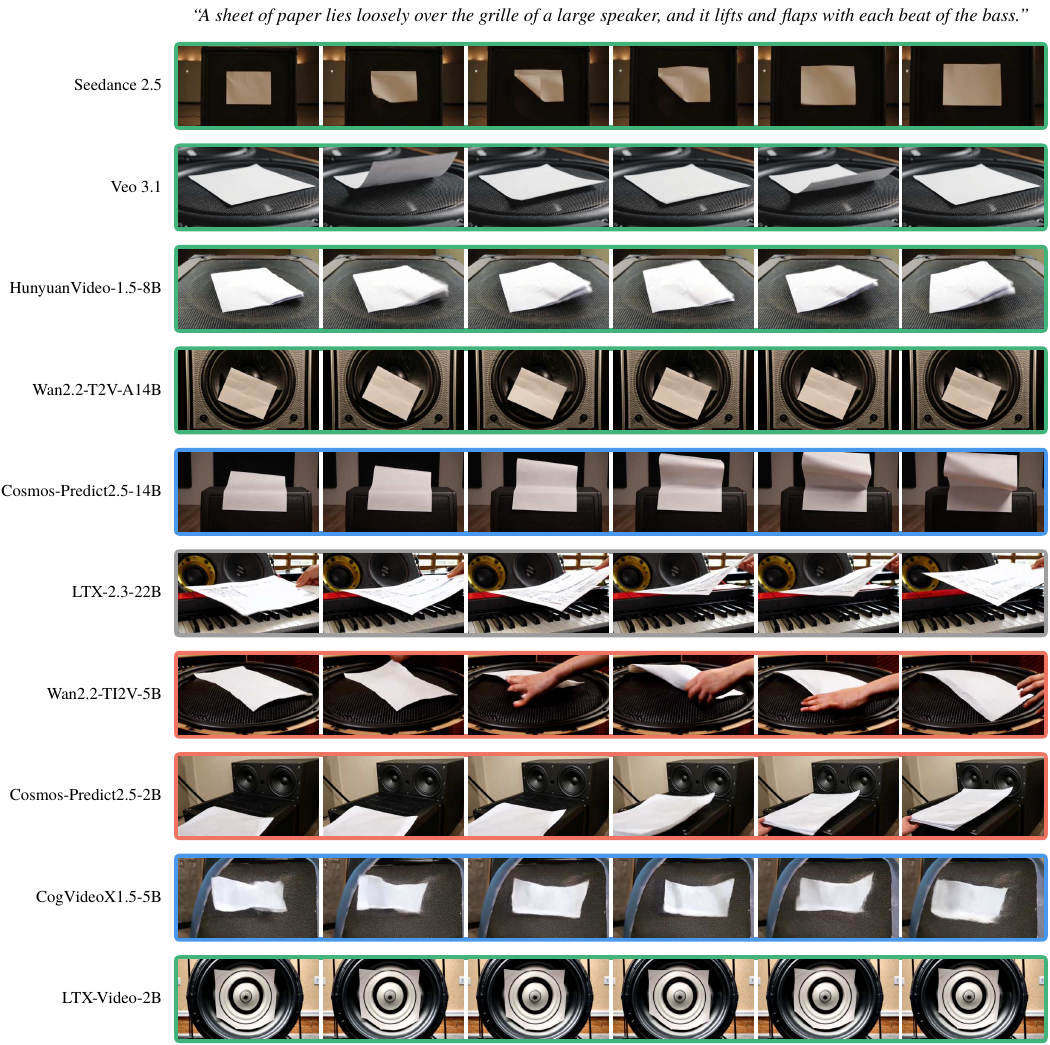}
\caption{Videos of ten generators for an acoustics caption; border colors as in Fig.~\ref{fig:qual_mech}.}
\label{fig:qual_acou}
\end{figure*}

\noindent\textbf{Generated videos.} Figs.~\ref{fig:qual_mech}--\ref{fig:qual_acou} show one caption per domain rendered by all ten generators. The two proprietary models render the event in every example, while open-source models most often fail before physics comes into play. Avoidance rarely looks like a frozen scene. Instead, the generator skips the action the caption describes: it either shows the outcome from the first frame, as in the fish already hanging from the rod (Cosmos-Predict2.5-14B, Fig.~\ref{fig:qual_elec}) or the phone resting on the pool edge throughout (CogVideoX1.5-5B, Fig.~\ref{fig:qual_mech}), or replaces the action with a related one, such as leafing through sheet music (LTX-2.3-22B, Fig.~\ref{fig:qual_acou}). When the event is shown, the physical errors are brief and local, such as the sheet of paper that emerges from beneath the speaker (Cosmos-Predict2.5-2B, Fig.~\ref{fig:qual_acou}): a single rubric item catches it, and the verdict carries the item's severity rather than a holistic impression.

\input{figs/fig_diag_examples}

\noindent\textbf{Judge diagnoses.} Fig.~\ref{fig:diag_examples} shows the output of PhyRubric-Judge on four videos from PhyRubric-Bench. The diagnosis follows the structure of the training targets: a timestamped description of what the video shows, the caption facts that are missing, and each physical violation with the moment it occurs. When the event passes (a), the description is a complete timeline of the event. When the event never occurs (b), the judge says so and marks PC as not assessable. When physics breaks (c, d), the judge names the violation and when it occurs, such as fish that tumble with nothing touching them or balls that appear and vanish on the table.

\clearpage    
\onecolumn
\raggedbottom  
\section{Prompts}
\label{sup:suppl_prompts}

This section lists the prompts used by every judge, and the instructions used to draft captions and rubrics (Sec.~\ref{sup:prompts_data}). Baseline prompts are reproduced verbatim from the official code;
\texttt{\{caption\}} is replaced by the caption of the clip and \texttt{<|video|>} (or the equivalent placeholder) by the
video tokens.

\subsection{Baseline Judges}
\label{sup:suppl_prompts_baselines}

\noindent\textbf{VideoCon-Physics}~\cite{bansal2025videophy} is queried twice per clip, once for SA and once for PC, and
scored as $p(\text{Yes})/(p(\text{Yes})+p(\text{No}))$ at the answer position.

\begin{promptbox}{VideoCon-Physics, SA prompt}
The following is a conversation between a curious human and AI assistant. The assistant gives helpful, detailed, and polite answers to the user's questions.\\
Human: <|video|>\\
Human: Does this video entail the description: "\{caption\}"?\\
AI:
\end{promptbox}

\begin{promptbox}{VideoCon-Physics, PC prompt}
The following is a conversation between a curious human and AI assistant. The assistant gives helpful, detailed, and polite answers to the user's questions.\\
Human: <|video|>\\
Human: Does this video follow the physical laws?\\
AI:
\end{promptbox}

\noindent\textbf{VideoPhy-2-AutoEval}~\cite{bansal2026videophy} is queried twice per clip; the first digit of the greedy
output is the score.

\begin{promptbox}{VideoPhy-2-AutoEval, SA prompt}
The following is a conversation between a curious human and an AI assistant. The assistant gives helpful, detailed, and polite answers to the user's questions.\\
Human: <|video|>\\
Human: Does this video match the description: "\{caption\}"? Please rate the video on a scale from 1 to 5, where 5 indicates a perfect match and 1 indicates no relevance.\\
AI:
\end{promptbox}

\begin{promptbox}{VideoPhy-2-AutoEval, PC prompt}
The following is a conversation between a curious human and an AI assistant. The assistant gives helpful, detailed, and polite answers to the user's questions.\\
Human: <|video|>\\
Human: Does this video adhere to the physical laws? Rate the video on a scale from 1 to 5, where 5 means full compliance and 1 means significant violations.\\
AI:
\end{promptbox}

\noindent\textbf{VideoScore}~\cite{he2024videoscore} uses one prompt for all five dimensions; the five scores are read from
the regression head, and we use \emph{text-to-video alignment} as SA and \emph{factual consistency} as PC.

\begin{promptbox}{VideoScore, regression prompt}
Suppose you are an expert in judging and evaluating the quality of AI-generated videos, please watch the following frames of a given video and see the text prompt for generating the video, then give scores from 5 different dimensions:\\
(1) visual quality: the quality of the video in terms of clearness, resolution, brightness, and color\\
(2) temporal consistency, both the consistency of objects or humans and the smoothness of motion or movements\\
(3) dynamic degree, the degree of dynamic changes\\
(4) text-to-video alignment, the alignment between the text prompt and the video content\\
(5) factual consistency, the consistency of the video content with the common-sense and factual knowledge\\[2pt]
for each dimension, output a float number from 1.0 to 4.0, the higher the number is, the better the video performs in that sub-score, the lowest 1.0 means Bad, the highest 4.0 means Perfect/Real (the video is like a real video)\\
Here is an output example:\\
visual quality: 3.2\\
temporal consistency: 2.7\\
dynamic degree: 4.0\\
text-to-video alignment: 2.3\\
factual consistency: 1.8\\[2pt]
For this video, the text prompt is "\{caption\}", all the frames of video are as follows:
\end{promptbox}

\noindent\textbf{VideoScore2}~\cite{he2025videoscore2} uses one prompt for its three dimensions and answers after a
reasoning trace; we use \emph{text-to-video alignment} as SA and \emph{physical/common-sense consistency} as PC.

\begin{promptbox}{VideoScore2, prompt}
You are an expert for evaluating AI-generated videos from three dimensions:\\
(1) visual quality -- clarity, smoothness, artifacts;\\
(2) text-to-video alignment -- fidelity to the prompt;\\
(3) physical/common-sense consistency -- naturalness and physics plausibility.\\[2pt]
Video prompt: \{caption\}\\[2pt]
Please output in this format:\\
visual quality: <v\_score>;\\
text-to-video alignment: <t\_score>,\\
physical/common-sense consistency: <p\_score>
\end{promptbox}

\subsection{PhyRubric-Judge and Zero-shot VLMs}
\label{sup:suppl_prompts_ours}

PhyRubric-Judge and the zero-shot Qwen3-VL models receive the same system prompt and the same user turn (the frames
followed by the caption). The judge is fine-tuned on this format, while the zero-shot models see it as is.

\begin{promptbox}{PhyRubric-Judge / zero-shot VLMs}
You are a judge of generated videos. Given the caption used to generate a video and uniformly sampled frames of it, score two things and explain.\\[2pt]
Semantic adherence (sa\_score, 1-5): 5 = everything the caption describes is visibly shown; 4 = the main objects and events are shown, only a minor detail is off; 3 = one main element of the caption is missing or wrong; 2 = two or more main elements are missing or wrong, but the clip is a recognisable attempt at the caption; 1 = no discernible relation to the caption.\\[2pt]
Physical commonsense (pc\_score, 1-5): 5 = no visible violation of everyday physics; 4 = only minor side details are wrong (colour, texture or position of untouched things); 3 = exactly one clear violation, and it is not the central interaction; 2 = the central interaction the caption is about does not happen, happens in reverse, or its mechanism is broken, and nothing else serious is wrong; or two clear violations that are both not central; 1 = the scene is impossible as a whole (matter passing through matter, objects appearing, vanishing, morphing or merging, gravity in the wrong direction); or the central interaction fails together with another clear violation; or three or more clear violations.\\[2pt]
pc\_assessable: true if the central physical event of the caption appears in the clip so that its physics can be judged; false if it never appears. When false, pc\_score only reflects scene-wide violations.\\[2pt]
Judge only what is visible. Do not infer unseen forces or causes. Do not penalise image quality, camera motion, style, how far a process has advanced, or anything that would need a numeric measurement.\\[2pt]
Answer with one JSON object: \{"sa\_score": int, "pc\_assessable": bool, "pc\_score": int, "reason": str\}. The reason states what the clip shows, what from the caption is missing, and which physical violations are seen, with timestamps.\\[4pt]
<video>\\
Caption: \{caption\}
\end{promptbox}

\subsection{Data Construction}
\label{sup:prompts_data}

The following instructions were given to the LLMs that drafted the captions and the rubrics (Sec.~\ref{sup:construction}), and to the agentic annotator (Sec.~\ref{sup:annotator}); for the annotator, file paths and tool invocation details are omitted. \texttt{\{domain\}}, \texttt{\{category\}}, and \texttt{\{category\_note\}} are filled per category, and \texttt{\{caption\}}, \texttt{\{violation\_types\}}, \texttt{\{expectation\}}, and \texttt{\{examples\}} per caption.

\begin{promptbox}{Caption drafting}
You write captions that test the physical commonsense of text-to-video models. Each caption describes an everyday scene with one visible dynamic event, and will later be rendered by video models and judged against a rubric.\\[2pt]
Domain: \{domain\}. Category: \{category\}. Scope of this category: \{category\_note\}\\[2pt]
Write \{n\} captions for this category. Each caption must satisfy all of the following.\\
1. One or two sentences in plain English. An everyday scene that a non-expert recognises from experience; no textbook demonstrations.\\
2. State the triggering event AND its visible outcome (falls off, is thrown forward, shatters, slides a short way and stops), so the phenomenon is guaranteed to be in frame and the ending is not left open.\\
3. Never state HOW the outcome happens physically (no ``accelerating'', ``in an arc'', ``bounces lower each time'', ``at the same moment as''). The manner belongs in the expectation, not the caption.\\
4. The outcome must hold for every reasonable rendering of the caption. Do not write an event whose outcome depends on a threshold (fast enough, hard enough), on skill, or on details the caption leaves open. If the outcome is conditional, change the scene or drop it.\\
5. No law names, no numbers or units, no decorative adjectives, no camera or shot directions (close-up, slow motion, the camera follows). Describe the scene, not how to film it.\\
6. The visible change must fit in a clip of about five seconds from one fixed viewpoint. Describe slow processes as an ongoing state or a before-and-after result, not as a completed action.\\
7. Include a person only when the human body itself is part of the physics (a hand warming ice, breath fogging a window). Otherwise use objects, animals, or an unseen actor.\\
8. Vary the objects, settings, and triggers across captions; no two captions should share the same scene.\\[2pt]
For each caption also give:\\
-- violation\_types: one to three tags from \{permanence\_count, shape\_identity, solidity, causeless\_motion, gravity\_direction, trajectory\_momentum, process\_direction, missing\_consequence, material\_response, attribute\_drift\}, naming the expectations this event puts at risk.\\
-- expectation: one or two sentences on HOW the stated outcome should unfold: its timing relative to the cause, its direction, whether it is gradual or abrupt, and what must not happen. It must hold for every reasonable rendering. It is never shown to the video model or the judge.\\[2pt]
Example:\\
caption: A block is pulled out of the bottom row of a tall wooden block tower and the tower collapses.\\
violation\_types: gravity\_direction, missing\_consequence\\
expectation: The collapse starts as the support is removed, the tower sways then falls, and the blocks scatter on the table; it does not stand for long on nothing.\\[2pt]
Return a JSON list of \{"caption", "violation\_types", "expectation"\}.
\end{promptbox}

\begin{promptbox}{Rubric drafting}
You write the rubric for one caption. The rubric has two groups of items. SA items ask whether the clip shows what the caption says; PC items ask whether what the clip shows is physically possible. Scores are derived from the item verdicts afterwards and are never rated directly.\\[2pt]
Caption: \{caption\}. Category: \{category\}. Scope of this category: \{category\_note\}. Violation types: \{violation\_types\}. Expectation: \{expectation\}.\\
Approved rubrics from the same domain, to match in structure, length and tone: \{examples\}\\[2pt]
SA items (3--6; at least two primary and one secondary).\\
-- One item per visible fact the caption requires: the objects, the action, the stated outcome. Nothing the caption does not say, and nothing about physics or quality.\\
-- Primary: facts without which the clip is not this caption. Secondary: qualifiers the caption states (once, slowly, wooden), stated counts, and, when the caption names several sub-actions, one item asking whether they happen in the caption's order.\\
-- A secondary item never restates a physical expectation, is never hypothetical, and never depends on an event the caption does not include. Every item must be decidable on any clip.\\[2pt]
PC items (six; a seventh scene-wide fatal item is appended automatically), in this order:\\
1. Core items, one per violation type that is the central mechanism, built from the expectation, in the order the event unfolds.\\
2. Major items: the accompanying physics of side elements that this scene puts at risk (gravity on a secondary object, material response, a secondary consequence).\\
3. One fatal item naming the scene's objects: nothing appears, vanishes, merges, morphs, or passes through anything.\\
4. One minor attribute-drift item: colour, texture, label and position of untouched things.\\
Each PC item has an expect and a fail. Expect says what a correct clip shows and thereby which moment, region or event the item is about. Fail is the only ground for a FAIL: two to four concrete visible events joined by ``; or'', each something a viewer can point at in the frames. Fail is not the negation of expect. Judge direction and order only, never magnitude or rate: a correct clip in which the process has barely advanced must pass. Never require an unseen cause (heat, current, sound, wind, a hand) to be visible. Blur, flicker, resolution, colour cast, style, and self-driven agents starting or stopping are not violations. Do not use vague words (unrealistic, implausible, properly, correctly, natural, smoothly); name the event. No two items share a fail text. Add a precondition only for scene content the caption does not guarantee; such an item is not\_shown when the precondition does not occur.\\[2pt]
Present tense, plain words, no numbers with units, no camera or image-quality language. Return the rubric as JSON in the format of the examples.
\end{promptbox}

\begin{promptbox}{Agentic annotator}
You label generated videos against a rubric, from the video frames alone. Work independently: you never see the labels of other videos or other labellers.\\
Do not judge from the caption. Captions say what was requested; generated videos often do something else. Mark a pass only when you can point to what you saw.\\[2pt]
Input per clip: caption; branch note: \{branch\_note\}; SA items S1.. (primary or secondary, with a check); PC items P1.. (severity, type, expect, fail, optional precondition).\\[2pt]
Decision rules.\\
-- SA item: pass if the clip visibly shows exactly what the check says, else fail. SA is never not\_shown. Read each check literally: a loosely similar action does not count (a hand resting on a car is not pushing it). A clip showing a different scene fails SA items and is NOT penalised in PC for that.\\
-- PC item, three steps. (1) Is the situation the item is about in the clip at all (the precondition occurs; the moment, region or event in expect appears)? If not: not\_shown. not\_shown means the situation is absent; it does not mean the correct behaviour is missing, and it does not mean unsure. (2) Situation present and you SEE an event listed in fail: fail, and copy the matching clause into matched\_fail. The fail text is the only ground for a fail. (3) Otherwise pass.\\
-- Something that looks wrong but that no fail clause names goes into uncovered, never into a fail. Before writing anything into uncovered, go through the fail clauses of every PC item once more.\\
-- End state present from the first frame: when the effect the caption describes is on screen from frame 0 and the action that should cause it never happens, the central event is absent. The SA item for the trigger fails; PC items tied to the trigger moment are not\_shown.\\
-- Judge only what is visible. Direction and order of change matter, not speed or how far it got. Not penalised: image quality (blur, flicker, colour cast); a heat source, power supply, sound, wind, breath, or hand that is not visible; camera motion; style and set dressing; anything that needs a numeric measurement; self-driven agents starting, stopping or turning.\\[2pt]
How to look (small objects and fast actions are easy to miss). Tools:\\
\hspace*{1em}overview(N=16): N evenly spaced frames of the whole clip, with timestamps.\\
\hspace*{1em}zoom(t0, t1, [box]): full-rate frames between t0 and t1 seconds, optionally cropped to a box given as fractions, at most 16 frames.\\
For EVERY clip: (a) make one overview and describe the whole clip to yourself before reading the items in detail; (b) then make at least one zoom on the region and time window where the key event should be; (c) zoom again if a decision hinges on a moment you have not seen at full rate. Do not decide an item from the overview alone when the object involved is small or the action is fast.\\[2pt]
Output per clip, as JSON:\\
\{"clip\_summary": "3-5 sentences: what the video actually shows, with timestamps",\\
\hspace*{1em}"items": [\{"id": "S1", "status": "pass|fail", "evidence": "one or two sentences saying what you SAW and when, not restating the rubric"\},\\
\hspace*{1em}\hspace*{1em}\{"id": "P1", "status": "pass|fail|not\_shown", "evidence": "...", "matched\_fail": "clause text or null"\}, ... every SA and PC item exactly once],\\
\hspace*{1em}"uncovered": [\{"what": "...", "when": "..."\}]\}
\end{promptbox}

%% file: Tab/benchmark.tex
\begin{table*}[t]
\centering
\caption{Comparison with prior physical commonsense benchmarks for video generation. Mechanics, Optics, Heat, Electricity, Acoustics: coverage of the five physical domains. Commonsense only: every caption tests physical commonsense that can be judged from everyday experience, without requiring formal physics knowledge. Rubric: each caption is paired with a human-verified rubric of verifiable items with explicit severities, from which the scores are derived rather than rated holistically. Train set: labeled training data released for training a VLM judge. Explainable: the judge reports which requirement fails and the supporting evidence. Open-source: a physics-specific judge is trained and its weights are released. \cmark: yes; \pmark: partially; \xmark: no; --: not applicable (human evaluation only, no VLM judge).}
\label{tab:benchmark}
\resizebox{1.0\textwidth}{!}{
\begin{tabular}{l ccccc c c c cc}
\toprule
 & \multicolumn{8}{c}{\textbf{Data}} & \multicolumn{2}{c}{\textbf{VLM judge}} \\
\cmidrule(lr){2-9} \cmidrule(lr){10-11}
\textbf{Benchmark} & \textbf{Mechanics} & \textbf{Optics} & \textbf{Heat} & \textbf{Electricity} & \textbf{Acoustics} & \begin{tabular}[c]{@{}c@{}}\textbf{Commonsense}\\\textbf{only}\end{tabular} & \textbf{Rubric} & \textbf{Train set} & \textbf{Explainable} & \textbf{Open-source} \\
\midrule
VideoPhy~\cite{bansal2025videophy}       & \cmark & \xmark & \xmark & \xmark & \xmark & \cmark & \xmark & \cmark & \xmark & \cmark \\
VideoPhy-2~\cite{bansal2026videophy}     & \cmark & \xmark & \xmark & \xmark & \xmark & \cmark & \xmark & \cmark & \xmark & \cmark \\
PhyWorldBench~\cite{gu2026phyworldbench} & \cmark & \cmark & \pmark & \pmark & \xmark & \xmark & \xmark & \xmark & \pmark & \xmark \\
PhyGenBench~\cite{meng2024towards}       & \cmark & \cmark & \cmark & \xmark & \xmark & \xmark & \xmark & \xmark & \xmark & \xmark \\
T2VPhysBench~\cite{guo2025t2vphysbench}  & \cmark & \cmark & \xmark & \xmark & \xmark & \xmark & \xmark & \xmark & -- & -- \\
PhysVidBench~\cite{sanli2026physvidbench} & \cmark & \xmark & \xmark & \xmark & \xmark & \cmark & \xmark & \xmark & \pmark & \xmark \\
\midrule
\rowcolor{gray!10}
PhyRubric (\textbf{Ours}) & \cmark & \cmark & \cmark & \cmark & \cmark & \cmark & \cmark & \cmark & \cmark & \cmark \\
\bottomrule
\end{tabular}
}
\end{table*}

%% file: Tab/taxonomy.tex
\begin{table*}[t]
\centering
\small
\caption{Taxonomy of PhyRubric-Bench. Each caption belongs to exactly one domain and one category.}
\label{tab:taxonomy}
\begin{tabular}{l p{0.62\linewidth} r r}
\toprule
Domain & Categories & \#Train & \#Eval \\
\midrule
Mechanics & gravity and fall, inertia, collision, friction and rolling, projectile, balance and toppling, tension, deformation, cutting and tearing, fluid, granular, gas and wind, biomechanics, soft material, occlusion and containment & 242 & 75 \\
Heat & melting and freezing, boiling and steam, condensation, burning and smoke, heat response & 200 & 50 \\
Optics & shadow, reflection, refraction and transparency, illumination, color mixing & 227 & 50 \\
Electricity & switch and light, plug and power, lightning, spark, magnet, static & 202 & 50 \\
Acoustics & struck vibration, sound-driven vibration, source motion & 76 & 20 \\
\midrule
Total & 34 categories & 947 & 245 \\
\bottomrule
\end{tabular}
\end{table*}

%% file: Tab/violation.tex
\begin{table*}[t]
\centering
\small
\caption{\textbf{Violation types of PhyRubric.} Each PC item has one type. Types are grouped by the order in which infants acquire physical knowledge: L1 core object principles, L2 everyday dynamics, and L3 attribute experience. \#Items counts the drafted PC items over all 1,192 rubrics, excluding the scene-wide item shared by every rubric.}
\label{tab:violation}
\begin{tabularx}{\textwidth}{@{}l c >{\raggedright\arraybackslash}X r@{}}
\toprule
Type & Level & Definition & \#Items \\
\midrule
Permanence \& count & L1 & Objects appear from nothing, vanish, teleport, duplicate, or merge. & 1,099 \\
Shape \& identity & L1 & Rigid objects morph or melt, change identity, or bodies grow or lose parts. & 492 \\
Solidity & L1 & Matter passes through matter. & 198 \\
Causeless motion & L1 & Inert objects start, stop, or change direction with nothing acting on them. & 752 \\
Gravity direction & L2 & Unsupported objects hover, rise, or fall the wrong way. & 735 \\
Trajectory \& momentum & L2 & Paths, speeds, or rebounds are inconsistent with how the object was set in motion. & 317 \\
Process direction & L2 & A process runs backwards, oscillates, or resets (\eg, melted ice refreezes). & 554 \\
Missing consequence & L2 & The central event does not produce its visible result, or produces it at the wrong place, time or direction. & 911 \\
Material response & L2 & A material behaves like another class (\eg, glass bends, water acts as a solid). & 800 \\
Attribute drift & L3 & The color, texture, or position of untouched objects changes between frames. & 1,301 \\
\midrule
Total & & & 7,159 \\
\bottomrule
\end{tabularx}
\end{table*}

%% file: figs/rubric_full.tex
\newcommand{\rubhead}[1]{\par\vspace{3pt}\noindent{\normalfont\bfseries #1}\par\vspace{1pt}}
\newcommand{\rubsa}[3]{\par\noindent\hangindent=10em\hangafter=1\makebox[2em][l]{#1}\makebox[8em][l]{[#2]}#3}
\newcommand{\rubpc}[3]{\par\vspace{3pt}\noindent #1\ \ [#2 | #3]}
\newcommand{\rubline}[2]{\par\noindent\hangindent=7em\hangafter=1\hspace*{2em}\makebox[5em][l]{#1}#2}
\begin{figure*}[t]
\begin{promptbox}{Caption: The key switch on the base of a desk lamp is turned, and the lamp lights.}
\rubhead{Semantic adherence (SA)}
\rubsa{S1}{primary}{A desk lamp with a key switch on its base.}
\rubsa{S2}{primary}{The key switch is turned.}
\rubsa{S3}{primary}{The lamp lights.}
\rubsa{S4}{secondary}{The lamp is dark before the turn.}
\rubsa{S5}{secondary}{The order is dark, turn, lit.}
\rubhead{Physical commonsense (PC)}
\rubpc{P1}{fatal}{permanence\_count}
\rubline{Expect:}{One lamp with one key switch; the lamp keeps its shape; the hand passes through nothing.}
\rubline{Fail:}{A second lamp appears; or the lamp vanishes; or the hand passes through the base; or the lamp changes shape.}
\rubpc{P2}{fatal}{permanence\_count}\ \ (scene-wide item, shared verbatim by all rubrics)
\rubline{Expect:}{Everything in view, named in the caption or not, keeps its number, identity and shape unless something visible acts on it, and nothing passes through anything else.}
\rubline{Fail:}{An object that nothing visible acts on appears from nothing; or vanishes while still in frame and not occluded; or duplicates or merges with another; or turns into a different object or a different shape between frames; or any object passes through another. Changes brought about by the caption's event or by a visible interaction, objects leaving the frame, and occlusion are not failures.}
\rubpc{P3}{core}{missing\_consequence}
\rubline{Expect:}{The lamp lights at the moment the key clicks over, and only then.}
\rubline{Fail:}{The key is turned and the lamp stays dark; or the lamp is lit before the key moves; or the lamp comes on with the key untouched; or the lamp comes on well after the turn with nothing else happening.}
\rubpc{P4}{core}{process\_direction}
\rubline{Expect:}{Once on, the lamp stays lit steadily for the rest of the clip.}
\rubline{Fail:}{The lamp goes off again by itself; or flickers on and off; or dims back to dark; or the key turns back by itself.}
\rubpc{P5}{major}{missing\_consequence}
\rubline{Expect:}{The lamp's light reaches its surroundings: the shade glows and the desk beneath brightens, with shadows pointing away from the lamp.}
\rubline{Fail:}{The bulb glows while the shade and desk stay as dark as before; or the surroundings brighten while the lamp stays dark; or the light comes from a place other than the lamp.}
\rubpc{P6}{major}{causeless\_motion}
\rubline{Expect:}{The lamp stays put as the key is turned; nothing else on the desk moves.}
\rubline{Fail:}{The lamp slides, tips or lifts at the turn; or the shade swings by itself; or items on the desk jump.}
\rubpc{P7}{minor}{attribute\_drift}
\rubline{Expect:}{The shade's color, the base and the desk stay the same apart from the light.}
\rubline{Fail:}{The shade changes color rather than brightness; or the base changes shape; or the desk's contents rearrange.}
\end{promptbox}
\caption{Full rubric for the caption ``The key switch on the base of a desk lamp is turned, and the lamp lights.'' Items are shown in descending order of severity, which differs from the drafting order in Sec.~\ref{sup:prompts_data} and does not affect scoring.}
\label{fig:rubric_full}
\end{figure*}

%% file: Tab/suppl_metric.tex
\begin{table*}[t]
\centering
\caption{Metrics reported by the original papers of prior judges (\cmark), and the full set we report (Ours), grouped into three families. Score: the judge score on which we compute each metric (S: soft; H: hard; pass line: each judge's own threshold, Tab.~\ref{tab:suppl_passlines}).}
\label{tab:suppl_metric_overview}
\resizebox{1.0\textwidth}{!}{
\begin{tabular}{ll c cccc c}
\toprule
\textbf{Family} & \textbf{Metric} & \textbf{Score} & \textbf{VideoCon-Physics} & \textbf{VideoPhy-2-AutoEval} & \textbf{VideoScore} & \textbf{VideoScore2} & \textbf{Ours} \\
\midrule
\multirow{2}{*}{A. Correlation} & Pearson $r$ & S / H & & \cmark & & \cmark & \cmark \\
 & Spearman $\rho$ & S / H & & & \cmark & & \cmark \\
\midrule
B. Ranking & ROC-AUC & S & \cmark & & & & \cmark \\
\midrule
\multirow{5}{*}{C. Agreement} & Binary Accuracy & pass line & & \cmark & & & \cmark \\
 & F1 (binary) & pass line & & \cmark & & & \cmark \\
\cmidrule(lr){2-8}
 & Exact accuracy & H & & & & \cmark & \cmark \\
 & Within-1 accuracy & H & & & & \cmark & \cmark \\
\bottomrule
\end{tabular}
}
\end{table*}

%% file: Tab/suppl_passlines.tex
\begin{table*}[t]
\centering
\caption{Native output, dimensions and pass line of each judge. The pass line is used only by binary agreement (family C). Integer judges without an official pass line use $4$, as for human labels. $^\ddagger$For exact and within-1 accuracy, VideoScore's 1--4 output is mapped to 1--5 by $\mathrm{round}(1.25x)$. $^\S$Soft score from its stated 1--5 distribution (Sec.~\ref{sup:suppl_baselines}).}
\label{tab:suppl_passlines}
\small
\setlength{\tabcolsep}{4pt}
\begin{tabular}{lllcc}
\toprule
\textbf{Judge} & \textbf{Dimension used (SA / PC)} & \textbf{Native output} & \textbf{Range} & \textbf{Pass line} \\
\midrule
Zero-shot VLMs & SA / PC (our prompt) & generated integer & 1--5 & 4 \\
VideoScore$^\ddagger$~\cite{he2024videoscore} & text-to-video alignment / factual consistency & regression output & $[1,4]$ & 4 after mapping \\
VideoScore2~\cite{he2025videoscore2} & T2V alignment / physical \& common-sense consistency & generated integer & 1--5 & 4 \\
VideoCon-Physics & SA / PC prompts & $p(\text{Yes})/(p(\text{Yes})+p(\text{No}))$ & $[0,1]$ & 0.5 (official) \\
VideoPhy-2-AutoEval & SA / PC & generated integer & 1--5 & 4 (official) \\
Gemini 4 Argon$^\S$~\cite{google2026gemini4argon} & SA / PC (our prompt) & generated integer & 1--5 & 4 \\
\midrule
PhyRubric-Judge (ours) & SA / PC & generated integer & 1--5 & 4 \\
\bottomrule
\end{tabular}
\end{table*}

%% file: Tab/familyA.tex
\begin{table}[t]
\centering
\caption{Family A: correlation between judge scores and human labels on PhyRubric-Bench. Hard: the generated integer score. Soft: the probability-weighted score (Eq.~\ref{eq:soft}), see Sec.~\ref{sup:suppl_baselines} for VideoCon-Physics, VideoScore, and Gemini 4 Argon. Higher is better. Per column, the best soft score is in \textbf{bold} and the best hard score is \underline{underlined}.}
\label{tab:suppl_judges_corr}
\small
\setlength{\tabcolsep}{3.5pt}
\begin{tabular}{ll cc cc}
\toprule
 & & \multicolumn{2}{c}{\textbf{SA}} & \multicolumn{2}{c}{\textbf{PC}} \\
\cmidrule(lr){3-4} \cmidrule(lr){5-6}
\textbf{Judge} & \textbf{Score} & $r$ & $\rho$ & $r$ & $\rho$ \\
\midrule
\multicolumn{6}{l}{\textit{Zero-shot VLMs}} \\
\multirow{2}{*}{Qwen3-VL-8B} & hard & 0.451 & 0.472 & 0.002 & 0.021 \\
 & soft & 0.459 & 0.495 & -0.001 & 0.043 \\
\multirow{2}{*}{Qwen3-VL-32B} & hard & 0.467 & 0.502 & 0.143 & 0.146 \\
 & soft & 0.487 & 0.546 & 0.143 & 0.143 \\
\multirow{2}{*}{Qwen3-VL-235B} & hard & 0.449 & 0.482 & 0.129 & 0.123 \\
 & soft & 0.464 & 0.540 & 0.137 & 0.130 \\
\midrule
\multicolumn{6}{l}{\textit{General video evaluators}} \\
VideoScore & soft & -0.015 & 0.001 & 0.110 & 0.153 \\
\multirow{2}{*}{VideoScore2} & hard & 0.314 & 0.334 & 0.106 & 0.105 \\
 & soft & 0.320 & 0.338 & 0.111 & 0.112 \\
\midrule
\multicolumn{6}{l}{\textit{Physics-specific evaluators}} \\
VideoCon-Physics & soft & 0.162 & 0.186 & 0.176 & 0.195 \\
\multirow{2}{*}{VideoPhy-2-AutoEval} & hard & 0.198 & 0.223 & 0.203 & 0.202 \\
 & soft & 0.210 & 0.245 & 0.232 & 0.237 \\
\midrule
\multicolumn{6}{l}{\textit{Proprietary VLM}} \\
\multirow{2}{*}{Gemini 4 Argon} & hard & \underline{0.580} & \underline{0.555} & \underline{0.251} & \underline{0.264} \\
 & soft & 0.604 & 0.566 & 0.311 & \textbf{0.333} \\
\midrule
\rowcolor{gray!10}
\cellcolor{gray!10} & hard & 0.543 & 0.553 & 0.244 & 0.246 \\
\rowcolor{gray!10}
\multirow{-2}{*}{\textbf{PhyRubric-Judge (ours)}} & soft & \textbf{0.608} & \textbf{0.603} & \textbf{0.317} & 0.317 \\
\bottomrule
\end{tabular}
\vspace{-8pt}
\end{table}

%% file: Tab/familyB.tex
\begin{table}[t]
\centering
\caption{Family B: ROC-AUC between soft judge scores and human labels, with labels binarised at $y \geq 4$. Notation as in Tab.~\ref{tab:suppl_judges_corr}. Higher is better ($0.5$: random); best per column in \textbf{bold}.}
\label{tab:suppl_judges_rank}
\small
\begin{tabular}{l cc}
\toprule
\textbf{Judge} & \textbf{SA} & \textbf{PC} \\
\midrule
\multicolumn{3}{l}{\textit{Zero-shot VLMs}} \\
Qwen3-VL-8B & 0.778 & 0.525 \\
Qwen3-VL-32B & 0.811 & 0.582 \\
Qwen3-VL-235B & 0.808 & 0.569 \\
\midrule
\multicolumn{3}{l}{\textit{General video evaluators}} \\
VideoScore & 0.477 & 0.589 \\
VideoScore2 & 0.684 & 0.557 \\
\midrule
\multicolumn{3}{l}{\textit{Physics-specific evaluators}} \\
VideoCon-Physics & 0.595 & 0.611 \\
VideoPhy-2-AutoEval & 0.633 & 0.632 \\
\midrule
\multicolumn{3}{l}{\textit{Proprietary VLM}} \\
Gemini 4 Argon & 0.839 & \textbf{0.680} \\
\midrule
\rowcolor{gray!10}
\textbf{PhyRubric-Judge (ours)} & \textbf{0.845} & \textbf{0.680} \\
\bottomrule
\end{tabular}
\end{table}

%% file: Tab/familyC.tex
\begin{table*}[t]
\centering
\caption{Family C: agreement with human labels on PhyRubric-Bench. (1) Binary: labels pass at $y \geq 4$ and each judge at its own pass line on its native output (Tab.~\ref{tab:suppl_passlines}; the hard score for generative judges); joint: a clip passes only if both SA and PC pass. (2) Score-level: exact and within-1 accuracy of the hard score against the human 1--5 label. VideoCon-Physics outputs a probability, which
has no 1--5 counterpart, so its score-level entries are left blank. Higher is better; best per column in \textbf{bold}.}
\label{tab:suppl_judges_agree}
\resizebox{0.9\textwidth}{!}{
\begin{tabular}{l cc cc cc cc cc}
\toprule
 & \multicolumn{4}{c}{\textbf{SA}} & \multicolumn{4}{c}{\textbf{PC}} & \multicolumn{2}{c}{\textbf{Joint}} \\
\cmidrule(lr){2-5} \cmidrule(lr){6-9} \cmidrule(lr){10-11}
 & \multicolumn{2}{c}{(1) Binary} & \multicolumn{2}{c}{(2) Score-level} & \multicolumn{2}{c}{(1) Binary} & \multicolumn{2}{c}{(2) Score-level} & \multicolumn{2}{c}{(1) Binary} \\
\cmidrule(lr){2-3} \cmidrule(lr){4-5} \cmidrule(lr){6-7} \cmidrule(lr){8-9} \cmidrule(lr){10-11}
\textbf{Judge} & Acc & F1 & Exact & W-1 & Acc & F1 & Exact & W-1 & Acc & F1 \\
\midrule
\multicolumn{11}{l}{\textit{Zero-shot VLMs}} \\
Qwen3-VL-8B & 0.576 & 0.443 & 0.238 & 0.632 & 0.585 & 0.731 & 0.435 & 0.596 & 0.546 & 0.354 \\
Qwen3-VL-32B & 0.533 & 0.440 & 0.287 & 0.587 & 0.601 & 0.696 & 0.390 & 0.559 & 0.592 & 0.392 \\
Qwen3-VL-235B & 0.522 & 0.431 & 0.264 & 0.565 & 0.604 & 0.751 & 0.569 & 0.605 & 0.481 & 0.341 \\
\midrule
\multicolumn{11}{l}{\textit{General video evaluators}} \\
VideoScore & 0.618 & 0.201 & 0.186 & 0.681 & 0.499 & 0.430 & 0.093 & 0.371 & 0.690 & 0.137 \\
VideoScore2 & 0.605 & 0.416 & 0.214 & 0.716 & 0.567 & 0.650 & 0.148 & 0.522 & 0.625 & 0.317 \\
\midrule
\multicolumn{11}{l}{\textit{Physics-specific evaluators}} \\
VideoCon-Physics & 0.679 & 0.295 & -- & -- & 0.470 & 0.314 & -- & -- & 0.814 & 0.125 \\
VideoPhy-2-AutoEval & 0.693 & 0.317 & 0.311 & 0.752 & 0.617 & 0.713 & 0.235 & 0.566 & 0.733 & 0.273 \\
\midrule
\multicolumn{11}{l}{\textit{Proprietary VLM}} \\
Gemini 4 Argon & \textbf{0.853} & 0.508 & 0.558 & \textbf{0.892} & 0.442 & 0.153 & 0.125 & 0.433 & \textbf{0.865} & 0.221 \\
\midrule
\rowcolor{gray!10}
\textbf{PhyRubric-Judge (ours)} & 0.836 & \textbf{0.547} & \textbf{0.614} & 0.846 & \textbf{0.648} & \textbf{0.757} & \textbf{0.599} & \textbf{0.636} & 0.844 & \textbf{0.498} \\
\bottomrule
\end{tabular}
}
\end{table*}

%% file: Tab/leaderboard_judge.tex
\begin{table}[t]
\centering
\caption{Leaderboard reproduced by judges. JP (\%) of each generator on PhyRubric-Bench, computed as in Tab.~\ref{tab:leaderboard} from human labels (Human) and from the hard scores of PhyRubric-Judge (Ours) and Gemini 4 Argon (Argon). The last three rows give the Pearson and Spearman correlation, and the mean absolute error (MAE, in JP points), of each judge's column against the human column across the ten generators. Both judges score all 2,433 clips.}
\label{tab:suppl_leaderboard_judge}
\small
\setlength{\tabcolsep}{5pt}
\begin{tabular}{l ccc}
\toprule
\textbf{Generator} & \textbf{Human} & \textbf{Ours} & \textbf{Argon} \\
\midrule
Seedance 2.5 & 44.20 & 32.30 & 12.98 \\
Veo 3.1 & 36.20 & 37.53 & 6.53 \\
HunyuanVideo-1.5-8B & 14.93 & 14.60 & 1.60 \\
Wan2.2-T2V-A14B & 14.87 & 20.13 & 1.60 \\
Cosmos-Predict2.5-14B & 12.22 & 19.34 & 2.93 \\
LTX-2.3-22B & 11.47 & 10.93 & 1.47 \\
Wan2.2-TI2V-5B & 7.87 & 15.27 & 1.20 \\
Cosmos-Predict2.5-2B & 6.69 & 11.87 & 2.19 \\
CogVideoX1.5-5B & 5.27 & 10.53 & 0.40 \\
LTX-Video-2B & 4.07 & 6.27 & 0.40 \\
\midrule
Pearson $r$ vs.\ human & -- & 0.93 & 0.94 \\
Spearman $\rho$ vs.\ human & -- & 0.87 & 0.81 \\
MAE vs.\ human & -- & 4.65 & 12.65 \\
\bottomrule
\end{tabular}
\end{table}

%% file: Tab/ablation.tex
\begin{table}[t]
\centering
\caption{Ablations of PhyRubric-Judge on PhyRubric-Bench. Pearson $r$ and ROC-AUC of the soft score against human labels, as in Tab.~\ref{tab:judge}. Every variant keeps the training recipe of Sec.~\ref{sec:exp_setup} and changes only the stated factor.}
\label{tab:ablation}
\small
\setlength{\tabcolsep}{3pt}
\begin{tabular}{l cc cc}
\toprule
 & \multicolumn{2}{c}{\textbf{SA}} & \multicolumn{2}{c}{\textbf{PC}} \\
\cmidrule(lr){2-3} \cmidrule(lr){4-5}
\textbf{Setting} & $r$ & AUC & $r$ & AUC \\
\midrule
Default (caption only) & 0.608 & 0.845 & 0.317 & 0.680 \\
\midrule
\multicolumn{5}{l}{\textit{(a) Rubric input}} \\
+ rubric input, scores only & 0.668 & 0.869 & 0.368 & 0.711 \\
+ rubric input, item verdicts & 0.568 & 0.821 & 0.254 & 0.629 \\
Zero-shot, caption only & 0.463 & 0.779 & 0.023 & 0.534 \\
Zero-shot, + rubric input & 0.368 & 0.688 & $-$0.071 & 0.447 \\
\midrule
\multicolumn{5}{l}{\textit{(b) Output format}} \\
Diagnosis first & 0.537 & 0.803 & 0.199 & 0.566 \\
Diagnosis as thinking & 0.547 & 0.804 & 0.199 & 0.570 \\
\midrule
\multicolumn{5}{l}{\textit{(c) Training objective}} \\
SFT, then GRPO (thinking-free) & 0.595 & 0.841 & 0.314 & 0.684 \\
SFT, then GRPO (with thinking) & 0.543 & 0.805 & 0.086 & 0.501 \\
\midrule
\multicolumn{5}{l}{\textit{(d) Training data}} \\
25\% of training data & 0.581 & 0.830 & 0.215 & 0.635 \\
50\% of training data & 0.588 & 0.837 & 0.247 & 0.653 \\
\bottomrule
\end{tabular}
\end{table}

%% file: figs/ablation_formats.tex
\begin{figure}[t]
\centering
\begin{promptbox}[unbreakable]{Input and output of each variant in Tab.~\ref{tab:ablation}}
\textbf{Default.} Input: frames + caption.\\
Output: \{"sa\_score", "pc\_assessable", "pc\_score", "reason"\}\medskip

\textbf{Rubric input, scores only.} Input: frames + caption + rubric.\\
Output: as Default.\medskip

\textbf{Rubric input, item verdicts.} Input: frames + caption + rubric.\\
Output: \{"summary", "items": [\\
\hphantom{Output: \{}\{"id": "S1", "status": "pass|fail", "evidence"\}, ...,\\
\hphantom{Output: \{}\{"id": "P1", "status": "pass|fail|not\_shown", "evidence"\}, ...]\}\\
Scores are derived from the verdicts with the scoring rules.\medskip

\textbf{Diagnosis first.} Input: as Default.\\
Output: \{"reason", "sa\_score", "pc\_assessable", "pc\_score"\}\medskip

\textbf{Diagnosis as thinking.} Input: as Default.\\
Output: <think> reason </think>\\
\hphantom{Output: }\{"sa\_score", "pc\_assessable", "pc\_score"\}\medskip

\textbf{SFT, then GRPO.} Input and output: as Default (thinking-free) or as Diagnosis as thinking (with thinking). GRPO changes the training objective, not the format.
\end{promptbox}
\caption{Input and output format of the ablation variants. The variants differ in whether the rubric is given and in where the scores and the diagnosis appear. The diagnosis text (\texttt{reason}) is identical wherever it is produced.}
\label{fig:ablation_formats}
\end{figure}

%% file: Tab/regen.tex
\begin{table}[t]
\centering
\caption{Regeneration with a diagnosis-driven rewrite on 100 captions. Mean SA and PC and pass rates (\%) of the blind labels for the original Wan2.2-T2V-A14B video and the two re-rendered conditions. $^*$: differs from the original video, $^\dagger$: differs from the new seed, at $p<0.05$ (paired Wilcoxon signed-rank test over captions). No PC difference is significant.}
\label{tab:regen}
\small
\setlength{\tabcolsep}{4pt}
\begin{tabular}{l cc ccc}
\toprule
 & \multicolumn{2}{c}{\textbf{Mean}} & \multicolumn{3}{c}{\textbf{Pass rate}} \\
\cmidrule(lr){2-3} \cmidrule(lr){4-6}
\textbf{Condition} & SA & PC & SA & PC & Joint \\
\midrule
Original video & 2.55 & 3.37 & 11 & 51 & 6 \\
New seed & 2.76$^{*}$ & 3.29 & 18 & 50 & 10 \\
Rewrite + diagnosis & 3.19$^{*\dagger}$ & 3.37 & 31 & 51 & 18 \\
\bottomrule
\end{tabular}
\end{table}

%% file: figs/regen_case1.tex
\begin{figure*}[t]
\centering
\includegraphics[width=\textwidth]{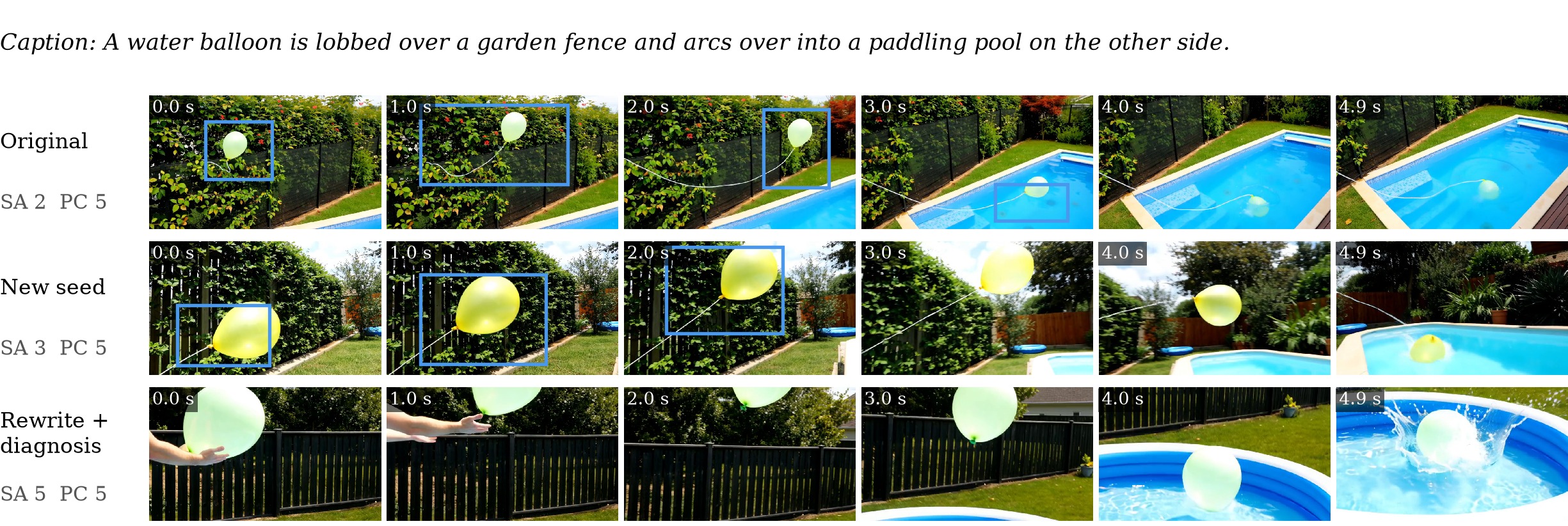}
\vspace{2pt}
\begin{tcolorbox}[enhanced, colback=white, colframe=black!60, boxrule=0.5pt, arc=0pt, sharp corners, colbacktitle=white, coltitle=black, fonttitle=\bfseries\footnotesize, titlerule=0.3pt, titlerule style={black!60}, left=6pt, right=6pt, top=3pt, bottom=3pt, toptitle=2pt, bottomtitle=2pt, fontupper=\footnotesize\ttfamily, title={Inputs of the two re-renders}]
\textbf{Caption (A, with a new seed).} A water balloon is lobbed over a garden fence and arcs over into a paddling pool on the other side.\medskip

\textbf{Diagnosis of the original video by PhyRubric-Judge (given to C).} A pale green balloon on a white string floats in front of a black fence and moves right and down across the frame, the camera panning with it. It reaches the pool edge at about 2.7s and drops into the water by 3.0s, then floats on the surface with the string trailing to the left. No lob or throw is seen. Missing from the caption: The balloon floats on a string in front of the fence; no throw or lob is seen. The balloon drifts down to the pool edge and drops in; no arc over the fence is seen. No lob is seen, so no lob-then-arc order. No violation of everyday physics is visible.\medskip

\textbf{Rewrite + diagnosis (C).} In a sunny garden, a person's hand lobs a pale green water balloon over a tall black fence. First the arm swings and releases the balloon, then it flies in a high arc above the fence and drops down into a paddling pool on the other side, splashing into the water.
\end{tcolorbox}
\caption{An example of regeneration. Wan2.2-T2V-A14B re-renders the caption from the original caption (Original), with a new seed (New seed), and from a prompt rewritten with the judge's diagnosis (Rewrite + diagnosis). Blue boxes mark SA violations, and the box lists the caption, the diagnosis, and the rewritten prompt. In the original video, the balloon hangs on a string and is never thrown, and a new seed changes little. The rewrite that sees the diagnosis produces the throw, the arc, and the splash.}
\label{fig:regen_case1}
\end{figure*}

%% file: Tab/error_types.tex
\begin{table}[t]
\centering
\caption{Violation types by domain on PhyRubric-Bench. Each cell is the percentage of assessable videos in that domain with at least one failed PC item of the given type, according to the human labels. Types are ordered by overall frequency; the highest domain of each type is in bold.}
\label{tab:error_types}
\small
\setlength{\tabcolsep}{4pt}
\resizebox{\linewidth}{!}{%
\begin{tabular}{l|ccccc|c}
\toprule
\textbf{Violation type} & Mech. & Opt. & Heat & Elec. & Acou. & All \\
\midrule
Missing consequence    &  9.1 & 16.1 & 13.5 & 24.6 & \textbf{44.4} & 17.3 \\
Permanence \& count    & \textbf{18.5} & 14.2 &  8.3 & 15.8 & 15.5 & 14.7 \\
Causeless motion       & 10.9 &  5.4 &  4.2 & 11.0 & \textbf{21.8} &  9.2 \\
Process direction      &  1.8 &  4.1 & 11.2 & \textbf{13.0} &  1.4 &  6.4 \\
Attribute drift        &  4.5 & \textbf{ 9.3} &  3.9 &  5.6 &  2.8 &  5.5 \\
Material response      & \textbf{ 8.6} &  1.6 &  4.9 &  2.0 &  3.5 &  4.7 \\
Gravity direction      &  5.9 &  1.0 &  4.4 & \textbf{ 6.5} &  5.6 &  4.7 \\
Shape \& identity      &  4.1 & \textbf{11.7} &  2.3 &  2.3 &  0.0 &  4.7 \\
Trajectory \& momentum & \textbf{ 8.2} &  0.5 &  0.3 &  2.5 &  3.5 &  3.5 \\
Solidity               &  2.5 &  0.3 &  0.0 & \textbf{ 2.8} &  0.0 &  1.4 \\
\midrule
\#Assessable videos    &  558 &  386 &  384 &  354 &  142 & 1,824 \\
\bottomrule
\end{tabular}}
\end{table}

%% file: Tab/suppl_error_types_gen.tex
\begin{table*}[t]
\centering
\caption{Violation types by generator on PhyRubric-Bench. Each cell is the percentage of assessable videos of that generator with at least one failed PC item of the given type, according to the human labels. The last row gives the number of assessable videos.}
\label{tab:error_types_gen}
\small
\setlength{\tabcolsep}{5pt}
\resizebox{\textwidth}{!}{%
\begin{tabular}{l|cc|cccccccc}
\toprule
 & Seedance & Veo & LTX-Video & Cosmos- & Wan2.2- & CogVideoX & Hunyuan & Wan2.2- & Cosmos- & LTX-2.3 \\
\textbf{Violation type} & 2.5 & 3.1 & 2B & Predict2.5-2B & TI2V-5B & 1.5-5B & Video-1.5-8B & T2V-A14B & Predict2.5-14B & 22B \\
\midrule
Missing consequence    & 11.3 & 14.7 & 22.1 & 14.4 & 22.2 & 17.6 & 20.0 & 16.3 & 18.8 & 18.7 \\
Permanence \& count    &  5.2 & 12.9 & 22.8 & 13.8 & 22.2 & 33.3 &  8.6 & 10.7 &  8.6 & 16.3 \\
Causeless motion       &  5.7 &  6.7 & 10.3 & 10.0 & 10.8 & 12.1 &  7.6 & 12.8 &  9.1 &  9.0 \\
Process direction      &  3.0 &  7.1 &  2.2 &  9.4 &  9.7 &  6.1 &  5.4 &  5.6 &  9.7 &  6.0 \\
Attribute drift        &  2.6 &  2.7 &  8.8 &  3.8 &  9.7 & 12.7 &  3.8 &  4.1 &  4.3 &  5.4 \\
Material response      &  2.6 &  0.9 &  7.4 &  6.2 &  9.7 &  7.9 &  2.7 &  4.6 &  3.8 &  3.6 \\
Gravity direction      &  2.2 &  2.2 &  7.4 &  4.4 &  4.5 &  9.1 &  3.2 &  7.1 &  4.3 &  4.2 \\
Shape \& identity      &  1.7 &  2.7 & 10.3 &  1.9 &  5.1 &  9.7 &  5.9 &  3.6 &  3.2 &  5.4 \\
Trajectory \& momentum &  1.7 &  4.0 &  4.4 &  3.1 &  4.5 &  4.2 &  3.2 &  3.6 &  3.8 &  2.4 \\
Solidity               &  0.0 &  2.2 &  0.7 &  0.6 &  2.3 &  1.2 &  1.1 &  2.0 &  1.1 &  2.4 \\
\midrule
\#Assessable videos    &  230 &  224 &  136 &  160 &  176 &  165 &  185 &  196 &  186 &  166 \\
\bottomrule
\end{tabular}}
\end{table*}

%% file: figs/fig_diag_examples.tex
\begin{figure*}[t]
\centering
\footnotesize
\includegraphics[width=\textwidth]{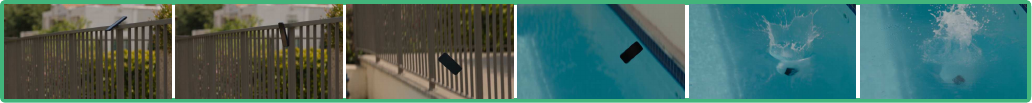}\\[-2pt]
\begin{promptbox}[unbreakable, fontupper=\scriptsize\ttfamily, top=1pt, bottom=1pt, left=4pt, right=4pt, toptitle=1pt, bottomtitle=1pt]{(a) Seedance 2.5, ``A phone balanced on a balcony rail slips off and falls into the swimming pool below with a splash.'' \hfill human: SA~5, PC~5}
\{"sa\_score": 5, "pc\_assessable": true, "pc\_score": 5, "reason": "A phone lies on a grey rail with a blurred garden behind it. At about 1.0s it tips and tumbles end over end, the camera pans and the rail goes out of frame by 2.0s. It falls through the air (2.0-2.7s) and enters the pool at about 2.7s. A splash and crown form at 3.0-3.7s, the phone sits in the water at the centre of the crown, and the crown fades by 5.0s. Everything the caption describes is shown. No violation of everyday physics is visible."\}
\end{promptbox}
\vspace{-2pt}
\includegraphics[width=\textwidth]{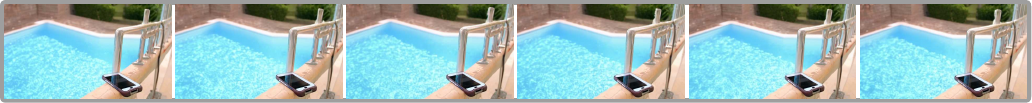}\\[-2pt]
\begin{promptbox}[unbreakable, fontupper=\scriptsize\ttfamily, top=1pt, bottom=1pt, left=4pt, right=4pt, toptitle=1pt, bottomtitle=1pt]{(b) CogVideoX1.5-5B, ``A phone balanced on a balcony rail slips off and falls into the swimming pool below with a splash.'' \hfill human: SA~2, PC~5 (event not shown)}
\{"sa\_score": 2, "pc\_assessable": false, "pc\_score": 5, "reason": "A phone in a case sits on a tan balcony rail beside a metal handrail, with a pool behind. It stays in place for the whole 5 s; it never slips off the rail and never falls. Missing from the caption: The phone stays on the rail for the full clip. No fall occurs. No splash appears. No slip, fall or splash sequence. The central physical event of the caption never occurs, so its physics cannot be judged."\}
\end{promptbox}
\vspace{-2pt}
\includegraphics[width=\textwidth]{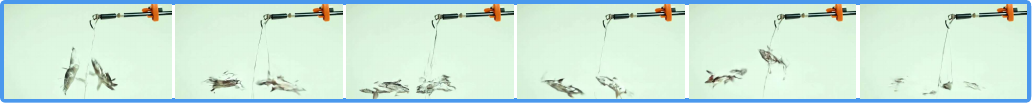}\\[-2pt]
\begin{promptbox}[unbreakable, fontupper=\scriptsize\ttfamily, top=1pt, bottom=1pt, left=4pt, right=4pt, toptitle=1pt, bottomtitle=1pt]{(c) CogVideoX1.5-5B, ``A toy fishing rod with a magnet on its line is dangled over paper fish fitted with steel clips, and a fish jumps up to the magnet and hangs from it.'' \hfill human: SA~2, PC~1}
\{"sa\_score": 2, "pc\_assessable": true, "pc\_score": 1, "reason": "A toy fishing rod with an orange handle hangs at the top right against a pale background. Several paper fish with clips are already in the air around the line from 0.3s, tumbling and flapping. A fish is pulled up to the hook at about 1.0-1.3s and 2.0-2.3s, then drops back down. At 3.3-3.7s a fish is pulled up again and hangs at the hook, then drops at 4.0s. The fish never hang from the line. Missing from the caption: Fish are already airborne at 0.3s and are not on paper lying under the rod. Fish are pulled up to the hook but never hang from the line. No fish is seen jumping up to the magnet. Physical violations: Fish are airborne and tumbling at 0.3-0.7s with nothing touching them. A fish is pulled up to the hook at 1.0-1.3s and 2.0-2.3s with no visible magnet."\}
\end{promptbox}
\vspace{-2pt}
\includegraphics[width=\textwidth]{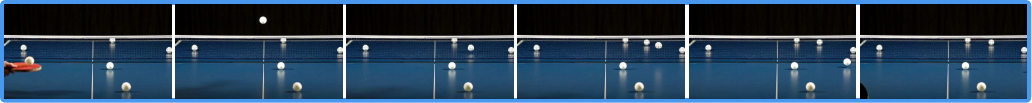}\\[-2pt]
\begin{promptbox}[unbreakable, fontupper=\scriptsize\ttfamily, top=1pt, bottom=1pt, left=4pt, right=4pt, toptitle=1pt, bottomtitle=1pt]{(d) Wan2.2-T2V-A14B, ``A ping-pong ball is served, arcs over the net, bounces once on the far side of the table, and drops off the end.'' \hfill human: SA~3, PC~1}
\{"sa\_score": 2, "pc\_assessable": true, "pc\_score": 1, "reason": "A blue table with a net and several balls. A paddle enters at 0.3s and a ball rises over the net at 0.7-1.3s. From 1.3s to 4.7s many balls appear and move around the table: a ball at the far end at 1.3s, another at 1.7s, a ball at the far end at 2.3s, and a ball at the far end at 3.7s. A paddle enters at the bottom at 4.7s. Missing from the caption: No clear bounce on the far side; balls just appear and move. No clear drop off the end; balls stay on the table. No serve, arc, bounce, drop sequence is clear. Physical violations: Many balls appear and move around the table (1.3-4.7s) and the ball count changes. A ball at the far end at 1.3s is gone by 1.7s and another ball appears at 1.7s."\}
\end{promptbox}
\caption{Outputs of PhyRubric-Judge on four videos from PhyRubric-Bench; (a)--(c) also appear in Figs.~\ref{fig:qual_mech}--\ref{fig:qual_elec}. Each block shows six frames of the video (border color as in Fig.~\ref{fig:qual_mech}), the generator, the caption and the human scores, followed by the judge's output verbatim. The diagnosis describes what the video shows with timestamps, lists what is missing from the caption, and names each physical violation: (a) the event is shown and passes; (b) the event never occurs, so PC is not assessable; (c) fish tumble in the air with nothing touching them; (d) balls appear and vanish on the table, so their number changes.}
\label{fig:diag_examples}
\end{figure*}

%% file: main.bib
@String(CVPR= {IEEE Conf. Comput. Vis. Pattern Recog.})

@String(ICCV= {Int. Conf. Comput. Vis.})

@String(CVPR  = {CVPR})

@String(ICCV  = {ICCV})

@inproceedings{huang2024vbench,
  title={Vbench: Comprehensive benchmark suite for video generative models},
  author={Huang, Ziqi and He, Yinan and Yu, Jiashuo and Zhang, Fan and Si, Chenyang and Jiang, Yuming and Zhang, Yuanhan and Wu, Tianxing and Jin, Qingyang and Chanpaisit, Nattapol and others},
  booktitle={2024 IEEE/CVF Conference on Computer Vision and Pattern Recognition (CVPR)},
  pages={21807--21818},
  year={2024},
  organization={IEEE}
}

@article{zheng2025vbench,
  title={Vbench-2.0: Advancing video generation benchmark suite for intrinsic faithfulness},
  author={Zheng, Dian and Huang, Ziqi and Liu, Hongbo and Zou, Kai and He, Yinan and Zhang, Fan and Gu, Lulu and Zhang, Yuanhan and He, Jingwen and Zheng, Wei-Shi and others},
  journal={arXiv preprint arXiv:2503.21755},
  year={2025}
}

@article{huang2025vbench++,
  title={Vbench++: Comprehensive and versatile benchmark suite for video generative models},
  author={Huang, Ziqi and Zhang, Fan and Xu, Xiaojie and He, Yinan and Yu, Jiashuo and Dong, Ziyue and Ma, Qianli and Chanpaisit, Nattapol and Si, Chenyang and Jiang, Yuming and others},
  journal={IEEE Transactions on Pattern Analysis and Machine Intelligence},
  year={2025},
  publisher={IEEE}
}

@inproceedings{liu2024evalcrafter,
  title={Evalcrafter: Benchmarking and evaluating large video generation models},
  author={Liu, Yaofang and Cun, Xiaodong and Liu, Xuebo and Wang, Xintao and Zhang, Yong and Chen, Haoxin and Liu, Yang and Zeng, Tieyong and Chan, Raymond and Shan, Ying},
  booktitle={2024 IEEE/CVF Conference on Computer Vision and Pattern Recognition (CVPR)},
  pages={22139--22149},
  year={2024},
  organization={IEEE}
}

@article{wang2026very,
  title={A very big video reasoning suite},
  author={Wang, Maijunxian and Wang, Ruisi and Lin, Juyi and Ji, Ran and Wiedemer, Thadd{\"a}us and Gao, Qingying and Luo, Dezhi and Qian, Yaoyao and Huang, Lianyu and Hong, Zelong and others},
  journal={arXiv preprint arXiv:2602.20159},
  year={2026}
}

@inproceedings{bansal2025videophy,
  title={Videophy: Evaluating physical commonsense for video generation},
  author={Bansal, Hritik and Lin, Zongyu and Xie, Tianyi and Zong, Zeshun and Yarom, Michal and Bitton, Yonatan and Jiang, Chenfanfu and Sun, Yizhou and Chang, Kai-Wei and Grover, Aditya},
  booktitle={International Conference on Learning Representations},
  volume={2025},
  pages={102075--102121},
  year={2025}
}

@inproceedings{bansal2026videophy,
  title={Videophy-2: A challenging action-centric physical commonsense evaluation in video generation},
  author={Bansal, Hritik and Peng, Clark and Bitton, Yonatan and Goldenberg, Roman and Grover, Aditya and Chang, Kai-Wei},
  booktitle={International Conference on Learning Representations},
  volume={2026},
  pages={118456--118470},
  year={2026}
}

@article{meng2024towards,
  title={Towards world simulator: Crafting physical commonsense-based benchmark for video generation},
  author={Meng, Fanqing and Liao, Jiaqi and Tan, Xinyu and Shao, Wenqi and Lu, Quanfeng and Zhang, Kaipeng and Cheng, Yu and Li, Dianqi and Qiao, Yu and Luo, Ping},
  journal={arXiv preprint arXiv:2410.05363},
  year={2024}
}

@inproceedings{gu2026phyworldbench,
  title={$ PhyWorldBench $: A Comprehensive Evaluation of Physical Realism in Text-to-Video Models},
  author={Gu, Jing and Liu, Xian and Zeng, Yu and Nagarajan, Ashwin and Zhu, Fangrui and Hong, Daniel and Fan, Yue and Yan, Qianqi and Zhou, Kaiwen and Liu, Ming-Yu and others},
  booktitle={International Conference on Learning Representations},
  volume={2026},
  pages={75130--75164},
  year={2026}
}

@article{unterthiner2018towards,
  title={Towards accurate generative models of video: A new metric \& challenges},
  author={Unterthiner, Thomas and Van Steenkiste, Sjoerd and Kurach, Karol and Marinier, Raphael and Michalski, Marcin and Gelly, Sylvain},
  journal={arXiv preprint arXiv:1812.01717},
  year={2018}
}

@article{wang2004image,
  title={Image quality assessment: from error visibility to structural similarity},
  author={Wang, Zhou and Bovik, Alan C and Sheikh, Hamid R and Simoncelli, Eero P},
  journal={IEEE transactions on image processing},
  volume={13},
  number={4},
  pages={600--612},
  year={2004},
  publisher={IEEE}
}

@inproceedings{hessel2021clipscore,
  title={Clipscore: A reference-free evaluation metric for image captioning},
  author={Hessel, Jack and Holtzman, Ari and Forbes, Maxwell and Le Bras, Ronan and Choi, Yejin},
  booktitle={Proceedings of the 2021 conference on empirical methods in natural language processing},
  pages={7514--7528},
  year={2021}
}

@inproceedings{he2024videoscore,
  title={Videoscore: Building automatic metrics to simulate fine-grained human feedback for video generation},
  author={He, Xuan and Jiang, Dongfu and Zhang, Ge and Ku, Max and Soni, Achint and Siu, Sherman and Chen, Haonan and Chandra, Abhranil and Jiang, Ziyan and Arulraj, Aaran and others},
  booktitle={Proceedings of the 2024 Conference on Empirical Methods in Natural Language Processing},
  pages={2105--2123},
  year={2024}
}

@inproceedings{pothiraj2026physics,
  title={Physics question scene graph: Fine-grained evaluation of physical plausibility in text-to-video generation},
  author={Pothiraj, Atin and Cho, Jaemin and Zhang, Yue and Stengel-Eskin, Elias and Bansal, Mohit},
  booktitle={European Conference on Computer Vision},
  pages={301--318},
  year={2026},
  organization={Springer}
}

@article{chen2026holistic,
  title={From Holistic Evaluation to Structured Criteria: Rubrics Across the Evolving LLM Landscape},
  author={Chen, Hao and Han, Ziyu and Yan, Yukun and Zhu, Qingfu and Sun, Maosong and Che, Wanxiang},
  journal={arXiv preprint arXiv:2606.08625},
  year={2026}
}

@article{spelke2007core,
  title={Core knowledge},
  author={Spelke, Elizabeth S and Kinzler, Katherine D},
  journal={Developmental science},
  volume={10},
  number={1},
  pages={89--96},
  year={2007},
  publisher={Wiley Online Library}
}

@article{lin2022infants,
  title={Infants’ physical reasoning and the cognitive architecture that supports it},
  author={Lin, Yi and Stavans, Maayan and Baillargeon, Ren{\'e}e},
  journal={Cambridge handbook of cognitive development},
  pages={168--194},
  year={2022}
}

@article{kubricht2017intuitive,
  title={Intuitive physics: Current research and controversies},
  author={Kubricht, James R and Holyoak, Keith J and Lu, Hongjing},
  journal={Trends in cognitive sciences},
  volume={21},
  number={10},
  pages={749--759},
  year={2017},
  publisher={Elsevier}
}

@misc{openai2026gpt56,
  author       = {{OpenAI}},
  title        = {{GPT-5.6} System Card},
  year         = {2026},
  month        = jul,
  howpublished = {\url{https://deploymentsafety.openai.com/gpt-5-6}},
  note         = {Accessed: 2026-10-05}
}

@misc{anthropic2026sonnet,
  author       = {{Anthropic}},
  title        = {System Card: {Claude Sonnet 5.5}},
  year         = {2026},
  month        = sep,
  howpublished = {\url{https://www.anthropic.com/claude-sonnet-5-5}},
  note         = {Accessed: 2026-10-06}
}

@misc{anthropic2026fable,
  author       = {{Anthropic}},
  title        = {System Card: {Claude Fable 5.1} \& {Claude Mythos 5.1}},
  year         = {2026},
  month        = sep,
  howpublished = {\url{https://www.anthropic.com/claude-fable-and-mythos-5-1}},
  note         = {Accessed: 2026-10-05}
}

@article{bai2025qwen3,
  title={Qwen3-vl technical report},
  author={Bai, Shuai and Cai, Yuxuan and Chen, Ruizhe and Chen, Keqin and Chen, Xionghui and Cheng, Zesen and Deng, Lianghao and Ding, Wei and Gao, Chang and Ge, Chunjiang and others},
  journal={arXiv preprint arXiv:2511.21631},
  year={2025}
}

@inproceedings{yang2025cogvideox,
  title={Cogvideox: Text-to-video diffusion models with an expert transformer},
  author={Yang, Zhuoyi and Teng, Jiayan and Zheng, Wendi and Ding, Ming and Huang, Shiyu and Xu, Jiazheng and Yang, Yuanming and Hong, Wenyi and Zhang, Xiaohan and Feng, Guanyu and others},
  booktitle={International Conference on Learning Representations},
  volume={2025},
  pages={83048--83077},
  year={2025}
}

@article{wu2025hunyuanvideo,
  title={Hunyuanvideo 1.5 technical report},
  author={Wu, Bing and Zou, Chang and Li, Changlin and Huang, Duojun and Yang, Fang and Tan, Hao and Peng, Jack and Wu, Jianbing and Xiong, Jiangfeng and Jiang, Jie and others},
  journal={arXiv preprint arXiv:2511.18870},
  year={2025}
}

@article{hacohen2024ltx,
  title={Ltx-video: Realtime video latent diffusion},
  author={HaCohen, Yoav and Chiprut, Nisan and Brazowski, Benny and Shalem, Daniel and Moshe, Dudu and Richardson, Eitan and Levin, Eran and Shiran, Guy and Zabari, Nir and Gordon, Ori and others},
  journal={arXiv preprint arXiv:2501.00103},
  year={2024}
}

@article{wan2025wan,
  title={Wan: Open and advanced large-scale video generative models},
  author={Wan, Team and Wang, Ang and Ai, Baole and Wen, Bin and Mao, Chaojie and Xie, Chen-Wei and Chen, Di and Yu, Feiwu and Zhao, Haiming and Yang, Jianxiao and others},
  journal={arXiv preprint arXiv:2503.20314},
  year={2025}
}

@article{agarwal2025cosmos,
  title={Cosmos world foundation model platform for physical ai},
  author={Agarwal, Niket and Ali, Arslan and Bala, Maciej and Balaji, Yogesh and Barker, Erik and Cai, Tiffany and Chattopadhyay, Prithvijit and Chen, Yongxin and Cui, Yin and Ding, Yifan and others},
  journal={arXiv preprint arXiv:2501.03575},
  year={2025}
}

@article{hacohen2026ltx,
  title={Ltx-2: Efficient joint audio-visual foundation model},
  author={HaCohen, Yoav and Brazowski, Benny and Chiprut, Nisan and Bitterman, Yaki and Kvochko, Andrew and Berkowitz, Avishai and Shalem, Daniel and Lifschitz, Daphna and Moshe, Dudu and Porat, Eitan and others},
  journal={arXiv preprint arXiv:2601.03233},
  year={2026}
}

@article{he2025videoscore2,
  title={Videoscore2: Think before you score in generative video evaluation},
  author={He, Xuan and Jiang, Dongfu and Nie, Ping and Liu, Minghao and Jiang, Zhengxuan and Su, Mingyi and Ma, Wentao and Lin, Junru and Ye, Chun and Lu, Yi and others},
  journal={arXiv preprint arXiv:2509.22799},
  year={2025}
}

@misc{seedance25,
  title        = {One-Take Creation, Flexible Referencing: Introducing Seedance 2.5},
  author       = {{ByteDance Seed}},
  year         = {2026},
  howpublished = {\url{https://seed.bytedance.com/en/blog/one-take-creation-flexible-referencing-introducing-seedance-2-5}},
  note         = {Blog post; accessed 2026-10-23}
}

@inproceedings{liu2023g,
  title={G-eval: NLG evaluation using gpt-4 with better human alignment},
  author={Liu, Yang and Iter, Dan and Xu, Yichong and Wang, Shuohang and Xu, Ruochen and Zhu, Chenguang},
  booktitle={Proceedings of the 2023 conference on empirical methods in natural language processing},
  pages={2511--2522},
  year={2023}
}

@article{wu2023q,
  title={Q-align: Teaching lmms for visual scoring via discrete text-defined levels},
  author={Wu, Haoning and Zhang, Zicheng and Zhang, Weixia and Chen, Chaofeng and Liao, Liang and Li, Chunyi and Gao, Yixuan and Wang, Annan and Zhang, Erli and Sun, Wenxiu and others},
  journal={arXiv preprint arXiv:2312.17090},
  year={2023}
}

@article{chen2023videocrafter1,
  title={Videocrafter1: Open diffusion models for high-quality video generation},
  author={Chen, Haoxin and Xia, Menghan and He, Yingqing and Zhang, Yong and Cun, Xiaodong and Yang, Shaoshu and Xing, Jinbo and Liu, Yaofang and Chen, Qifeng and Wang, Xintao and others},
  journal={arXiv preprint arXiv:2310.19512},
  year={2023}
}

@article{zheng2024open,
  title={Open-sora: Democratizing efficient video production for all},
  author={Zheng, Zangwei and Peng, Xiangyu and Yang, Tianji and Shen, Chenhui and Li, Shenggui and Liu, Hongxin and Zhou, Yukun and Li, Tianyi and You, Yang},
  journal={arXiv preprint arXiv:2412.20404},
  year={2024}
}

@article{sun2024sora,
  title={From sora what we can see: A survey of text-to-video generation},
  author={Sun, Rui and Zhang, Yumin and Shah, Tejal and Sun, Jiahao and Zhang, Shuoying and Li, Wenqi and Duan, Haoran and Wei, Bo and Ranjan, Rajiv},
  journal={arXiv preprint arXiv:2405.10674},
  year={2024}
}

@article{zhu2024sora,
  title={Is sora a world simulator? a comprehensive survey on general world models and beyond},
  author={Zhu, Zheng and Wang, Xiaofeng and Zhao, Wangbo and Min, Chen and Li, Bohan and Deng, Nianchen and Dou, Min and Wang, Yuqi and Shi, Botian and Wang, Kai and others},
  journal={arXiv preprint arXiv:2405.03520},
  year={2024}
}

@article{ma2025step,
  title={Step-video-t2v technical report: The practice, challenges, and future of video foundation model},
  author={Ma, Guoqing and Huang, Haoyang and Yan, Kun and Chen, Liangyu and Duan, Nan and Yin, Shengming and Wan, Changyi and Ming, Ranchen and Song, Xiaoniu and Chen, Xing and others},
  journal={arXiv preprint arXiv:2502.10248},
  year={2025}
}

@article{baillargeon2004infants,
  title={Infants' physical world},
  author={Baillargeon, Ren{\'e}e},
  journal={Current directions in psychological science},
  volume={13},
  number={3},
  pages={89--94},
  year={2004},
  publisher={SAGE Publications, Sage CA: Los Angeles, CA}
}

@inproceedings{sun2025t2v,
  title={T2v-compbench: A comprehensive benchmark for compositional text-to-video generation},
  author={Sun, Kaiyue and Huang, Kaiyi and Liu, Xian and Wu, Yue and Xu, Zihan and Li, Zhenguo and Liu, Xihui},
  booktitle={2025 IEEE/CVF Conference on Computer Vision and Pattern Recognition (CVPR)},
  pages={8406--8416},
  year={2025},
  organization={IEEE}
}

@article{liu2023fetv,
  title={Fetv: A benchmark for fine-grained evaluation of open-domain text-to-video generation},
  author={Liu, Yuanxin and Li, Lei and Ren, Shuhuai and Gao, Rundong and Li, Shicheng and Chen, Sishuo and Sun, Xu and Hou, Lu},
  journal={Advances in Neural Information Processing Systems},
  volume={36},
  pages={62352--62387},
  year={2023}
}

@article{liao2024evaluation,
  title={Evaluation of text-to-video generation models: A dynamics perspective},
  author={Liao, Mingxiang and Lu, Hannan and Zhang, Xinyu and Wan, Fang and Wang, Tianyu and Zhao, Yuzhong and Zuo, Wangmeng and Ye, Qixiang and Wang, Jingdong},
  journal={Advances in Neural Information Processing Systems},
  volume={37},
  pages={109790--109816},
  year={2024}
}

@article{guo2025t2vphysbench,
  title={T2vphysbench: A first-principles benchmark for physical consistency in text-to-video generation},
  author={Guo, Xuyang and Huo, Jiayan and Shi, Zhenmei and Song, Zhao and Zhang, Jiahao and Zhao, Jiale},
  journal={arXiv preprint arXiv:2505.00337},
  year={2025}
}

@inproceedings{motamed2026generative,
  title={Do generative video models understand physical principles?},
  author={Motamed, Saman and Culp, Laura and Swersky, Kevin and Jaini, Priyank and Geirhos, Robert},
  booktitle={2026 IEEE/CVF Winter Conference on Applications of Computer Vision (WACV)},
  pages={948--958},
  year={2026},
  organization={IEEE}
}

@article{kang2024far,
  title={How far is video generation from world model: A physical law perspective},
  author={Kang, Bingyi and Yue, Yang and Lu, Rui and Lin, Zhijie and Zhao, Yang and Wang, Kaixin and Huang, Gao and Feng, Jiashi},
  journal={arXiv preprint arXiv:2411.02385},
  year={2024}
}

@inproceedings{km2026phyeduvideo,
  title={PhyEduVideo: A benchmark for evaluating text-to-video models for physics education},
  author={KM, Megha Mariam and Arun, Aditya and Laskar, Zakaria and Jawahar, CV},
  booktitle={2026 IEEE/CVF Winter Conference on Applications of Computer Vision (WACV)},
  pages={8690--8699},
  year={2026},
  organization={IEEE}
}

@inproceedings{sanli2026physvidbench,
  title={PhysVidBench: Language-Grounded Evaluation of Physical Commonsense in Text-to-Video Models},
  author={Sanli, Enes and Tezcan, Baris Sarper and Erdem, Erkut and Erdem, Aykut},
  booktitle={Findings of the Association for Computational Linguistics: EMNLP 2026},
  pages={45311--45359},
  year={2026}
}

@article{lin2026phyground,
  title={PhyGround: Benchmarking Physical Reasoning in Generative World Models},
  author={Lin, Juyi and Akbari, Arash and He, Yumei and Zhao, Lin and Zhang, Haichao and Akbari, Arman and Xu, Xingchen and Lu, Zoe Y and Nan, Enfu and Deng, Hokin and others},
  journal={arXiv preprint arXiv:2605.10806},
  year={2026}
}

@article{chen2024mllm,
  title={Mllm-as-a-judge: Assessing multimodal llm-as-a-judge with vision-language benchmark},
  author={Chen, Dongping and Chen, Ruoxi and Zhang, Shilin and Liu, Yinuo and Wang, Yaochen and Zhou, Huichi and Zhang, Qihui and Wan, Yao and Zhou, Pan and Sun, Lichao},
  journal={arXiv preprint arXiv:2402.04788},
  year={2024}
}

@inproceedings{pu2025judge,
  title={Judge anything: Mllm as a judge across any modality},
  author={Pu, Shu and Wang, Yaochen and Chen, Dongping and Chen, Yuhang and Wang, Guohao and Qin, Qi and Zhang, Zhongyi and Zhang, Zhiyuan and Zhou, Zetong and Gong, Shuang and others},
  booktitle={Proceedings of the 31st ACM SIGKDD Conference on Knowledge Discovery and Data Mining V. 2},
  pages={5742--5753},
  year={2025}
}

@inproceedings{han2025video,
  title={Video-bench: Human-aligned video generation benchmark},
  author={Han, Hui and Li, Siyuan and Chen, Jiaqi and Yuan, Yiwen and Wu, Yuling and Deng, Yufan and Leong, Chak Tou and Du, Hanwen and Fu, Junchen and Li, Youhua and others},
  booktitle={2025 IEEE/CVF Conference on Computer Vision and Pattern Recognition (CVPR)},
  pages={18858--18868},
  year={2025},
  organization={IEEE}
}

@article{jonsson2007use,
  title={The use of scoring rubrics: Reliability, validity and educational consequences},
  author={Jonsson, Anders and Svingby, Gunilla},
  journal={Educational research review},
  volume={2},
  number={2},
  pages={130--144},
  year={2007},
  publisher={Elsevier}
}

@inproceedings{kim2024prometheus,
  title={Prometheus: Inducing fine-grained evaluation capability in language models},
  author={Kim, Seungone and Shin, Jay and Jang, Joel and Longpre, Shayne and Lee, Hwaran and Yun, Sangdoo and Shin, Ryan and Kim, Sungdong and Thorne, James and Seo, Minjoon and others},
  booktitle={International Conference on Learning Representations},
  volume={2024},
  pages={29927--29962},
  year={2024}
}

@inproceedings{lee2024prometheus,
  title={Prometheus-vision: Vision-language model as a judge for fine-grained evaluation},
  author={Lee, Seongyun and Kim, Seungone and Park, Sue and Kim, Geewook and Seo, Minjoon},
  booktitle={Findings of the Association for Computational Linguistics: ACL 2024},
  pages={11286--11315},
  year={2024}
}

@inproceedings{kim2025biggen,
  title={The biggen bench: A principled benchmark for fine-grained evaluation of language models with language models},
  author={Kim, Seungone and Suk, Juyoung and Cho, Ji Yong and Longpre, Shayne and Kim, Chaeeun and Yoon, Dongkeun and Son, Guijin and Cho, Yejin and Shafayat, Sheikh and Baek, Jinheon and others},
  booktitle={Proceedings of the 2025 Conference of the Nations of the Americas Chapter of the Association for Computational Linguistics: Human Language Technologies (Volume 1: Long Papers)},
  pages={5877--5919},
  year={2025}
}

@article{cook2024ticking,
  title={Ticking all the boxes: Generated checklists improve llm evaluation and generation},
  author={Cook, Jonathan and Rockt{\"a}schel, Tim and Foerster, Jakob and Aumiller, Dennis and Wang, Alex},
  journal={arXiv preprint arXiv:2410.03608},
  year={2024}
}

@article{arora2025healthbench,
  title={Healthbench: Evaluating large language models towards improved human health},
  author={Arora, Rahul K and Wei, Jason and Hicks, Rebecca Soskin and Bowman, Preston and Qui{\~n}onero-Candela, Joaquin and Tsimpourlas, Foivos and Sharman, Michael and Shah, Meghan and Vallone, Andrea and Beutel, Alex and others},
  journal={arXiv preprint arXiv:2505.08775},
  year={2025}
}

@inproceedings{hu2023tifa,
  title={Tifa: Accurate and interpretable text-to-image faithfulness evaluation with question answering},
  author={Hu, Yushi and Liu, Benlin and Kasai, Jungo and Wang, Yizhong and Ostendorf, Mari and Krishna, Ranjay and Smith, Noah A},
  booktitle={2023 ieee/cvf international conference on computer vision (iccv)},
  pages={20349--20360},
  year={2023},
  organization={IEEE}
}

@inproceedings{cho2024davidsonian,
  title={Davidsonian scene graph: Improving reliability in fine-grained evaluation for text-to-image generation},
  author={Cho, Jaemin and Hu, Yushi and Baldridge, Jason and Garg, Roopal and Anderson, Peter and Krishna, Ranjay and Bansal, Mohit and Pont-Tuset, Jordi and Wang, Su},
  booktitle={International conference on learning representations},
  volume={2024},
  pages={15625--15645},
  year={2024}
}

@inproceedings{zhang2025human,
  title={Human-activity agv quality assessment: A benchmark dataset and an objective evaluation metric},
  author={Zhang, Zhichao and Sun, Wei and Li, Xinyue and Li, Yunhao and Ge, Qihang and Jia, Jun and Zhang, Zicheng and Ji, Zhongpeng and Sun, Fengyu and Jui, Shangling and others},
  booktitle={Proceedings of the 33rd ACM International Conference on Multimedia},
  pages={6771--6780},
  year={2025}
}

@article{ma2025evaluating,
  title={Evaluating and predicting distorted human body parts for generated images},
  author={Ma, Lu and Cao, Kaibo and Liang, Hao and Lin, Jiaxin and Li, Zhuang and Liu, Yuhong and Zhang, Jihong and Zhang, Wentao and Cui, Bin},
  journal={arXiv preprint arXiv:2503.00811},
  year={2025}
}

@article{hu2021lora,
  title={Lora: Low-rank adaptation of large language models},
  author={Hu, Edward J and Shen, Yelong and Wallis, Phillip and Allen-Zhu, Zeyuan and Li, Yuanzhi and Wang, Shean and Wang, Lu and Chen, Weizhu},
  journal={arXiv preprint arXiv:2106.09685},
  year={2021}
}

@article{pearson1895note,
  title={Note on regression and inheritance in the case of two parents},
  author={Pearson, Karl},
  journal={proceedings of the royal society of London},
  volume={58},
  pages={240--242},
  year={1895},
  publisher={JSTOR}
}

@article{hanley1982meaning,
  title={The meaning and use of the area under a receiver operating characteristic (ROC) curve.},
  author={Hanley, James A and McNeil, Barbara J},
  journal={Radiology},
  volume={143},
  number={1},
  pages={29--36},
  year={1982}
}

@misc{google2026gemini4argon,
  author       = {{Google DeepMind}},
  title        = {Gemini 4 Argon: Our Next Era of Frontier Intelligence},
  year         = {2026},
  month        = sep,
  howpublished = {\url{https://blog.google/innovation-and-ai/models-and-research/gemini-models/gemini-4-argon/}},
  note         = {Accessed: 2026-11-05}
}

@misc{google2025veo31,
  author       = {{Google DeepMind}},
  title        = {Veo 3.1},
  year         = {2025},
  month        = oct,
  howpublished = {\url{https://deepmind.google/models/veo/}},
  note         = {Accessed: 2026-11-05}
}

@article{shao2024deepseekmath,
  title={Deepseekmath: Pushing the limits of mathematical reasoning in open language models},
  author={Shao, Zhihong and Wang, Peiyi and Zhu, Qihao and Xu, Runxin and Song, Junxiao and Bi, Xiao and Zhang, Haowei and Zhang, Mingchuan and Li, YK and Wu, Yang and others},
  journal={arXiv preprint arXiv:2402.03300},
  year={2024}
}

@incollection{bode2025deltaai,
  title={DeltaAI: A National Resource for AI/ML Research},
  author={Bode, Brett and Bauer, Gregory and Herriott, Laura and Kindratenko, Volodymyr and Gropp, William},
  booktitle={Practice and Experience in Advanced Research Computing 2025: The Power of Collaboration},
  pages={1--4},
  year={2025}
}

@incollection{boerner2023access,
  title={Access: Advancing innovation: Nsf’s advanced cyberinfrastructure coordination ecosystem: Services \& support},
  author={Boerner, Timothy J and Deems, Stephen and Furlani, Thomas R and Knuth, Shelley L and Towns, John},
  booktitle={Practice and experience in advanced research computing 2023: Computing for the common good},
  pages={173--176},
  year={2023}
}
